\documentclass[twoside,11pt]{article}

\usepackage{jmlr2e}
\usepackage{algorithm}
\usepackage{algpseudocode}
\algrenewcommand\algorithmicindent{0.8em}

\algtext*{EndFunction}
\usepackage{varwidth}

\usepackage{booktabs}

\usepackage{tikz}
\usetikzlibrary{calc,fit,positioning}
\usetikzlibrary{bayesnet}

\usepackage{xspace}
\usepackage{textcomp}
\newcommand{\textapproxVGI}{\raisebox{0.5ex}{\texttildelow}}
\usepackage{multicol}

\usepackage{commath}
\usepackage{amsmath}					
\usepackage{mathtools}
\usepackage{bm}
\allowdisplaybreaks %

\usepackage{thm-restate}

\usepackage{xcolor}
\usepackage{xifthen}
\usepackage{verbatim}

\usepackage{lastpage}
\usepackage{hyperref}

\newcommand{\ie}{that is\xspace}
\newcommand{\eg}{e.g.\xspace}

\newcommand{\paragraphJMLR}[1]{\paragraph{#1}}

\definecolor{revisiontwocolor}{RGB}{175, 2, 2}
\newcommand\revisiontwo[1]{#1}

\makeatletter
\newcommand{\oset}[3][0ex]{%
  \mathrel{\mathop{#3}\limits^{
    \vbox to#1{\kern-2\ex@
    \hbox{$\scriptstyle#2$}\vss}}}}
\makeatother

\DeclareMathOperator*{\argmax}{arg\,max}

\DeclareMathOperator*{\idxop}{idx}
\newcommand\idx[1]{\idxop(#1)}

\newcommand\mc[1]{\multicolumn{1}{l}{#1}} %

\DeclarePairedDelimiterX{\infdivx}[2]{(}{)}{%
  #1\;\delimsize|\delimsize|\;#2%
}
\newcommand{\KLD}[2]{D_{\text{KL}}\infdivx{#1}{#2}}

\newcommand{\obs}{\text{obs}}
\newcommand{\mis}{\text{mis}}

\newcommand{\sik}{{(i, k)}}
\newcommand{\nj}{{\smallsetminus j}}
\newcommand{\thetab}{{\bm{\theta}}}
\newcommand{\phib}{{\bm{\phi}}}
\newcommand{\phibj}[1][]{%
    {\ifthenelse{\equal{#1}{}}{\phib_j}{\phib_j^{(#1)}}}%
}

\newcommand{\gammab}{{\bm{\gamma}}}
\newcommand{\pib}{{\bm{\pi}}}
\newcommand{\x}{{\bm{x}}}

\newcommand{\m}{{\bm{m}}}

\newcommand{\tx}{{\bm{\tilde x}}}

\newcommand{\xm}{\x_\mis}
\newcommand{\txm}{{\bm{\tilde x}_\mis}}

\newcommand{\xo}{\x_\obs}
\newcommand{\xz}{\x_z}
\newcommand{\txz}{{\bm{\tilde x}_z}}
\newcommand{\xj}{x_{j}}

\newcommand{\xnj}{\x_{\nj}}
\newcommand{\xmnj}{\x_{\mis\nj}}
\newcommand{\xonj}{\x_{\obs\nj}}
\newcommand{\txmnj}{\tx_{\mis\nj}}

\newcommand{\txnj}{{\tx_{\nj}}}
\newcommand{\txj}{{\tilde{x}_j}}

\newcommand{\pt}{{p_\thetab}}

\newcommand{\ps}{{p_*}}
\newcommand{\qp}{q_\phib}
\newcommand{\qpj}{q_{\phibj}}

\newcommand{\E}{{\mathbb{E}}}

\newcommand{\J}{{\mathcal{J}}}
\newcommand{\D}{\mathcal{D}}
\newcommand{\B}{\mathcal{B}}

\newcommand{\Lc}{\mathcal{L}}

\newcommand{\JVGI}{\J_\text{VGI}}

\newcommand{\hJVGI}{\hat \J_\text{VGI}}

\newcommand{\Lmcvi}{\Lc_\text{MCVI}}

\DeclareRobustCommand*{\VBGIVAE}{VBGI\nobreakdash-VAE\xspace}
\DeclareRobustCommand*{\VBGIVAEM}{VBGI\nobreakdash-VAE\nobreakdash-M\xspace}

\newcommand{\Fb}{\bm{F}}
\newcommand{\Fo}{\Fb_\obs}
\newcommand{\Fm}{\Fb_\mis}
\newcommand{\Psib}{\bm{\Psi}}
\newcommand{\Psio}{\Psib_\obs}
\newcommand{\Psim}{\Psib_\mis}
\newcommand{\z}{\bm{z}}

\newcommand{\mub}{\bm{\mu}}
\newcommand{\muo}{\mub_\obs}
\newcommand{\mum}{\mub_\mis}

\newcommand{\N}{\mathcal{N}}
\newcommand{\Sigmab}{\bm{\Sigma}}
\newcommand{\Cb}{\bm{C}}

\newcommand{\psidiag}{\text{diag}(\Psib)}

\newcommand{\ub}{\bm{u}}

\makeatletter
\newcommand*{\indep}{%
  \mathbin{%
    \mathpalette{\@indep}{}%
  }%
}
\newcommand*{\nindep}{%
  \mathbin{%
    \mathpalette{\@indep}{\not}%
  }%
}
\newcommand*{\@indep}[2]{%
  \sbox0{$#1\perp\m@th$}%
  \sbox2{$#1=$}%
  \sbox4{$#1\vcenter{}$}%
  \rlap{\copy0}%
  \dimen@=\dimexpr\ht2-\ht4-.2pt\relax
  \kern\dimen@
  {#2}%
  \kern\dimen@
  \copy0 %
} 
\makeatother

\jmlrheading{24}{2023}{1-\pageref{LastPage}}{11/21; Revised 3/23}{6/23}{21-1373}{Vaidotas Simkus, Benjamin Rhodes and Michael U. Gutmann}

\ShortHeadings{Variational Gibbs Inference}{Simkus, Rhodes and Gutmann}
\firstpageno{1}

\begin{document}

\title{Variational Gibbs Inference\\ for Statistical Model Estimation from Incomplete Data}

\author{\name Vaidotas Simkus \email vaidotas.simkus@ed.ac.uk \\
       \name Benjamin Rhodes \email ben.rhodes@ed.ac.uk \\
       \name Michael U.\ Gutmann \email michael.gutmann@ed.ac.uk \\
       \addr School of Informatics\\
       University of Edinburgh}

\editor{Mohammad Emtiyaz Khan}

\maketitle

\begin{abstract}%
Statistical models are central to machine learning with broad applicability across a range of downstream tasks. The models are controlled by free parameters that are typically estimated from data by maximum-likelihood estimation or approximations thereof. 
However, when faced with real-world data sets many of the models run into a critical issue: they are formulated in terms of fully-observed data, whereas in practice the data sets are plagued with missing data.
The theory of statistical model estimation from incomplete data is conceptually similar to the estimation of latent-variable models, where powerful tools such as variational inference (VI) exist. 
However, in contrast to standard latent-variable models, parameter estimation with incomplete data often requires estimating exponentially-many conditional distributions of the missing variables, hence making standard VI methods intractable.
We address this gap by introducing variational Gibbs inference (VGI), a new general-purpose method to estimate the parameters of statistical models from incomplete data.
We validate VGI on a set of synthetic and real-world estimation tasks, estimating important machine learning models such as variational autoencoders and normalising flows from incomplete data.
The proposed method, whilst general-purpose, achieves competitive or better performance than existing model-specific estimation methods.
\end{abstract}


\section{Introduction}
\label{sec:intro}
\revisiontwo{This paper introduces a new general-purpose method to estimate statistical models from incomplete data that is well-suited for modern (deep) statistical models. Estimating statistical models is one of the core tasks in machine learning because the fitted model} can be used in many practical downstream tasks, such as, classification, prediction, anomaly detection, data augmentation, and missing data imputation \citep[\eg][Chapter~5.1.1]{Goodfellow2016}.
However, most of the current methods require large amounts of fully-observed data at training time, and hence they remain largely impractical in many real-world domains that are overwhelmed with incomplete data.
For example, the vast amounts of data gathered by online systems is sparse, with ratings data used in recommender systems often missing 95-99\% of the total data \citep{Marlin2011}. Similarly, a review of medical trial studies has identified that 95\% of studies contained missing data, with as much as 70\% of the data values missing in some studies \citep{bellHandlingMissingData2014}. 
This prevalence of missing data in real-world scenarios warrants a need for principled approaches to efficiently handle missing data in machine learning.%

\revisiontwo{A principled classical approach to estimating statistical models from incomplete data is expectation-maximisation \citep[EM,][]{Dempster1977} which aims at maximising the likelihood, however, it is mostly limited to simple models.}
Monte Carlo EM \citep[MCEM,][]{Wei1990} is a less limited version of classical EM that can be understood as an iterative method that fits the statistical model on imputations derived from itself. \revisiontwo{However, exact conditional sampling for modern statistical models is typically impossible.}
One way to \revisiontwo{(approximately)} sample imputations from a joint statistical model is via Markov chain Monte Carlo methods \citep[MCMC, \eg][Chapter~27.4]{Barber2017}, however they tend to be computationally intensive and hence scale poorly to larger data sets \citep{Blei2017}. 
\revisiontwo{Variational inference \citep[VI,][]{Jordan1999} is often a computationally more performant alternative, while sometimes lacking the asymptotic exactness guarantees of MCMC.
In case of missing data, however, as we will elaborate in the paper (Sections~\ref{sec:background-density-estimation-and-vi}\nobreakdash{-}\ref{sec:background-vi}), 
existing (amortised) VI methods would require $2^d-1$ variational distributions, one for each non-trivial pattern of missingness, and thus scale poorly with the dimension of the data $d$. Their applicability to estimating statistical models from incomplete data has thus been strongly limited and our paper addresses this gap in the literature.}

\subsection{Main Contributions}
\revisiontwo{
Our main contribution is a novel general-purpose method for estimating statistical models from incomplete data. The method combines the computational performance of VI and the expressiveness of Markov chains, which makes it well-suited for modern (deep) statistical models.
Crucially, the method only requires $d$ rather than $2^d-1$ variational distributions and thereby overcomes the limitations of existing (amortised) VI methods that prevented their use for model estimation from incomplete data. We achieve this reduction from exponential to linear growth by leveraging techniques that are related to those used by popular imputation methods (see Section~\ref{sec:background-fcs}). As the proposed method is based on variational inference (VI) and the Gibbs sampler, we call it variational Gibbs inference (VGI).
}

\begin{figure}[p]
    \centering
    \vspace{-2em}
    \includegraphics[width=0.99\linewidth]{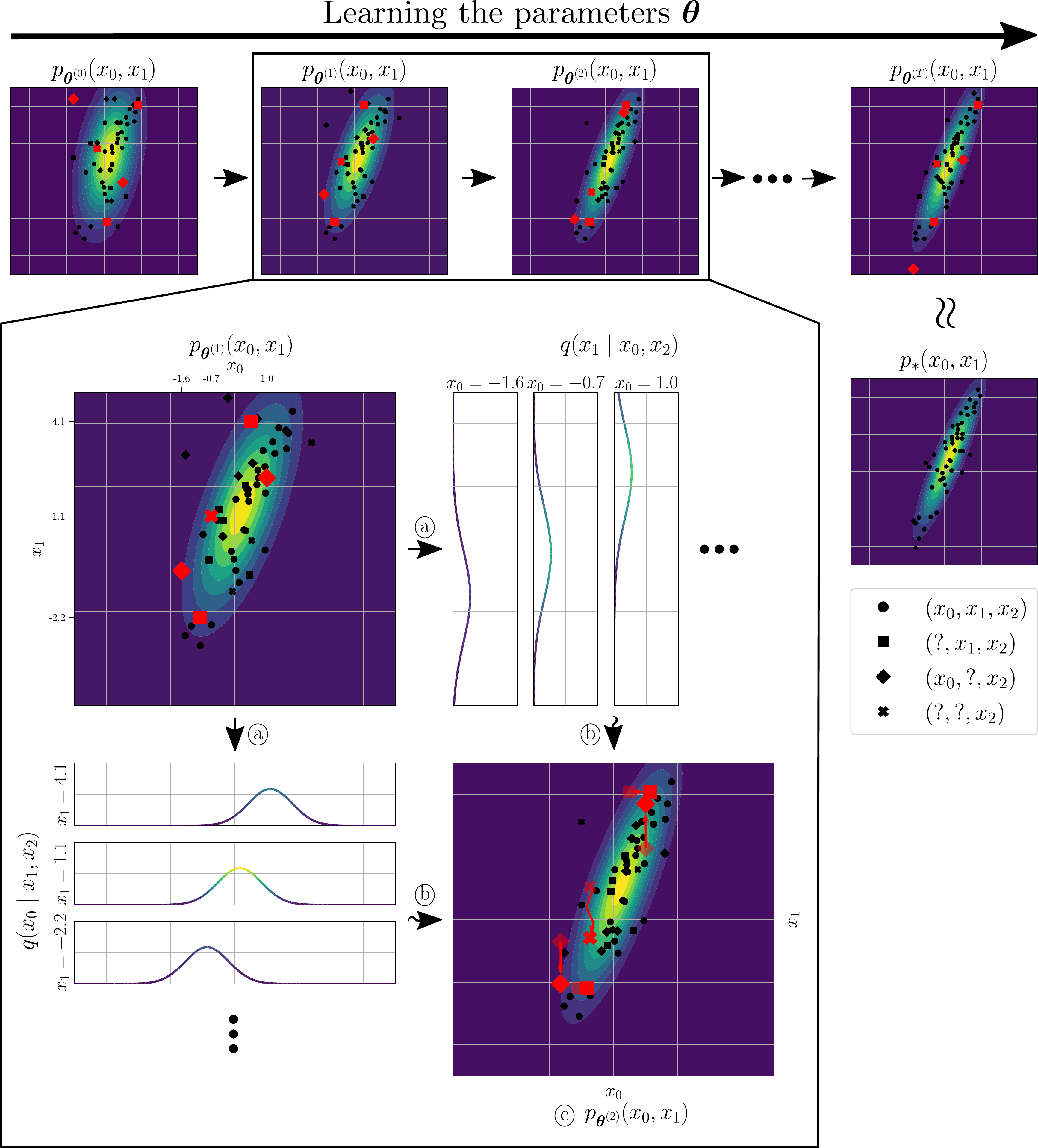}
    \caption[A schematic illustration of the VGI method on a toy model.]{A schematic illustration of our method on a 3D toy model $\pt(x_1, x_2, x_3)$, with missingness only in the first two dimensions for ease of exposition.\\
    The top row along the arrow shows contours of the model $p_{\thetab^{(t)}}$ with parameters $\thetab^{(t)}$ learnt iteratively as it approaches the complete-data estimate $p_*$.\\
    The zoomed-in pane details an iteration of the algorithm, where:\\
    \textcircled{a} univariate variational conditionals $q(x_j \mid \xnj)$ are learnt from $\pt(\x)$ and\\
    \textcircled{b} used %
    to update the imputations via pseudo-Gibbs sampling; \\
    \textcircled{c} $\pt(\x)$ is \revisiontwo{learnt} using the imputed data \revisiontwo{via} a variational objective.\\
    Compared to standard VI, our approach reduces the number of variational distributions that need to be learnt from $2^{d} - 1$  to $d$.
    }
    \label{fig:illustration}
\end{figure}

\revisiontwo{Figure~\ref{fig:illustration} illustrates VGI.}\footnote{\revisiontwo{An interactive demo is available at \href{https://github.com/vsimkus/variational-gibbs-inference\#vgi-demo}{github.com/vsimkus/variational-gibbs-inference}.}}
The method starts with $K$ initial random imputations of the missing values of each incomplete data-point.
Each iteration of our algorithm has two steps---a learning and an imputation step. 
In the learning step, the statistical model and the variational \revisiontwo{distributions} are updated by maximising a variational lower-bound on the log-likelihood ({\footnotesize \textcircled{a}} and {\footnotesize \textcircled{c}} in the figure). 
In the imputation step, the missing values are imputed via pseudo-Gibbs sampling using the learnt variational \revisiontwo{distributions} ({\footnotesize\textcircled{b}} in the figure). 
These imputations are \emph{persistent} and updated in subsequent iterations.
In this way, the initial imputations iteratively adapt to the target statistical model, such that they follow the joint distribution of the missing variables conditional on the observed data.
Moreover, our method facilitates parameter sharing across missingness patterns by using an amortised inference model and is amenable to parallelisation, which further increases computational efficiency.

\subsection{Overview of the Paper}

The paper is organised as follows. 
Section~\ref{sec:background} provides background on statistical model estimation and standard variational inference for incomplete data, discusses the technical gap in the literature on amortised variational inference of missing data, and describes a classical approach for missing data imputation that is related to our work.

In Section~\ref{sec:vgi}, we derive the VGI optimisation objective (Section~\ref{sec:var-objective}), present the VGI algorithm (Section~\ref{sec:vgi-algorithm}), and discuss practical considerations for modelling the variational Gibbs conditionals (Section~\ref{sec:variational-model}). 
We further introduce a block-Gibbs version of VGI (Section~\ref{sec:vbgi}), which we later use \revisiontwo{in Section~\ref{sec:vae-experimental-settings}} to adapt our method specifically to the variational autoencoder \citep[VAE,][]{Kingma2013,Rezende2014a}. 
The details of evaluating models fitted with VGI for model selection using incomplete held-out data are presented next (Section~\ref{sec:vgi-eval-held-out}).
A discussion in Section~\ref{sec:vgi-related-methods} on the similarities and differences between VGI and related methods closes Section~\ref{sec:vgi}.

In Section~\ref{sec:toy-experiments}, we validate and analyse our method on low- and high-dimensional toy problems where analytical solutions exist. 
In Section~\ref{sec:vae-experiments} we compare VGI against model-specific estimation methods on a VAE model and show that VGI produces competitive results in terms of model accuracy.
In Section~\ref{sec:flow-experiments} we apply our general-purpose method to normalising flow estimation \citep{Rezende2015} and show that it can outperform a flow-specific estimation method.

In Section~\ref{sec:discussion} we summarise our findings and discuss possible future research directions.

\section{Background}
\label{sec:background}
In this section we provide the background on statistical model estimation and variational inference with missing data,
amortised variational inference, and a popular missing data imputation method based on conditional modelling. We highlight the shortcomings of those methodologies that the proposed method addresses.

\subsection{Model Estimation and Variational Inference with Incomplete Data}
\label{sec:background-density-estimation-and-vi}

Statistical model estimation from observed data is typically solved via maximum-likelihood estimation (MLE). Following \citet{Little2002} the marginal observed likelihood for incomplete data is defined via a joint model $\pt(\xo, \m)$ of the observed variables of interest $\xo$ and the binary missingness mask $\m$,
\begin{align}
    \pt(\xo, \m) = \int \pt(\xo, \xm) p(\m \mid \xo, \xm) \dif \xm,
    \label{eq:incomplete-joint-model}
\end{align}
where the complete variable of interest $\x$ is defined by its observed and missing components $\x = (\xo, \xm)$, $\pt(\xo, \xm)$ is the statistical model of the data, and $p(\m \mid \xo, \xm)$ is the missingness model representing the missing data mechanism. 
In this paper we assume that data are missing at random (MAR) or missing completely at random (MCAR) so that the missingness model can be ignored when estimating the statistical model of the data $\pt(\x)$ \citep{Rubin1976}. The maximum likelihood estimate of the parameters $\thetab$ is then given by
\begin{align}
\argmax_\thetab \prod_{i=1}^N \pt(\xo^i, \m^i) = \argmax_\thetab \prod_{i=1}^N \int \pt(\xo^i, \xm^i) \dif \xm^i, \label{eq:incomplete-mar-mle}
\end{align}
where $N$ is the number of data-points and $i$ denotes the index of a data-point.
A brief review of the different missingness mechanisms and a proof of the above is provided in Appendix~\ref{apx:ignorable-missingness}.

The integral over the missing \revisiontwo{components} in \eqref{eq:incomplete-mar-mle} is typically intractable, which renders the likelihood and hence standard maximum-likelihood estimation intractable. 
Exactly the same problem occurs when estimating latent-variable models. 
Indeed, we can consider the missing \revisiontwo{components} to be latent variables and hence obtain a tractable lower bound on the log-likelihood as done in variational inference \citep[VI,][]{Jordan1999}. Following the standard derivation of the evidence lower-bound (ELBO) \citep[\eg][Chapter~11.2]{Barber2017}, the bound on the log-likelihood for $N$ incomplete data points is
\begin{align*}
    \frac{1}{N} \sum_{i=1}^N \log \pt(\xo^i) &\geq 
        \frac{1}{N} \sum_{i=1}^N \E_{q(\xm^i \mid \xo^i)} \left[ \log \frac{\pt (\xo^i, \xm^i)}{q(\xm^i \mid \xo^i)} \right], %
\end{align*}
where %
$q(\xm^i \mid \xo^i)$ are the variational distributions. 
\revisiontwo{Maximising the ELBO with respect to the model parameters $\thetab$ and variational distributions $q \in \mathcal{Q}$ in some distributional family $\mathcal{Q}$ yields an approximate MLE solution to \eqref{eq:incomplete-mar-mle}.
If the variational distributions are equal to the model conditionals $\pt(\xm^i \mid \xo^i)$ for all $\xo^i$ then the bound is tight and the solution can be made exact.\footnote{\revisiontwo{Note that to be able to satisfy this condition the variational family should include the model conditionals $\pt(\xm^i \mid \xo^i) \in \mathcal{Q}$.
Still, such specification may not guarantee the exactness of the MLE solution, since due to local optima in the ELBO the variational distributions may not perfectly match the model conditionals.}}}

\begin{figure}[t]
\centering
    \aboverulesep=0ex
    \belowrulesep=0ex
    \renewcommand{\arraystretch}{1.5}
    \begin{tabular}{c|c|c|c|c|}
        \mc{}       & \mc{}       & \mc{}       & \mc{}       & \mc{}        \\ \cmidrule{2-5}
        $\x^{1}$  & $x_1^{1}$  & $?$  & $x_3^{1}$  & $x_4^{1}$ \\ \cmidrule{2-5}
        $\x^{2}$  & $?$  & $x_2^{2}$  & $x_3^{2}$  & $?$ \\ \cmidrule{2-5}
        $\x^{3}$  & $?$  & $?$  & $?$  & $x_4^{3}$ \\ \cmidrule{2-5}
        \vdots      & \multicolumn{4}{|c|}{\vdots} \\ \cmidrule{2-5}
    \end{tabular}
    \hspace{1em}
    \begin{tabular}{c|c|c|c|c|}
        \mc{}       & \mc{}       & \mc{}       & \mc{}       & \mc{}        \\ \cmidrule{2-5}
        $\m^{1}$  & 1  & 0  & 1  & 1 \\ \cmidrule{2-5}
        $\m^{2}$  & 0  & 1  & 1  & 0 \\ \cmidrule{2-5}
        $\m^{3}$  & 0  & 0  & 0  & 1 \\ \cmidrule{2-5}
        \vdots      & \multicolumn{4}{|c|}{\vdots} \\ \cmidrule{2-5}
    \end{tabular}
    \hspace{1em}
    \begin{tabular}{c}
    $q(\xm^i \mid \xo^i)$\\ \midrule
    $q(x^{1}_2 \mid x^{1}_1, x^{1}_3, x^{1}_4)$\\
    $q(x^{2}_1, x^{2}_4 \mid x^{2}_2, x^{2}_3)$\\
    $q(x^{3}_1, x^{3}_2, x^{3}_3 \mid x^{3}_4)$\\
    \vdots
    \end{tabular}
    \caption{Left: Observed incomplete data. Middle: missingness mask. There are potentially as many as $2^d$ different missingness patterns. Right: corresponding variational posterior distributions.}
    \label{fig:incomplete-data-example}
\end{figure}

Whilst it is straightforward to obtain a tractable lower bound on the log-likelihood, there is a crucial complication in the missing data problem that sets it apart from standard (amortised) VI problems: for $d$-dimensional data, there is not one but possibly $2^d - 1$ such variational conditional distributions, namely one for each non-trivial pattern of missingness, as illustrated in Figure~\ref{fig:incomplete-data-example}.
This raises fundamental technical issues that have prevented the application of VI to parameter estimation from incomplete data. We discuss these complications in the next subsection.

\subsection{The Difficulty of Amortising Missing Variable Inference}
\label{sec:background-vi}

Classical variational inference requires the specification of one variational distribution $q$ per observed data-point, but such an approach is computationally inefficient in modern large-scale use-cases due to a lack of parameter sharing.
An amortised version of VI \citep{Gershman2014} deals with this issue by incorporating global parameter sharing. 
The key idea in amortised VI is to parametrise the conditional variational distribution by a deterministic function, or an inference network, of the observed inputs with globally shared parameters $\phib$.
However, in the missing data setting we want to represent all $2^d-1$ conditional distributions caused by the different missingness patterns. 
\revisiontwo{The exponential growth in the number of missingness patterns means that the na\"ive approach of using one inference network per pattern results in a lack of parameter sharing across data-points even for moderate-dimensional data, thus cancelling the computational advantages of amortised VI}.

\revisiontwo{
Efficiently amortising a variational distribution $\qp(\xm \mid \xo)$ for any $(\xm, \xo) \in \x$ entails simultaneously dealing with two problems: (i) handling all possible combinations of variables in the conditioning set, that is, for all $\xo \in \x \setminus \xm$, and (ii) constructing pdfs/pmfs for arbitrary sets of target variables $\xm \in \x \setminus \xo$.
Existing work has focused on the first problem, approaching it by either simply fixing the input dimensionality of the inference network to $d$ and then padding the missing inputs with zeroes \citep{Nazabal2018, Mattei2019}, or using permutation-invariant network architectures \citep{Ma2019}. 
However, there is no work in the VI literature that addresses the second problem.\footnote{But see Section~\ref{sec:vgi-related-methods} for related work.} Dealing with this problem requires care---placing restrictions on the variational family may result in a biased estimate of the target statistical model (as shown below). Hence, to match the possibly complex conditional distributions of the target model we would like to use unrestricted probabilistic models for the variational family whilst still being able to take advantage of the increased efficiency of amortised VI.
}

\begin{figure}[tb]
    \centering
    \includegraphics[width=0.8\linewidth]{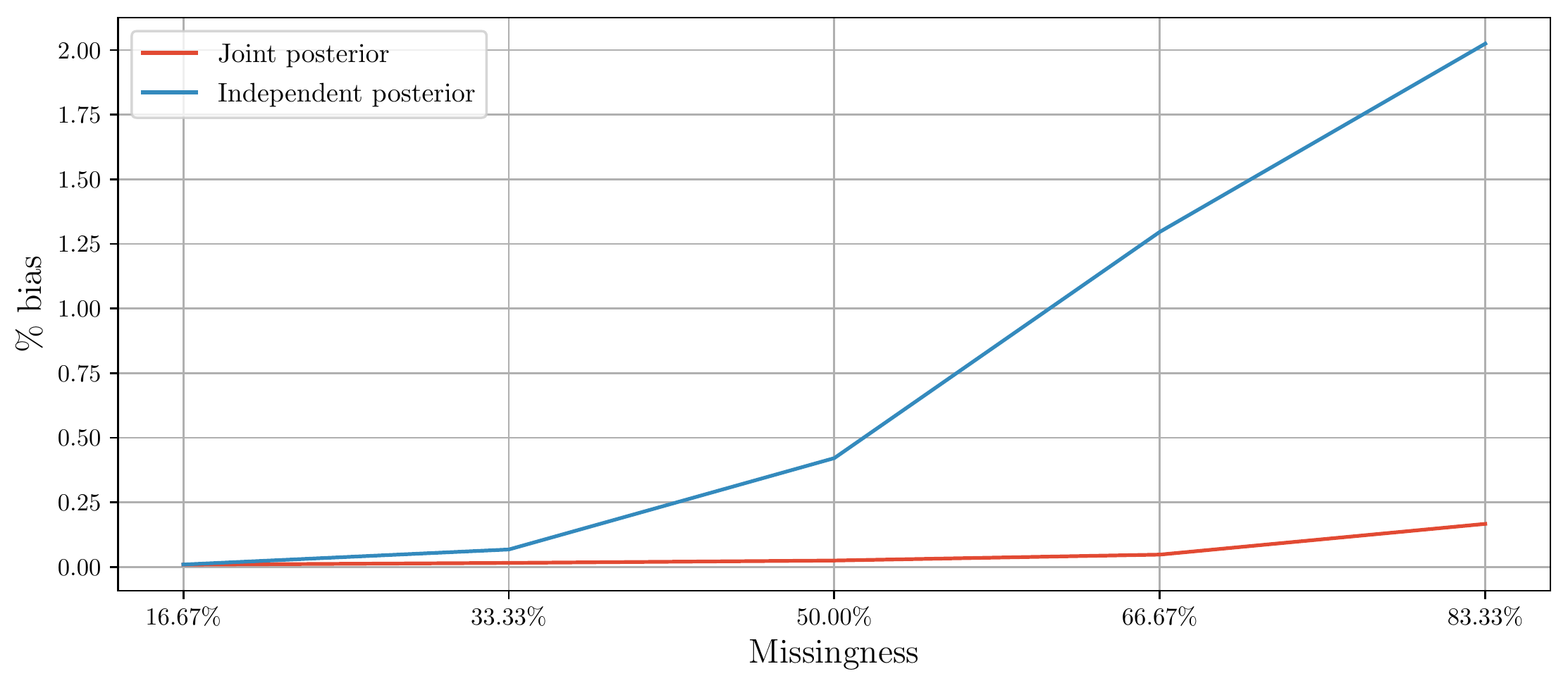}
    \caption{Percent bias measured on the test log-likelihood of a toy factor analysis model with correct \revisiontwo{(red)} and incorrect \revisiontwo{(blue)} modelling assumptions. \revisiontwo{Log-likelihood was computed on complete test data.} 
    We observe that with the incorrect independence assumption $\xj \indep \xmnj \mid \xo$ for $\forall j \in \idx{\m}$ the bias is significantly larger compared to the correctly-specified joint model. Percent bias is computed as $100 \cdot \abs{(\E \left[ \ell(\pt) \right] - \ell(\ps) )/ \ell(\ps)}$ \citep[\eg][Chapter~2.5]{VanBuuren2018}, where $\ell(\pt)$ and $\ell(\ps)$ are the average log-likelihood on test data using the fitted $\pt$ and the ground truth $\ps$ models\revisiontwo{, and the expectation is over different runs of the learning algorithm.}}
    \label{fig:toy-em-bias}
\end{figure}

\revisiontwo{We illustrate how restricting the variational family can reduce the quality of the fitted target model. For example,} taking inspiration from mean-field VI \citep[\eg][Section~10.1.1]{Bishop2006} one may assume independence of the missing variables given the observed and work with variational distributions of the form
\begin{align*}
    \qp(\xm^i \mid \xo^i) = \prod_{j \in \idx{\m^i}} \qp(\xj^i \mid \xo^i),
\end{align*}
where $\idx{\m^i}$ denotes the set of indices of the missing values in the data-point. 
However, in Figure~\ref{fig:toy-em-bias} we show that such independence assumption can significantly bias the learnt probabilistic model by artificially reducing correlations in the completed data, with the bias increasing with the fraction of missingness.

In this work, we introduce a novel variational method that can fit statistical models using only $d$ variational conditionals, thus facilitating parameter sharing without introducing strong statistical assumptions \revisiontwo{or restricting their modelling capability}.

\subsection{Imputing Missing Data with Fully-Conditional Models}
\label{sec:background-fcs}

For multivariate missing data, imputations can be generated by iteratively sampling univariate conditional distributions $f(\xj \mid \xnj; \phibj)$ for $\forall j \in \idx{\m}$ with local parameters $\phibj$. 
\revisiontwo{Imputation methods in this family, which learn the conditionals and impute the data iteratively,} are known as fully-conditional specification (FCS) or sequential iterative regression methods \citep{Brand1999, Rubin2003, VanBuuren2006}. 
They have been essential tools in statistical analysis of incomplete data for over two decades.

The most popular of the FCS imputation methods is multivariate imputation by chained equations \citep[MICE,][]{VanBuuren2000}. %
MICE is an iterative framework that starts with a random imputation from the observed data and then sequentially imputes each incomplete variable one-by-one. 
Each step consists of fitting a conditional distribution $f(\xj \mid \xnj; \phibj)$ on the data-points where $\xj$ is observed \revisiontwo{via regression} and then imputing the $\xj$ in data-points where it is not observed \revisiontwo{by sampling the learnt conditional} (see Appendix~\ref{apx:mice} for more details about MICE).
\revisiontwo{To capture the uncertainty of the missing variables, the MICE procedure is usually independently} repeated $K$ number of times to obtain $K$ imputations of the missing data, an approach known as multiple-imputation \citep[MI,][]{Rubin1987}.
\revisiontwo{
After imputation, statistical models $\pt(\x)$ can be fit on the imputed data sets using standard methods for complete data.\footnote{However, the standard multiple-imputation workflow for parameter estimation is generally not applicable to statistical models when their parameters are not identifiable, as is most often the case in deep generative models, and hence caution must be taken (see Appendix~\ref{apx:combining-mi}).} We provide more background on such an ``impute-then-fit'' approach in Appendix~\ref{apx:combining-mi}.
}

\revisiontwo{
Whilst the sampling procedure bears some similarity to Gibbs sampling \citep{Geman1984}, see Appendix~\ref{apx:gibbs-sampling} for a short review, an important theoretical difference between Gibbs sampling and FCS is that in standard Gibbs sampling one starts with a joint model of the target distribution and then, by decomposing it into full-conditionals, samples a Markov chain that will eventually converge to the joint target distribution.
On the other hand, the univariate distributions with disjoint parameters $\phibj$ in FCS are not guaranteed to have a joint distribution \citep[][Section~1.6]{Arnold1999}, which is why this approach has been called pseudo-Gibbs sampling \citep{Heckerman2000}.\footnote{Sometimes also called incompatible Gibbs sampling or compound conditional specification \citep{Rubin2003}.}
Nevertheless, despite the possibly incompatible univariate distributions, it has been empirically found that the procedure can generate good imputations of multivariate missing values in many practical settings \citep{Rubin2003, VanBuuren2006}, which motivates us to use a similar factorisation to represent the variational distributions of the missing variables.
}

FCS has several desirable properties: (i) it easily lends itself to the specification of flexible imputation models since even for univariate Gaussian conditionals the joint distribution can be complex and multi-modal \citep[\eg][Section~3.4]{Arnold1999}, 
(ii) heterogenous data (mixed continuous and discrete) can be handled by using different distribution families for each conditional,
(iii) it alleviates the problem of having to learn  $2^d - 1$ joint missing variable distributions into one that requires learning only $d$ univariate Gibbs conditionals, 
and (iv) univariate conditional distributions can be easily constrained to prevent invalid imputation values.

\revisiontwo{
Inspired by the empirical success of FCS methods on multiple-imputation tasks, in the next section we propose a variational approach that characterises the joint variational distributions $\qp(\xm \mid \xo)$ via $d$ variational full-conditional models. 
Thus, the joint variational distribution $\qp(\xm \mid \xo)$ can be made very flexible since we neither require strong statistical independence assumptions nor restrictive distributional assumptions on the variational families (see Section~\ref{sec:background-vi}),
which enables us to well optimise the ELBO and hence obtain a good (approximate) MLE estimate of the statistical model $\pt(\x)$.
In this way, we generalise variational inference of univariate missing data to the multivariate case, akin to how FCS generalises univariate regression-based imputation to the multivariate setting \citep[\eg][Chapter~4.5.1]{VanBuuren2018}.
}

\section{Variational Gibbs Inference}
\label{sec:vgi}

We present variational Gibbs inference (VGI) to estimate the parameters of statistical models from incomplete data by maximising a variational lower-bound on the log-likelihood.
The method requires only $d$ variational conditionals and uses an iterative algorithm that alternates between two steps: (i) the learning step which fits the statistical model and the variational conditionals by optimising the variational objective, and (ii) the imputation-step which updates persistent pseudo-Gibbs chains that provide the imputations for each incomplete data-point.
The following sections introduce the variational objective (Section~\ref{sec:var-objective}) and the VGI algorithm (Section~\ref{sec:vgi-algorithm}), discuss practicalities to consider when modelling the variational conditionals (Section~\ref{sec:variational-model}), explain how VGI can be adapted to specific statistical models, such as latent-variable models (Section~\ref{sec:vbgi})%
, describe the details of evaluating the method on incomplete held-out data (Section~\ref{sec:vgi-eval-held-out}),
and finally discusses the related work (Section~\ref{sec:vgi-related-methods}). %

\subsection{The Variational Objective}
\label{sec:var-objective}

We derive a variational ELBO to estimate the statistical model $\pt(\x)$ from incomplete data using the marginal distributions $f_\phib^t(\xm \mid \xo)$ of Markov chains with learnable parameters $\phib$ as the variational distribution of the missing variables. %
\revisiontwo{Maximising the objective allows us} to learn the parameters $\thetab$ of the model $\pt(\x)$ and the parameters $\phib$ of the Gibbs transition \emph{kernel} $\tau_\phib(\xm \mid \xo, \txm)$,\footnote{We use $\txm$ to denote the previous imputation value before the transition.} or equivalently, the parameters of the $d$ univariate Gibbs conditionals (\emph{variational conditionals}) $q_{\phibj}(\xj \mid \txmnj, \xo)$ which characterise the kernel. 
\revisiontwo{By learning the kernel $\tau_\phib$ we can match the marginal distributions of the imputation Markov chains $f_\phib^t$ to the conditional distributions of the target model $\pt(\xm \mid \xo)$.}
Consequently, the variational conditionals $q_{\phibj}(\xj \mid \xmnj, \xo)$ are learnt to approximate the true conditionals $\pt(\xj \mid \xmnj, \xo)$.

Representing the variational distribution implicitly via the samples of a Markov chain allows the VGI method to work with a flexible variational model for adaptive imputations, which in turn enables tightening of the variational lower-bound and thus the maximisation of the likelihood. Moreover, by representing the kernel with $d$ variational conditionals, VGI achieves efficient amortisation of the exponentially-many distributions of missing data, similar to fully-conditional imputation methods discussed in Section~\ref{sec:background-fcs}. 
A summary of the key terms in VGI is provided in Table~\ref{tab:term-glossary}.

\begin{table}[bt]
\centering
    \begin{tabular}{lll}
        \textbf{Term} & \textbf{Reference} & \textbf{Description} \\\midrule
        $\pt(\x)$ & Sec.~\ref{sec:background-density-estimation-and-vi} & Statistical model of the data \\
        $\qpj(\xj \mid \xmnj, \xo)$ & Sec.~\ref{sec:variational-model} & Variational conditionals \\
        $\tau_\phib(\xm \mid \xo, \txm)$ & Eq.~\eqref{eq:gibbs-transition} & Gibbs transition kernel \\
        $\tilde \tau_\phib(\xm \mid \xo, \txm)$ & Eq.~\eqref{eq:extended-gibbs-transition} & Extended(-Gibbs) transition kernel (see Appendix~\ref{apx:vgi-derivation-extended}) \\
        $f^{t}_\phib(\xm \mid \xo)$ & Eq. \eqref{eq:marginal-mc-distribution} & Marginal distribution of a Markov chain at step $t$\\
        $\JVGI^t(\thetab, \phib; \xo)$ & Eq. \eqref{eq:vgi-objective} & The VGI objective for a single data-point $\xo$ at step $t$
    \end{tabular}
    \caption{A glossary of key terms used in VGI.}
    \label{tab:term-glossary}
\end{table}

We start our derivation with the standard variational ELBO, but with the marginal distribution $f_\phib^t$ of a Markov chain as the variational distribution
\begin{align}
    \log \pt(\xo) &\geq  \Lc(\thetab, f_\phib^t; \xo),&  \Lc(\thetab, f_\phib^t; \xo)&= \E_{f_\phib^t(\xm \mid \xo)}\left[ \log \frac{\pt(\xm, \xo)}{f_\phib^t(\xm \mid \xo)} \right],\label{eq:elbo-marginal-markov-chain} %
\end{align}
where $t$ denotes the iteration of our algorithm.
To maximise the likelihood of the model $\pt$ we want the bound to be sufficiently tight, and hence $f_\phib^t$ needs be flexible and not biased by the choice of the initial imputation distribution $f^0$. 
The flexibility of the marginal distribution of a Markov chain depends on the kernel, and the bias induced by the choice of $f^0$ decreases with the number of steps $t$ in the Markov chain \citep[\eg][Chapter~4.4]{Cover2006}. Hence we want $\tau_\phib$ to be flexible and $t$ to become sufficiently large.

However, \revisiontwo{we cannot evaluate the above lower-bound} since the imputation distribution $f_\phib^t(\xm \mid \xo)$ is implicitly-defined by the Markov chain. 
Moreover, maximising the bound with respect to the parameters $\thetab$ and $\phib$ with gradient-based methods would be expensive for large $t$ and might suffer from gradient instability.
Rather than estimating the gradients of the above lower-bound with respect to $\phib$ over the full length of the Markov chain, we propose ``cutting'' the chain just before the current transition and optimising \revisiontwo{the transition kernel} $\tau_\phib$ greedily as we sample.
The imputations obtained in iteration $t$ will then be re-used in the next iteration to further improve $\tau_\phib$ and $\pt$.
This allows us to derive a tractable and efficient variational method.

We choose the kernel, which transforms the imputations from the previous iteration, $\txm$, and produces updated imputations $\xm$, to be Gibbs (hence the name of the method) and define it as follows
\begin{align}
    \tau_\phib(\xm \mid \xo, \txm) &= \sum_{j \in \idx{\m}} \pib(j) \qpj(\xj \mid \xmnj, \xo) \delta (\xmnj - \txmnj), \label{eq:gibbs-transition}
\end{align}
where the dimension index $j \in \idx{\m}$ is controlled by a fixed uniform selection distribution $\pib(j)$ of a random-scan Gibbs sampler, $\xj$ and $\xmnj$ denote the $j$-th missing dimension and the remaining missing dimensions of $\xm$ respectively, and the kernel is specified by the $d$ variational conditionals $\qpj(\xj \mid \xmnj, \xo)$.
Alternatively, we can also let the updated imputation $\xj$ depend on the previous imputation value $\txj$, which gives variational conditionals of the form $\qpj(\xj^t \mid \xo, \txj^{t-1}, \txmnj^{t-1})$. We call a kernel $\tilde \tau_\phib$ (or subsequently the variational model) that uses this form of conditionals an \emph{extended-Gibbs} kernel (or, for conciseness, \emph{extended} kernel) and provide an analogous derivation of this section in Appendix~\ref{apx:vgi-derivation-extended}.

Marginalising out $\txm$ in \eqref{eq:gibbs-transition} with respect to the \emph{imputation distribution} $f^{t-1}$ from the previous step in the Markov chain gives the imputation distribution after a single Gibbs update
\begin{align}
    f_\phib^t(\xm \mid \xo) &= \int \tau_\phib(\xm \mid \xo, \txm) f^{t-1}(\txm \mid \xo) \dif \txm \label{eq:marginal-mc-distribution}\\
    &= \sum_{j \in \idx{\m}} \pib(j) f_\phib^t(\xm \mid \xo, j) %
    \nonumber\\
    & = \E_{ \pib(j)} \left[ f_\phib^t(\xm \mid \xo, j) \right], \label{eq:joint-var-distribution}
\end{align}
where $f_\phib^t(\xm \mid \xo, j)$ is the imputation distribution $f^{t-1}$ updated in dimension $j$,
 \begin{align}  
    f_\phib^t(\xm \mid \xo, j) &= \qpj(\xj \mid \xmnj, \xo) f^{t-1}(\xmnj \mid \xo). \label{eq:joint-var-distribution-j}
\end{align}
 Note that the absence of $\phib$ in $f^{t-1}$ corresponds to the aforementioned ``cutting'' of the chain.

Now, we continue with the standard ELBO from \eqref{eq:elbo-marginal-markov-chain} and use \eqref{eq:joint-var-distribution} and \eqref{eq:joint-var-distribution-j} to derive the variational Gibbs ELBO at iteration $t$ 
\begin{align}
    \Lc^{t}(\thetab, \phib; \xo)
    &\oset[1.2ex]{\eqref{eq:joint-var-distribution}}{=} \E_{ \pib(j) f_\phib^t(\xm \mid \xo, j)} \left[ \log \frac{\pt(\xo, \xm)}{f_\phib^t(\xm \mid \xo)} \right] \nonumber\\
    &=\E_{ \pib(j) f_\phib^t(\xm \mid \xo, j)} \left[ \log \frac{\pt(\xo, \xm)}{f_\phib^t(\xm \mid \xo, j)}+\log \frac{f_\phib^t(\xm \mid \xo, j)}{f_\phib^t(\xm \mid \xo)} \right] \nonumber\\
    &= \E_{ \pib(j) f_\phib^t(\xm \mid \xo, j)} \left[ \log \frac{\pt(\xo, \xm)}{f_\phib^t(\xm \mid \xo, j)} \right] \nonumber\\
    &\phantom{==} + \E_{\pib(j)} \KLD{f_\phib^t(\xm \mid \xo, j)}{f_\phib^t(\xm \mid \xo)} \nonumber\\
    &\geq \E_{ \pib(j) f_\phib^t(\xm \mid \xo, j)} \left[ \log \frac{\pt(\xo, \xm)}{f_\phib^t(\xm \mid \xo, j)} \right] \nonumber\\
    &\,\begin{aligned}%
    \oset[1.2ex]{\eqref{eq:joint-var-distribution-j}}{=}& \,\E_{ \pib(j) f^{t-1}(\xmnj \mid \xo) \qpj(\xj \mid \xmnj, \xo)} \left[ \log \frac{\pt(\xo, \xm)}{\qpj(\xj \mid \xmnj, \xo)} \right]\\
    &- \E_{{\pib(j)} f^{t-1}(\xmnj \mid \xo)} \left[ \log f^{t-1}(\xmnj \mid \xo) \right],
    \end{aligned} \label{eq:variational-Gibbs-elbo}
\end{align}
where the inequality follows from the non-negativity of KL divergence.
If $\xm$ is independent of $j$, which holds for the stationary distribution of the Markov chain characterised by the variational conditionals (see Appendix~\ref{apx:gibbs-kernel-stationary-distribution}), then the KL divergence term is zero and maximising \eqref{eq:variational-Gibbs-elbo} is equivalent to maximising \eqref{eq:elbo-marginal-markov-chain}.

\revisiontwo{The ELBO in \eqref{eq:variational-Gibbs-elbo} can be optimised efficiently with respect to $\thetab$ and $\phib$ since only samples from the imputation distribution $f^{t-1}(\xm \mid \xo)$ are needed and the intractable entropy term in the last line of the equation does not depend on the parameters $\thetab$ or $\phib$, and hence does not need to be computed.}
Removing the entropy term, we obtain the variational Gibbs inference (VGI) objective $\JVGI^t$ at iteration $t$ for one incomplete data-point $\xo$:\footnote{The objective using the extended kernel (Appendix~\ref{apx:vgi-derivation-extended}) is almost the same, but the variational conditional additionally includes a dependency on $\txj$ and hence depends on all of the imputed variables from the previous iteration $(\txj, \xmnj) \sim f^{t-1}(\xm \mid \xo)$.}
\begin{align}
   \hspace{-2.2ex} \JVGI^t(\thetab, \phib; \xo) &= \E_{{\pib(j)} f^{t-1}(\xmnj \mid \xo) \qpj(\xj \mid \xmnj, \xo) } \left[ \log \frac{\pt(\xj, \xmnj, \xo)}{\qpj(\xj \mid \xmnj, \xo)} \right] \label{eq:vgi-objective}
\end{align}
\revisiontwo{%
Importantly, while cutting the chain removes the entropy of $f^{t-1}$ from the objective function, the entropy of the variational conditionals remains. This prevents the collapse of the variational conditionals to point masses and consequently also prevents the collapse of the joint imputation distribution $f^{t}$
that is sampled using the fitted conditionals.%
\footnote{\revisiontwo{Note that the imputation distribution $f^{t-1}(\xmnj \mid \xo)$ is kept fixed when we maximise the objective $\JVGI^t(\thetab, \phib; \xo)$ with respect to $\thetab$ and $\phib$, due to the ``cutting'' of the Markov chain. After updating the parameters, we update $f^{t-1}(\xmnj \mid \xo)$ via the updated transition kernel according to \eqref{eq:marginal-mc-distribution}.}
}}

For $N$ incomplete data-points we will maximise the averaged objective to obtain the parameter estimates $\hat \thetab$ and $\hat \phib$ at iteration $t$
\begin{align}
    \hat \thetab^{t}, \hat \phib^{t} = \argmax_{\thetab, \phib} \frac{1}{N} \sum_{i=1}^N \JVGI^t(\thetab, \phib; \xo^i). \label{eq:vgi-objective-avg}
\end{align}
In practice, we optimise \eqref{eq:vgi-objective-avg} using stochastic gradient ascent and approximate the expectations in $\JVGI^t$ with Monte Carlo integration
\begin{gather}
    \hJVGI^t(\thetab, \phib; \xo) = \frac{1}{K} \sum_{k=1}^K \frac{1}{M} \sum_{m=1}^M
    \left[ \log \frac{\pt(x_{j^m}^{k}, \x_{\mis \smallsetminus j^{m}}^{k}, \xo)}{\qpj(x_{j^m}^{k} \mid \x_{\mis \smallsetminus j^m}^{k}, \xo)} \right], \label{eq:vgi-objective-approx}%
\end{gather}
where $x_{j^m}^k \sim \qpj(x_{j^m}^{k} \mid \x_{\mis \smallsetminus j^m}^{k}, \xo)$, $j^m \sim \pib(j)$, $\x_{\mis \smallsetminus j^m}^k \sim f^{t-1}(\x_{\mis \smallsetminus j^m} \mid \xo)$, and $K$ is the number of imputations for each incomplete data-point, $M$ is the number of samples used to approximate the expectation with respect to $j$ (and $\xj$), %
and $f^{t-1}$ is represented via samples from a fixed number of Markov chains. We empirically found that using small $M$ and $K$ was sufficient in most of our experiments.\footnote{We found $K \in [5, 10]$ and $M \in [1, 10]$ sufficient in our experiments.}

To summarise, we have derived the key VGI objective $\JVGI^t$, which maximises a lower-bound on the log-likelihood using samples from a marginal distribution of a variational Markov chain at any iteration $t$. 
Importantly, the objective uses only $d$ variational conditionals to represent all $2^d-1$ possible conditional distributions of missing data.
We next describe the VGI algorithm which integrates the optimisation of the iteration-dependent variational objective $\JVGI^t$ and sampling from the Markov chains $f^t(\xm \mid \xo)$ into an iterative procedure.

\subsection{The VGI Algorithm}
\label{sec:vgi-algorithm}

The variational Gibbs inference (VGI) algorithm is shown in Algorithm~\ref{alg:VGI} with subroutines summarised in Algorithms~\ref{alg:var-warm-up}\nobreakdash{-}\ref{alg:pseudo-Gibbs}.\footnote{Note that in our experiments we use optimised implementation of the algorithms by parallelising the loops where possible. Our implementation is available at \url{https://github.com/vsimkus/variational-gibbs-inference}.} 
The core objective of the algorithm is to fit a statistical model $\pt$ on incomplete data set $\D$ using a variational model $\qp$ of the Gibbs conditionals that is learnt jointly with $\pt$.
The method uses stochastic gradient optimisation \citep[\eg][]{Spall2003, ruderOverviewGradientDescent2017} to learn the parameters $\thetab$ and $\phib$ efficiently by processing the data set in mini-batches, which are randomly chosen subsets of the full data set $\D$.
Algorithm~\ref{alg:VGI} consists of three stages: initialisation, model warm-up, and the main iterative stage, which we describe below.

\paragraphJMLR{Initialisation.} In line~\ref{alg:line:create-initial-imputations} the algorithm starts with an incomplete data set $\D = \{(\xo^i, \m^i)\}_i$ and produces a $K$-times imputed data set $\D_K = \{ (\xo^i, \m^i, \xm^{(i, 1)}, \ldots, \xm^{(i, K)})\}_i$, where each incomplete data-point is imputed using an initial imputation distribution $\xm^\sik \sim f^0(\xm \mid \xo^i)$.
The $f^0$ should be chosen such that it provides a good starting distribution for the Gibbs sampler. Empirically we found that choosing $f^0$ to be the marginal empirical distribution worked well, which corresponds to the maximum-entropy distribution of the missing data.

\begin{algorithm}[tpb]
    \caption{Variational Gibbs inference (VGI) algorithm}
    \label{alg:VGI}
    
    \textbf{Input:} $\pt(\x)$, statistical model with parameters $\thetab$\\
    \phantom{\textbf{Input:}} $\qpj(\xj \mid \xnj)$ for $j \in \{1 \ldots d\}$, variational conditional models with parameters $\phib$\\
    \phantom{\textbf{Input:}} $\D$, incomplete data set\\
    \phantom{\textbf{Input:}} $K$, number of imputations of each incomplete data-point\\
    \phantom{\textbf{Input:}} $f^0(\xm \mid \xo)$, initial imputation distribution\\
    \phantom{\textbf{Input:}} $\alpha_\thetab$ and $\alpha_\phib$, the parameter learning rates\\
    \phantom{\textbf{Input:}} \emph{max\_epochs}, number of epochs\\
    \textbf{Output:} $\thetab, \phib$, and $K$-times imputed data $\D_K$
    \begin{algorithmic}[1]
        \State \textbf{Create} $K$-times imputed data $\D_K$ using $f^0$ \label{alg:line:create-initial-imputations}
        \State \textbf{Warm up} the parameters $\phib$ of the variational model $\qp$ using Algorithm~\ref{alg:var-warm-up} \label{alg:line:var-warmup}
        \State \textbf{Warm up} the parameters $\thetab$ of the statistical model $\pt$ using Algorithm~\ref{alg:model-warm-up} \label{alg:line:density-warmup}
        \For{$t$ in $[1, \text{\emph{max\_epochs}}]$}
            \For{mini-batch $\B_K$ in $\D_K$}
                \State \textbf{Update} imputations in $\B_K$ using pseudo-Gibbs sampler (see Algorithm~\ref{alg:pseudo-Gibbs}) \label{alg:line:Gibbs-update}
                \State \textbf{Store} the updated imputations in $\B_K$ for use in the next epoch
                \State \textbf{Compute} $\hJVGI^t$ in \eqref{eq:vgi-objective-approx} using Algorithm~\ref{alg:vgi-objective}
                \State $\thetab^t = \thetab^{t-1} + \alpha_\thetab \nabla_\thetab \hJVGI^t$ \Comment{Update params of $\pt$ with a stochastic gradient step} \label{alg:line:parameter-update-theta}
                \State $\phib^t = \phib^{t-1} + \alpha_\phib \nabla_\phib \hJVGI^t$ \Comment{Update params of $\qp$ with a stochastic gradient step} \label{alg:line:parameter-update-phi}
          \EndFor
        \EndFor 
    \end{algorithmic}
\end{algorithm}

\paragraphJMLR{Warm-up.} The warm-up stage (lines~\ref{alg:line:var-warmup}\nobreakdash-\ref{alg:line:density-warmup}) has two parts: a variational model $\qp$ initialisation stage (line~\ref{alg:line:var-warmup})  and a statistical model $\pt$ warm-up stage (line~\ref{alg:line:density-warmup}), which we describe below in the given order. We note that the warm-up stage can be optional, however we empirically found that it allows the models to take reasonable initial values, which can sometimes improve the model fit or stabilise the learning for a small additional initialisation cost.

The warm-up procedure for the variational model is outlined in pseudo-code in Algorithm~\ref{alg:var-warm-up}.
In standard variational inference for latent variable models, the variational model is often initialised randomly. However, in the missing-data case, starting-out with randomly-initialised variational conditionals may be sub-optimal since there is usually some observed data that could be used to reasonably initialise the variational model.
Therefore, to make use of the available observed data, we suggest pre-training the variational inference networks on the observed values using maximum-likelihood estimation and stochastic gradient ascent (SGA) via
\begin{align}
    \hat \phib &= \argmax_{\phib} \J_\phib(\D_K), \label{eq:var-model-pretraining}\\
     \J_\phib(\D_K) &= \sum_{j=1}^d \frac{1}{\lvert \obs(\D_K, j) \rvert} \sum_{i \in \obs(\D_K, j)} \frac{1}{K} \sum_{k=1}^K \log \qpj(\xj^i \mid \xm^\sik, \xonj^i), \nonumber
\end{align}
where $\obs(\D_K, j)$ represents the set of indices of data-points that are observed in the $j$-th dimension. This pre-training procedure is probabilistic regression for all dimensions $j \in [1, d]$ using the data where the $j$-th dimension is observed, with the missing values imputed with a baseline imputation method (for example, samples from the empirical marginals).\footnote{\revisiontwo{Care must be taken when warming-up the extended-Gibbs kernel to avoid poor initialisation of the conditionals, see technical aside in Appendix~\ref{apx:vgi-derivation-extended}.}}
We find that a small number of pre-training iterations is often sufficient to favourably initialise the variational model, which can boost the performance of the method and stabilise training in the initial iterations of the main stage of the algorithm.

\begin{figure}[tpb]
\centering
\vspace{-1em}
\begin{algorithm}[H]
    \caption{Variational model warm-up}
    \label{alg:var-warm-up}
    \textbf{Input:} $\qpj(\xj \mid \xnj)$ for $j \in \{1 \ldots d\}$, variational conditionals; $\D_K$, $K$-times imputed data;\\
    \phantom{\textbf{Input:}} $\alpha_\phib$, the parameter learning rate; \emph{var\_warmup\_epochs}, number of epochs.\\
    \textbf{Output:} $\phib$ initialised on observed data using \eqref{eq:var-model-pretraining}\\
    \begin{varwidth}[t]{0.43\textwidth}
       \begin{algorithmic}[1]
            \For{$t_w$ in $[1, \emph{var\_warmup\_epochs}]$}
                \For{mini-batch $\B_K$ in $\D_K$}
                    \State $\J_\phib = 0$
                    \For{$j$ in $[1, d]$}
                        \State $\J_\phib \mathrel{+}= \frac{1}{\lvert \obs(\B_K, j) \rvert}\J^j_\phib(\B_K, j)$
                    \EndFor
                    \State $\phib \leftarrow \phib + \alpha_\phib \nabla_\phib \J_\phib$
                \EndFor
            \EndFor
        \algstore{var-model-warmup}
        \end{algorithmic}
    \end{varwidth}\hspace{0.5em}
    \begin{varwidth}[t]{0.58\textwidth}
    \begin{algorithmic}[1]
    \algrestore{var-model-warmup}
        \Function{$\J^j_\phib$}{$\B_K, j$}:
        \State $\J^j_\phib = 0$
        \For{$\xo^i, \m^i, \xm^{(i, 1)}, \ldots, \xm^{(i, K)}$ in $\B_K$}
            \If{$\xj^i$ is observed}
                \State $\J^j_\phib \mathrel{+}= \frac{1}{K} \sum_{k=1}^K \log \qpj(\xj^i \mid \xm^\sik, \xonj^i)$ %
            \EndIf
        \EndFor
        \State \textbf{return} $\J^j_\phib$
        \EndFunction
    \end{algorithmic}
    \end{varwidth}
\end{algorithm}\vspace{-1em}
\begin{algorithm}[H]
    \caption{Compute VGI objective in \eqref{eq:vgi-objective-approx}}
    \label{alg:vgi-objective}
    \textbf{Input:} $\pt(\x)$, statistical model; $\qpj(\xj \mid \xnj)$ for $j \in \{1 \ldots d\}$, variational conditionals;\\
    \phantom{\textbf{Input:}} $\B_K$, $K$-times imputed mini-batch; $M$, number of missing dimensions to sample.\\
    \textbf{Output:} $\hJVGI(\thetab, \phib; \xo)$ averaged over all $\xo^i$ in mini-batch $\B_K$\\
    \begin{varwidth}[t]{0.54\textwidth}
       \begin{algorithmic}[1]
            \State $\hJVGI = 0$ 
            \For{$\xo^i, \m^i, \xm^{(i, 1)}, \ldots, \xm^{(i, K)}$ in $\B_K$}
                \For{$k$ in $[1, K]$}
                        \State $\hJVGI \mathrel{+}= \frac{1}{\lvert \B_K \rvert K} \hJVGI^{(i, k)}(\xo^i, \xm^{(i, k)})$
                \EndFor
            \EndFor
        \algstore{vgi-estimate}
        \end{algorithmic}
    \end{varwidth}\hspace{0.5em}
    \begin{varwidth}[t]{0.48\textwidth}
    \begin{algorithmic}[1]
    \algrestore{vgi-estimate}
        \Function{$\hJVGI^{(i, k)}$}{$\xo^i, \xm^{(i, k)}$}:
            \State $\hJVGI^{(i, k)} = 0$
            \For{$m$ in $[1, M]$}
                \State $j^m \sim \pib(j)$ \Comment{Sample from $\idx{\m^i}$} %
                \State $x^k_{j^m} \sim \qpj(x^k_{j^m} \mid \x^{(i, k)}_{\mis\smallsetminus j^m}, \xo^i)$ %
                \State $\hJVGI^{(i, k)} \mathrel{+}= \log \frac{\pt(x_{j^m}^{k}, \x_{\mis \smallsetminus j^{m}}^{(i, k)}, \xo^i)}{\qpj(x_{j^m}^{k} \mid \x_{\mis \smallsetminus j^m}^{(i, k)}, \xo^i)}$ %
            \EndFor
            \State \textbf{return} $\frac{1}{M} \hJVGI^{(i, k)}$
        \EndFunction
    \end{algorithmic}
    \end{varwidth}
\end{algorithm}
\vspace{-2em}
{
\begin{minipage}[t]{.47\textwidth}
\begin{algorithm}[H]
    \caption{Statistical model warm-up}
    \label{alg:model-warm-up}
    \textbf{Input:} $\pt(\x)$, statistical model\\
    \phantom{\textbf{Input:}} $\qpj(\xj \mid \xnj)$ for $j \in \{1 \ldots d\}$\\
    \phantom{\textbf{Input:}} $\D_K$, K-times imputed data\\
    \phantom{\textbf{Input:}} $\alpha_\thetab$, the parameter learning rate\\
    \phantom{\textbf{Input:}} \emph{model\_warmup\_epochs}, \# of epochs\\
    \textbf{Output:} $\thetab$ initialised for the main loop
    \begin{algorithmic}[1]
        \For{$t_w$ in $[1, \emph{model\_warmup\_epochs}]$}
            \For{mini-batch $\B_K$ in $\D_K$}
                \State \textbf{Estimate} $\hJVGI$ using Algorithm~\ref{alg:vgi-objective}
                \State $\thetab \leftarrow \thetab + \alpha_\thetab \nabla_\thetab \hJVGI$
            \EndFor
        \EndFor
    \end{algorithmic}
\end{algorithm}
\end{minipage}
\hfill
\begin{minipage}[t]{.48\textwidth}
\begin{algorithm}[H]
    \caption{(Pseudo-)Gibbs sampling}
    \label{alg:pseudo-Gibbs}
    \textbf{Input:} $\qpj(\xj \mid \xnj)$ for $j \in \{1 \ldots d\}$\\
    \phantom{\textbf{Input:}} $\B_K$, K-times imputed mini-batch\\
    \phantom{\textbf{Input:}} $G$, number of Gibbs update steps\\
    \textbf{Output:} $\B_K$ with updated imputations
    \begin{algorithmic}[1]
        \For{$\xo^i, \m^i, \xm^{(i, 1)}, \ldots, \xm^{(i, K)}$ in $\B_K$}
            \For{$k$ in $[1, K]$}
                \For{$g$ in $[1, G]$}
                    \State $j \sim \pib(j)$ \Comment{Sample from $\idx{\m^i}$}
                    \State $\xj \sim \qpj(\xj \mid \xmnj^\sik, \xo^i)$ %
                    \State $\xm^\sik \leftarrow (\xmnj^\sik, \xj)$ %
              \EndFor
            \EndFor
        \EndFor 
    \end{algorithmic}
\end{algorithm}
\end{minipage}
}
\end{figure}

Next, the algorithm performs SGA on the parameters $\thetab$ of the statistical model $\pt$ using the variational Gibbs objective $\hJVGI$ in \eqref{eq:vgi-objective-approx} (see Algorithms~\ref{alg:vgi-objective} and \ref{alg:model-warm-up}). 
This kind of ``warm-up'' allows the parameters $\thetab$ to take on reasonable initial values before the main iterative stage starts.
We qualitatively find that the warm-up stage generally needs to be performed for a small number of iterations until the change in $\hJVGI$ between consecutive iterations falls below a threshold.
Note that we keep the variational parameters $\phib$ fixed at this stage so that the variational model does not deteriorate while the model $\pt$ is being initialised.
Also, in this stage the imputed data in $\D_K$ are not yet updated, they still follow the initial imputation distribution $f^0$. This is because we empirically found that early Gibbs sampling can cause divergent imputation chains, training instability, or getting stuck in a local optima.

\paragraphJMLR{Main stage.} The main stage of the algorithm iterates between updating the imputations in $\D_K$ and fitting the parameters $\thetab$ and $\phib$ (see lines~\ref{alg:line:Gibbs-update}\nobreakdash-\ref{alg:line:parameter-update-phi}). For each mini-batch $\B_K$ we first update the imputed values $\xm^{(i, 1)}, \ldots, \xm^{(i, K)}$ using $G$ Gibbs updates with the variational conditionals (see Algorithm~\ref{alg:pseudo-Gibbs}) and then update the parameters $\thetab$ and $\phib$ using a single stochastic gradient of the variational Gibbs objective $\hJVGI^t$ in \eqref{eq:vgi-objective-approx}, see Algorithm~\ref{alg:vgi-objective} on computing the objective. These two steps (imputation and parameter update) are then repeated until convergence or until the computational budget is exhausted.

We find that using a small number of Gibbs updates $G$ is generally sufficient and preferable in terms of the trade-off between compute cost and convergence rate.\footnote{We used $G \in [1, 5]$ in our experiments.} Moreover, a large value of $G$ may cause training instability due to the generalisation gap of the variational conditionals (see Section~\ref{sec:vgi-eval-held-out} for more details).
Updating the imputations via Gibbs sampling using the variational conditionals and storing them for the next iteration corresponds to updating the marginal Markov chain distribution $f^t(\xm \mid \xo)$. 
Reusing the imputations from the previous iteration, rather than re-sampling from scratch at each iteration, is akin to using persistent chains that have previously been used in a different context in Markov  chain  Monte  Carlo  methods \citep[\eg\ persistent contrastive divergence,][]{younesConvergenceMarkovianStochastic1999, Tieleman2008}.

\subsection{Choosing the Variational Model}
\label{sec:variational-model}

The performance of amortised variational methods, and hence also VGI, critically depends on the choice of the functional family and the expressivity of the inference networks used to parametrise them.
To maximise the compatibility of the statistical model $\pt$ and the variational model $\qp$, we suggest the variational family should be chosen as one that includes or is close to the family of the statistical model. If the target model $\pt$ is a deep model, the variational inference networks should use architectural blocks in the neural network that are similar to the ones used in the target model.

We must also consider how to specify and parametrise the $d$ variational conditionals.
A straightforward way to specify them is to use $d$ inference networks with parameters $\phibj$, one for each variational conditional.
However, such an approach would be parameter-inefficient and scale poorly to higher dimensional data. 
To address this, we suggest using the extended-Gibbs conditionals $\qpj(\xj^t \mid \xo, \txj^{t-1}, \txmnj^{t-1})$ (Appendix~\ref{apx:vgi-derivation-extended}), which allow the imputation $\xj^t$ to depend on the previous imputation value $\txj^{t-1}$.
Then, we can use a single partially-shared neural network where the parameters and the computations are shared in the first part of the network for all conditional distributions.
These two simple modifications allows us to scale VGI to higher-dimensional data (see Appendix~\ref{apx:variational-network} for a more detailed discussion).
We investigate the effects of parameter-sharing and extended conditionals in Section~\ref{sec:fa-accuracy-variational-model}.

\revisiontwo{
Finally, an important caveat of our approach is that, due to approximation errors, the fitted variational conditionals may not correspond to a joint distribution, which mirrors a similar caveat of the FCS methods outlined in Section~\ref{sec:background-fcs}. Hence, pseudo-Gibbs sampling using the fitted conditionals may diverge (we revisit this in Section~\ref{sec:vgi-eval-held-out} when discussing model selection).\footnote{Note that, after learning, we advocate using the estimated $p_\theta$ for generating imputations if they were required and not the variational conditionals.} 
Nevertheless, we show in the experimental sections that the VGI method works well despite the lack of such convergence guarantees, similar to the existing FCS imputations methods (see Section~\ref{sec:background-fcs}).
}

\subsection{Variational Block-Gibbs Inference and Latent-Variable Models}
\label{sec:vbgi}

Thus far we have considered the general case where the number of possible missingness patterns is $2^{d}$, with $d$ being the dimensionality of the data. In some cases, missingness in a block of dimensions may be coupled such that either all of the dimensions in the block are missing or all are observed, which reduces the number of possible missingness patterns to $2^s$, where $s$ is the number of such missing variable blocks, with $s=d$ if the missingness in no dimension is coupled.

We can adapt the VGI lower-bound in \eqref{eq:variational-Gibbs-elbo} and equivalently the VGI objective in \eqref{eq:vgi-objective} to this scenario by letting $j$ denote the index of a block of missing dimensions, rather than the index of a single missing dimension. 
Hence, the univariate $\xj$ in \eqref{eq:variational-Gibbs-elbo} and \eqref{eq:vgi-objective} become potentially multivariate random variables $\x_j$.
We must then specify a variational model for each block of missing dimensions, where we may choose to specify independent variational models or partially-shared models as discussed in Section~\ref{sec:variational-model}.
The Gibbs update step in line~\ref{alg:line:Gibbs-update} of Algorithm~\ref{alg:VGI} then corresponds to the update of a block-Gibbs sampler.
Hence, we refer to this method as the variational block-Gibbs inference (VBGI).

A particularly important instance of coupled missingness is the latent-variable model $\pt(\xo, \xm \mid \xz)\pt(\xz)$, where all of the latent-variables $\xz$ are missing together.
If the missingness of no other dimension is coupled, then $j \in \idx{\m} \cup \{ z \}$ refers to either any one of the missing dimensions or all of the latents.
Rewriting $\JVGI^t$ for this model we get the following objective:\footnote{Similar to the standard VGI case, the VBGI objective can optionally be adapted to use the extended-Gibbs conditionals (see Appendix~\ref{apx:vgi-derivation-extended}).}
\begin{align}
    \J^t_\text{VBGI}(\thetab, \phib; \xo) &= \E_{f^{t-1}(\txm, \txz \mid \xo)} \Bigg( \nonumber \\
    &\phantom{=}\sum_{j \in \idx{\m}} \pib(j) \E_{\qpj(\xj \mid \txmnj, \txz, \xo) } \left[ \log \frac{\pt(\xj, \txmnj, \xo \mid \txz) \pt(\txz)}{\qpj(\xj \mid \txmnj, \txz, \xo)} \right] \nonumber\\
    &\phantom{=}+ \pib(j=z) \E_{q_{\phib_z}(\xz \mid \txm, \xo) } \left[ \log \frac{\pt(\txm, \xo \mid \xz) \pt(\xz)}{q_{\phib_z}(\xz \mid \txm, \xo)} \right] \Bigg), \label{eq:vbgi-latent-objective}
\end{align}
where $q_{\phib_z}(\xz \mid \txm, \xo)$ is the variational conditional for the latents $\xz$. The grouping of missing dimensions allows us to adapt VGI to latent-variable models. %
It forms the basis for the variational autoencoder-specific version of VGI that we introduce in Section~\ref{sec:vae-experiments}. 

\subsection{Evaluating VGI on Incomplete Held-Out Data}
\label{sec:vgi-eval-held-out}

A common step in a machine learning workflow is the validation of the estimated model on a held-out data set, which is used in early stopping or model selection \citep[\eg][]{Goodfellow2016}.
In the incomplete data case, we can expect that the held-out data are also incomplete.
One approach is to estimate the marginal log-likelihood $\log \pt(\xo)$ using importance sampling or sequential Monte Carlo methods \citep[\eg][]{Barber2017}, however these approaches can be too computationally expensive to obtain reliable estimates of the marginal on the held-out data, especially if we wanted to evaluate many models.
Alternatively, a validation loss, analogue to the training loss, is commonly used to select a model that minimises it. 
The evaluation of the VGI objective in \eqref{eq:vgi-objective} on incomplete held-out data requires the iterative evaluation of the variational conditionals (Gibbs sampling) on unseen data. 
However, evaluating these conditionals on unseen data has known pitfalls. We will first review these pitfalls, then explain how it affects the evaluation of VGI, and finally suggest a simple procedure that enables the evaluation of the VGI objective on held-out data and produces sample imputations.

\citet{cremerInferenceSuboptimalityVariational2018} have shown that amortised variational distributions may be significantly biased compared to the optimal variational distribution in the same variational family, a phenomenon they called the amortisation gap. They have also demonstrated that the amortisation gap is typically larger on held-out data if the variational model is fitted only to the training data; we will refer to this as the \emph{inference generalisation gap} \citep[see also the concurrent work by][]{zhangGeneralizationGapAmortized2021}.
The reason for the generalisation gap is overfitting of the variational inference network to its inputs, \ie the training data, which is generally a consequence of using finite-sized training data sets. 
Importantly, the inference generalisation gap can appear even when the target model $\pt$ has not been overfitted.

The same considerations apply to VGI since it is an amortised variational method.
In particular, the generalisation gap in VGI can cause the imputation Markov chains to diverge due to the compounding effect of the generalisation errors. This prevents a meaningful evaluation of VGI on the held-out data. We have empirically observed that the chances of divergent behaviour increases with the number of variational conditional models.
\citet{matteiRefitYourEncoder2018} and \citet{cremerInferenceSuboptimalityVariational2018} suggested to deal with the inference generalisation gap by fine-tuning the variational models on the held-out data. 
To avoid information leakage from the held-out data into the model $\pt$, the estimated model and the corresponding parameters $\thetab$ are kept fixed during fine-tuning. Moreover, the fine-tuned variational conditionals should not be used as part of the VGI training. A simple way to achieve this is to make a copy of the variational model before fine-tuning and discarding it afterwards.

In Appendix~\ref{apx:vgi-finetuning}, Algorithm~\ref{alg-apx:VGI-finetuning} we show the VGI fine-tuning procedure.
The fine-tuning starts with incomplete held-out data and then fills\revisiontwo{-in} the missing values with random draws from the initial imputation distribution $f^0$ as in Algorithm~\ref{alg:VGI}. 
Then, during the first iteration it performs several rounds of Gibbs sampling over all missing values using the learnt variational conditionals to improve the imputations.
To mitigate possible divergent behaviour due to the inference generalisation gap in the first iteration, the procedure rejects any values outside of the observed-data hypercube, defined by the minimum and maximum values in the observed data.
After the imputation warm-up in the first iteration, the algorithm continues akin to the learning of $\phib$ in standard in VGI, fine-tuning the variational kernel $\tau_\phib$ and updating the imputations with a small number of Gibbs updates in each iteration, where all proposed imputations are accepted---the acceptance region is no longer necessary, since the kernel is being adapted to the (imputed) held-out data.
At the end of the procedure, we obtain the VGI loss on the held-out data as well as imputations of the missing values.

The iterative validation procedure described in this section would be expensive if it were used to continually monitor the validation loss during training. However, the validation loss does not need to be computed at every iteration, computing it only every few iterations is often sufficient. Moreover, we have empirically found that the cost of fine-tuning was often only a small fraction of the training cost.

\subsection{Related Work}
\label{sec:vgi-related-methods}
We here discuss work that is closely related or shares similarities to VGI.
\paragraphJMLR{Monte Carlo expectation-maximisation.}
Monte Carlo EM \citep[MCEM,][]{Wei1990} has been proposed as an extension of the classical expectation-maximisation \citep[EM,][]{Dempster1977} algorithm to the setting where the required expectation with respect to the conditional distributions of missing data does not have a closed form expression but is approximated by a Monte Carlo sample average. The method requires sampling from the conditional distribution of missing data at each iteration.
Then, like VGI, MCEM iteratively maximises the ELBO using sample imputations of the missing data. 
In contrast to VGI, MCEM does not use a variational approximation but instead attempts to sample the true conditional distribution, for example, with methods such as rejection sampling \citep[\eg][Chapter~27.1.2]{Barber2017}.

However, sampling from the conditional distributions in higher dimensions usually requires Markov chain Monte Carlo methods \citep[MCMC, \eg][Chapter~27.4]{Barber2017}, which only asymptotically sample from the exact target distribution. 
In practice, MCMC is used to sample chains of fixed length for computational reasons and hence it may not sample the true conditional distribution.
As a consequence, using samples from an unconverged Markov chain in MCEM may adversely affect the learnt model $\pt$.

Similar to MCEM with MCMC, VGI also samples a Markov chain of imputations using the variational kernel. However, in contrast to MCEM, where the MCMC sampler needs to be restarted at each iteration and run for a sufficient amount of time, in VGI we use a single persistent Markov chain that we update throughout training. This allows our method to eventually sample better imputations of the missing data given that the algorithm is run for long enough.
In Section~\ref{sec:flow-experiments} we estimate a normalising flow model \citep{Rezende2015, Papamakarios2021} from incomplete data with VGI and MCEM using flow-specific MCMC and find that VGI outperforms MCEM both in terms of the computational performance and accuracy.
In an attempt to improve the performance of MCEM, we have empirically investigated the use of persistent chains in a Metropolis-Hastings version of MCMC on normalising flows, but found that this drastically reduced the acceptance rate of the proposed transitions and therefore also reduced the accuracy of the estimated statistical model.

\paragraphJMLR{Markov chain variational inference.}
\citet{Salimans2015} proposed a powerful framework for combining Markov chain Monte Carlo methods and variational inference for latent-variable models called Markov chain variational inference (MCVI).
Similar to VGI they propose a lower-bound on the log-likelihood using a Markov chain characterised via a variational transition kernel $\tau$.
However, the two methods differ fundamentally in their goals: VGI attempts to \emph{learn} a statistical model $\pt$ from incomplete data while MCVI attempts to efficiently and accurately approximate a conditional distribution under a \emph{fixed} latent-variable model.
To achieve their goal, MCVI aims to optimise the variational kernel over the full length of the Markov chain.
In order to avoid computing the intractable integral required to obtain the marginal distribution of a Markov chain $f_\phib^t$ they propose a further lower-bound on the ELBO in \eqref{eq:elbo-marginal-markov-chain} using a ``reverse'' transition kernel $r$, which predicts the reverse path of the Markov chain sampled using the variational kernel $\tau$. 
A na\"ive application of MCVI to learn a statistical model would be expensive, since it requires simulating long Markov chains of imputations and then backpropagating the gradients through the sampling path.
To simplify the optimisation problem, the authors of MCVI have also briefly considered a sequential (greedy) approach similar to ours.
However, as shown in Section~\ref{sec:var-objective}, VGI does not need to learn an auxiliary model of the ``reverse'' kernel $r$, and in fact, replacing $r$ in the MCVI lower-bound with the true reverse transition \citep[\eg][Section~1.4]{Murray2007} recovers the tighter lower-bound in \eqref{eq:elbo-marginal-markov-chain} (see Appendix~\ref{apx:mcvi}), which is approximately marginalised in VGI.

A further difference between VGI and MCVI is that MCVI has been developed for the latent-variable and not the missing data setting. Hence, unlike VGI, it does not deal with the problem of how to handle the $2^d-1$ possible patterns of missingness and the consequent exponential growth in the number of required variational distributions.

\paragraphJMLR{Coordinate ascent variational inference.}
The VGI objective in \eqref{eq:vgi-objective} is related to coordinate ascent variational inference \citetext{CAVI, \citealp[\eg][Chapter~10.1.1]{Bishop2006}; \citealp[][Section~2.4]{Blei2017}}, which also considers the update of only a single dimension using a univariate variational distribution. However, the previous works only considered fully-factorised variational distributions (mean-field assumption), which can significantly bias the estimate of the statistical model as we have demonstrated in Section~\ref{sec:background-vi}.
In contrast, our method works without introducing the mean-field factorisation using a highly-flexible implicit variational distribution given by the marginal of a pseudo-Gibbs sampler, which is adapted to the model $\pt$ by learning the transition kernel (univariate variational conditionals) of the sampler.

\paragraphJMLR{Arbitrarily-conditional models.}
\revisiontwo{A separate line of research focuses on constructing models such that any conditional distribution $\pt(\x_{\text{u}} \mid \xo)$, where $\x_\text{u} \subseteq \x \smallsetminus \xo$ may be arbitrarily chosen, is directly available \citep{Li2020a}. 
One could use such models to construct an amortised variational distribution for arbitrary missingness patterns $\pt(\xm \mid \xo)$ and use it in the variational ELBO to fit a target statistical model. 
However, unlike the variational conditionals in VGI that we can choose freely, such arbitrarily-conditional models typically have restricted modelling capabilities compared to their non-conditional counterparts.}

\paragraphJMLR{Substantive model compatible FCS.}
As discussed in Section~\ref{sec:background-fcs} our solution to the $2^d-1$ conditional missing data distributions bears similarity to existing imputation methods in the fully-conditionally specified (FCS) family. 
Standard FCS methods are independent of the target analysis model. However, \citet{bartlettMultipleImputationCovariates2015} have proposed a modified version of the method, called substantive model compatible FCS (SMC-FCS), to generate imputations that are congenial with target (nonlinear) regression model. There is thus some similarity to VGI in the sense that both methods take the target model into account. However, SMC-FCS is about imputing data compatible with regression models (and hence is \revisiontwo{for} supervised learning), while our method estimates joint statistical models (and hence is \revisiontwo{focused} on unsupervised learning). From our understanding SMC-FCS does not apply or readily extend to our setting.

\paragraphJMLR{Bayesian data augmentation.}
Alternatively to maximum-likelihood estimation (MLE) methods, such as the EM or VI, one can also use Bayesian inference to learn models from incomplete data. 
The advantage of Bayesian methods is that they provide a natural way to evaluate the epistemic uncertainty of the model.
One classical method is the \revisiontwo{data augmentation (DA) algorithm} \citep{Tanner1987}. %
In DA, one must first specify a prior over the parameters of the model $p(\thetab)$ and assume initial imputations $X_\mis^{(0)}$.\footnote{We use the capital $X$ to denote all data-points $\x^i$ in the data set.} 
Then, the algorithm iteratively samples the posterior over the parameters $\thetab^{(t)} \sim p(\thetab \mid X_\mis^{(t-1)}, X_\obs)$ and updates imputations (or augmentations) $X_\mis^{(t)} \sim p(X_\mis^{(t)} \mid X_\obs, \thetab^{(t)})$. 
After an initial warm-up (or burn-in) period, the algorithm produces samples of the missing values and model parameters from the posterior distribution $p(X_\mis, \thetab \mid X_\obs)$.
Similar to VGI, the iterative procedure in DA is a Gibbs sampler. The main difference between the two is that DA treats the parameters of the model just like the missing variables and samples them accordingly.
However, for complex models, just like in the MCEM case, the required distributions are usually only known up to a proportionality and hence computationally expensive MCMC methods would be required to sample them.
Moreover, Bayesian inference for modern statistical models, such as VAEs and flows, suffers from scalability issues and is still an active area of research \citep{galUncertaintyDeepLearning2016, Maddox2019, izmailovWhatAreBayesian2021, abdarReviewUncertaintyQuantification2021}.

\section{Experiments on Toy Models}
\label{sec:toy-experiments}
In this section we demonstrate VGI on low- and high-dimensional factor analysis models. We analyse the accuracy of the learnt statistical models and the variational conditionals as well as the effect of the extended-Gibbs variational conditionals. The code for this and the following experimental sections is available at \url{https://github.com/vsimkus/variational-gibbs-inference}.

\subsection{Factor Analysis Model}
\label{sec:fa-model}

Factor analysis \citep[FA, \eg][Chapter~21.1]{Barber2017} is a linear latent-variable model that is often used to discover unobserved factors $\z$ from observed data $\x$.
The prior distribution of $\z$ is assumed to be a multivariate standard Gaussian $p(\z) = \N(\z; \bm{0}, \bm{I})$, and the distribution of $\x$ given a value of $\z$ is
\begin{align*}
    \pt(\x \mid \z) = \N(\x; \Fb \z + \mub, \Psib),\nonumber
\intertext{and the marginal distribution is}
    \pt(\x) = \N(\x; \mub ,\Fb\Fb^\top + \Psib),%
\end{align*}
where the parameters $\thetab$ are the mean vector $\mub$, the factor matrix $\Fb$, and the diagonal matrix $\Psib$ defining the variances of the observation noise. 

FA is a linear version of the variational autoencoder \citep[VAE,][]{Kingma2013, Rezende2014a} that is often used as a toy model to analyse new methods \citep[\eg][]{Williams2018}. Hence, we start our analysis on a FA model before proceeding to evaluate VGI on more complex models in the next sections. 
Given its relative simplicity, it is possible to fit a FA model on incomplete data with the expectation-maximisation (EM) algorithm (see Appendix~\ref{apx:em-for-fa} for details). The EM algorithm provides a best-case solution that we can use to gauge the accuracy of the model estimated by VGI (see below).
Moreover, reference conditional distributions $\pt(\xj \mid \xnj)$ can be computed analytically \citep{Petersen2012}, such that we can evaluate the accuracy of the variational conditionals, which is instrumental to producing good imputations and subsequently---to achieving an accurate fit of the model.

\subsection{Data}
\label{sec:fa-data}
We evaluate VGI on two data sets:
\begin{description}
    \item[Toy data.] A synthetic data set generated with a 6-dimensional FA model with a 2-dimensional latent space (see Appendix~\ref{apx:toy-ground-truth} for the ground truth parameters). The training data set has 6400 data-points and the test data set has 5000 data-points.
    \item[FA-Frey.] A synthetic 560-dimensional data set based on the Frey data,\footnote{The original data set is available at \url{https://cs.nyu.edu/~roweis/data/frey_rawface.mat}.} where a FA model with 43 latent dimensions was first fitted on the original data and then used to synthesise a new data set. The training data set has 2400 data-points and the test data set has 3000 data-points.
\end{description}

We consider five fractions of missingness in the training data, ranging from 16.6\% to 83.3\%, and simulate incomplete training data by generating a binary missingness mask uniformly at random (MCAR). 
The data-points that are rendered fully missing are removed from the training set, since doing so does not affect the maximum-likelihood estimate under MAR or MCAR missingness.

Where the experiments are repeated multiple times to obtain confidence intervals, the underlying training data are kept constant and only the missingness mask is re-sampled. Thus, the results demonstrate the accuracy of the estimation methods in the presence of missing data, rather than the variability of the model estimate on different realisations from the ground truth data generating distribution.

\subsection{Experimental Settings}
\label{sec:fa-vgi-settings}

In our evaluation, we assume that the statistical model $\pt$ is well-specified. That is, we fit a FA model to data following a FA model with the same latent dimensionality, so that the evaluation can focus on the effect of missing data, rather than on robustness to model specification.  We parametrise $\Psib$ as $\Psib = \exp(\gammab)$, initialise the parameter $\Fb$ with samples from a standard normal distribution, and set $\mub$ and $\gammab$ to $\bm{0}$ and $\bm{1}$ respectively. 

We specify the variational conditionals in VGI to be univariate Gaussians whose parameters $\log \sigma^2_j$ and $\mu_j$ are given by the outputs of a fully-connected neural network.
For the toy data we use the standard variational conditionals with an independent network of two hidden layers for each conditional. 
To make the computations more efficient on the FA-Frey data, we use the extended variational conditionals with a partially-shared network: the first two hidden layers share parameters and computations for all conditional distributions and the last hidden layer has independent parameters for each distribution. 
The networks use leaky ReLU activation functions with negative slope of $0.01$ and the weights are initialised using Kaiming initialisation \citep{He2015}.

We compute the univariate Gaussian entropy terms in \eqref{eq:vgi-objective} analytically using \citep[\eg][]{Norwich1993}
\begin{align*}
    - \E_{\N(\xj ; \mu_j, \sigma_j^2)}\left[ \log \N(\xj ; \mu_j, \sigma_j^2)\right]%
    &= \frac{1}{2} \log(2 \pi \sigma_j^2) + \frac{1}{2},
\end{align*}
where $\mu_j$ and $\log \sigma_j^2$ are given by the inference networks with input $(\txmnj, \xo)$, or $(\txm, \xo)$ in the case of the extended variational conditionals.

We fit the model $\pt(\x)$ and the variational model using Algorithm~\ref{alg:VGI}. We use $K=5$ imputation chains for each incomplete data-point, and $G=3$ (toy-data) and $G=5$ (FA-Frey) Gibbs updates. In the Monte Carlo averaging in $\hJVGI$ we select $M=1$ (toy-data) and $M=10$ (FA-Frey) missing dimensions.
To compute the gradients with respect to the variational parameters we use the reparametrisation trick \citep{Kingma2013}.
The \revisiontwo{model} parameters $\thetab = (\Fb, \mub, \gammab)$ are optimised using the Adam optimiser \citep{Kingma2014Adam}, whereas the variational parameters $\phib$ are fitted using AMSGrad \citep{Reddi2018}, since we found that using Adam on the variational parameters caused training instability.

\subsection{Comparison Methods}
\label{sec:fa-comparison-methods}
We compare VGI against the following methods:
\begin{description}
    \item[EM (Complete).] The ideal case where no data is missing, fitted using EM for FA.
    \item[Empirical sample imputation.] A weak baseline where the incomplete data is $K=5$ times imputed with random draws from the empirical distribution of the observed values, then the FA model is fitted as described in Appendix~\ref{apx:combining-mi}. 
    \item[EM \citep{Dempster1977}.] An optimal method where the model $\pt$ is fitted using EM for FA with missing data (see Appendix~\ref{apx:em-for-fa}). This method presents the best-case performance that could be achieved with a variational method, which is equivalent to setting the variational distribution in VI equal to the true conditional distribution \citep[\eg][Chapter~11.2.2]{Barber2017}. Also, in contrast to the other methods, where SGA is used, we here compute the updated distribution parameters analytically.
    \item[MICE \citep{VanBuuren2000}.] A pseudo-Gibbs sampler as described in Section~\ref{sec:background-fcs} that has been widely used in statistical analysis of incomplete data. MICE code has originally been made available in R \citep{VanBuuren2000}. We have implemented MICE in Python using the IterativeImputer and Bayesian linear ridge regression as the conditional imputation models, as implemented in the scikit-learn package \citep{Pedregosa2011}.
    MICE is used to produce $K=5$ imputations and then the FA model is fitted as described in Appendix~\ref{apx:combining-mi}.
    MICE is another strong baseline because it should be able to produce imputations that are congenial to the target model, since both MICE conditionals and the target FA model are linear Gaussian models.
\end{description}

\subsection{Accuracy of the Fitted FA Model}
\label{sec:fa-model-accuracy}

In this toy setting no over-fitting was observed, hence the model parameters $\hat \thetab$ from the final training iteration were used in the following evaluation for all methods.

\begin{figure}[tb]
    \centering
    \includegraphics[width=\textwidth]{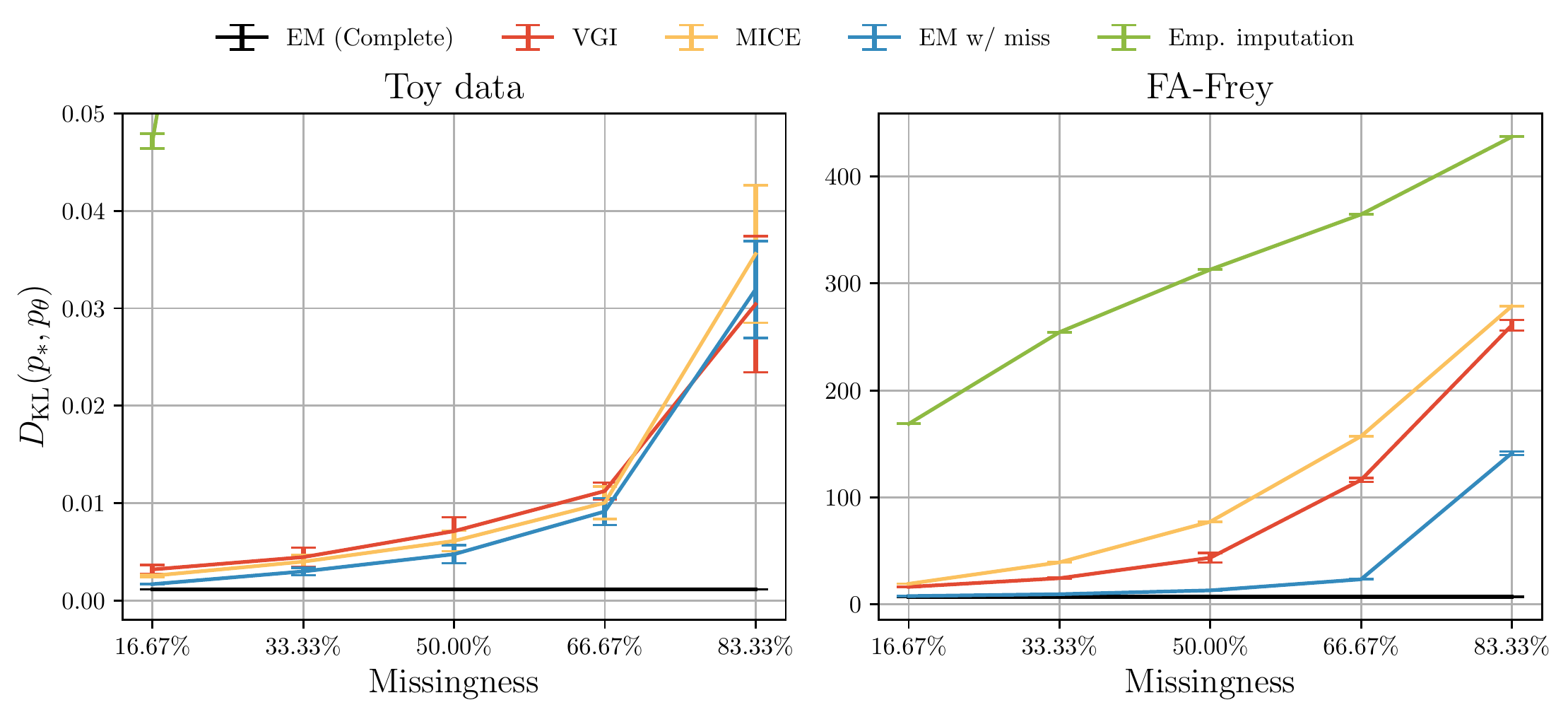}
    \caption{The accuracy of the fitted models $\pt$ as measured by $\KLD{\ps}{\pt}$ on toy (left) and FA-Frey (right) data. Smaller values are better. \revisiontwo{The error bars show standard error on the mean of five experiments with different missingness masks and model initialisation.} In the left figure, empirical sample imputation performed significantly worse than the posterior approximating methods and hence is not shown for missingness larger than 16.67\%.}
    \label{fig:toy-fitted-model-results}
\end{figure}

We evaluate the accuracy of the fitted FA model using the Kullback--Leibler divergence $\KLD{\ps(\x)}{\pt(\x)}$ between the ground truth model $\ps$ and the fitted statistical model $\pt$ on the two synthetic data sets, shown in Figure~\ref{fig:toy-fitted-model-results}. 
It can be immediately seen that the posterior-approximating methods (VGI, MICE, and EM) perform significantly better than the simple empirical imputation baseline showing the clear advantage of these methods over ad-hoc approaches commonly used in practice as a quick fix for missing data. 

As expected, MICE performs well on both data sets since the linear imputation model is congenial with the data distribution and the target model.
VGI performs comparably to MICE on the toy data (note the overlapping error bars) and shows significant improvement on the FA-Frey data. The better performance on FA-Frey can be attributed to the fact that contrary to MICE, where missing value imputations are generated prior and independently of the model $\pt$, in VGI the imputations are generated with respect to the model $\pt$ and are not static throughout training, thus better representing the uncertainty of the imputed values.\footnote{\revisiontwo{We note that the performance of MICE could be improved by generating more imputations, however with additional computations.}}
On the toy data, both VGI and MICE achieved a performance that is comparable to the optimal EM solution, but on the FA-Frey data, the gap between them and EM increases with missingness. 
We show in Figure~\ref{fig-apx:fafrey-kldiv-mcem} of Appendix~\ref{apx:additional-figures} that a similar gap appears between EM and Monte Carlo EM \citep[MCEM,][]{Wei1990} with SGA.
Hence, we attribute the performance gap to the stochasticity in the optimisation---Monte Carlo averaging and stochastic gradient ascent---used within MCEM, MICE, and VGI, and hence the gap could be reduced, given sufficient compute resources, via standard means in stochastic optimisation.

\subsection{Assessing Estimation Consistency}
\label{sec:fa-model-estimation-consistency}

\revisiontwo{We further evaluate the statistical consistency of VGI on the toy FA model. The learning rate was decayed according to a cosine schedule in order to satisfy the convergence conditions of stochastic gradient ascent (SGA) \citep[\eg][Chapter~4.3.2]{Spall2003}. Note that in these experiments the variational family of the variational conditionals includes the true conditional distributions, and hence the estimator is expected to be unbiased.}

\begin{figure}[tpb]
    \centering
    \includegraphics[width=1\textwidth]{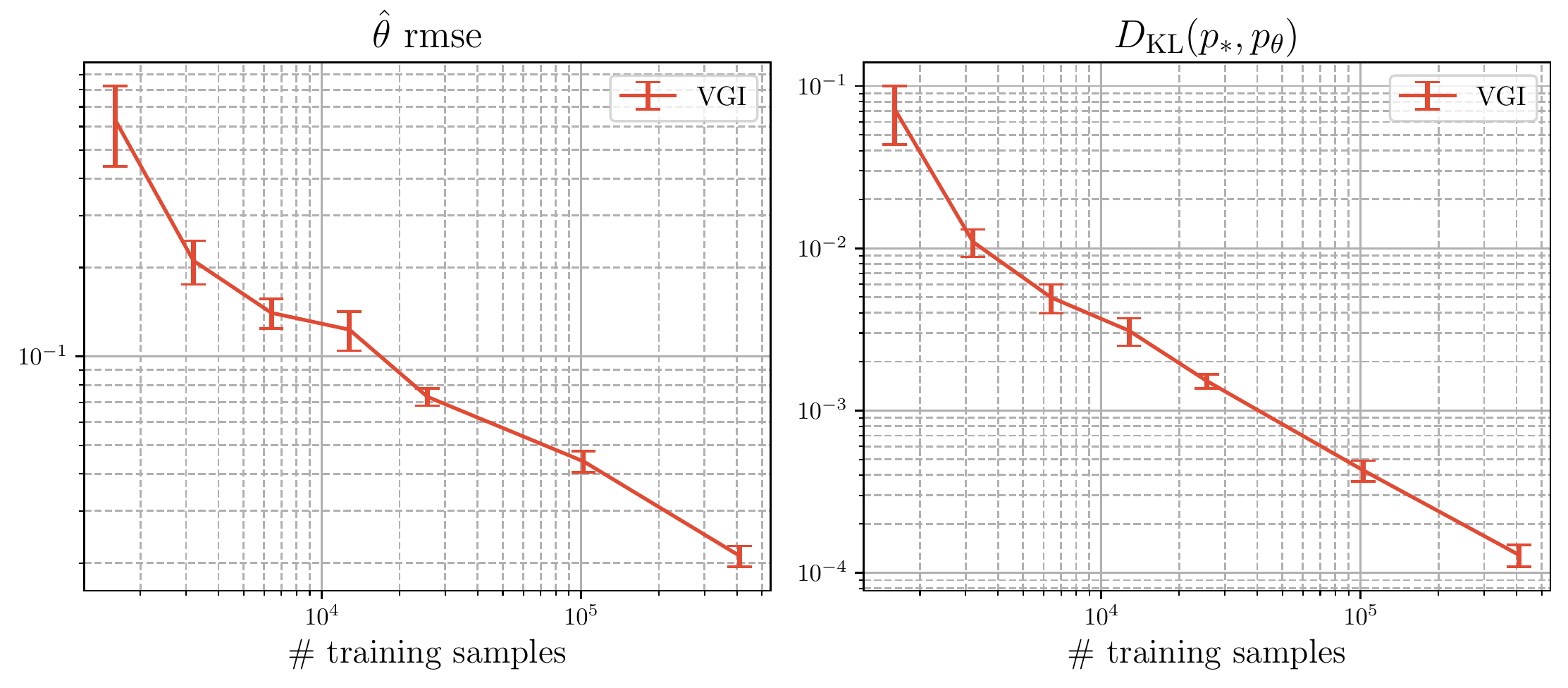}
    \caption{\revisiontwo{The accuracy of the fitted toy FA model in terms of RMSE with the ground truth parameters (left) and KL divergence to the ground truth model (right) as a function of the number of training samples (on a log-log scale). The error bars show standard error on the mean of five experiments with different missingness masks and model initialisation.}}
    \label{fig:fa-toy-fitted-consistency}
\end{figure}

\revisiontwo{In Figure~\ref{fig:fa-toy-fitted-consistency} we plot the log-log curves of the number of incomplete training data-points versus the root mean squared error (RMSE) of the fitted model parameters (left subfigure) and the KL divergence (right subfigure).\footnote{\revisiontwo{The factor loading matrix of the fitted model has been rotated using the (orthogonal) Procrustes rotation \citep[][]{tenbergeOrthogonalProcrustesRotation1977} before computing the RMSE to resolve the partial non-identifiability of the parameter.}}
The plots show a linear behavior in the logarithmic domain, which indicates consistency and conforms with the asymptotic normality of the MLE theory \citep[\eg][Chapter~9.7]{wassermanAllStatisticsConcise2005}.
We note that a few points in the figure fall above the linear curve. We attribute this to normal sample variation and to false convergence of SGD due to a potentially sub-optimal learning rate decay schedule \citep[][Chapter~4.3.2]{Spall2003} that could be addressed via hyper-parameter search.
}

\subsection{Accuracy of the Variational Model}
\label{sec:fa-accuracy-variational-model}

\begin{figure}[tpb]
    \includegraphics[width=1\textwidth]{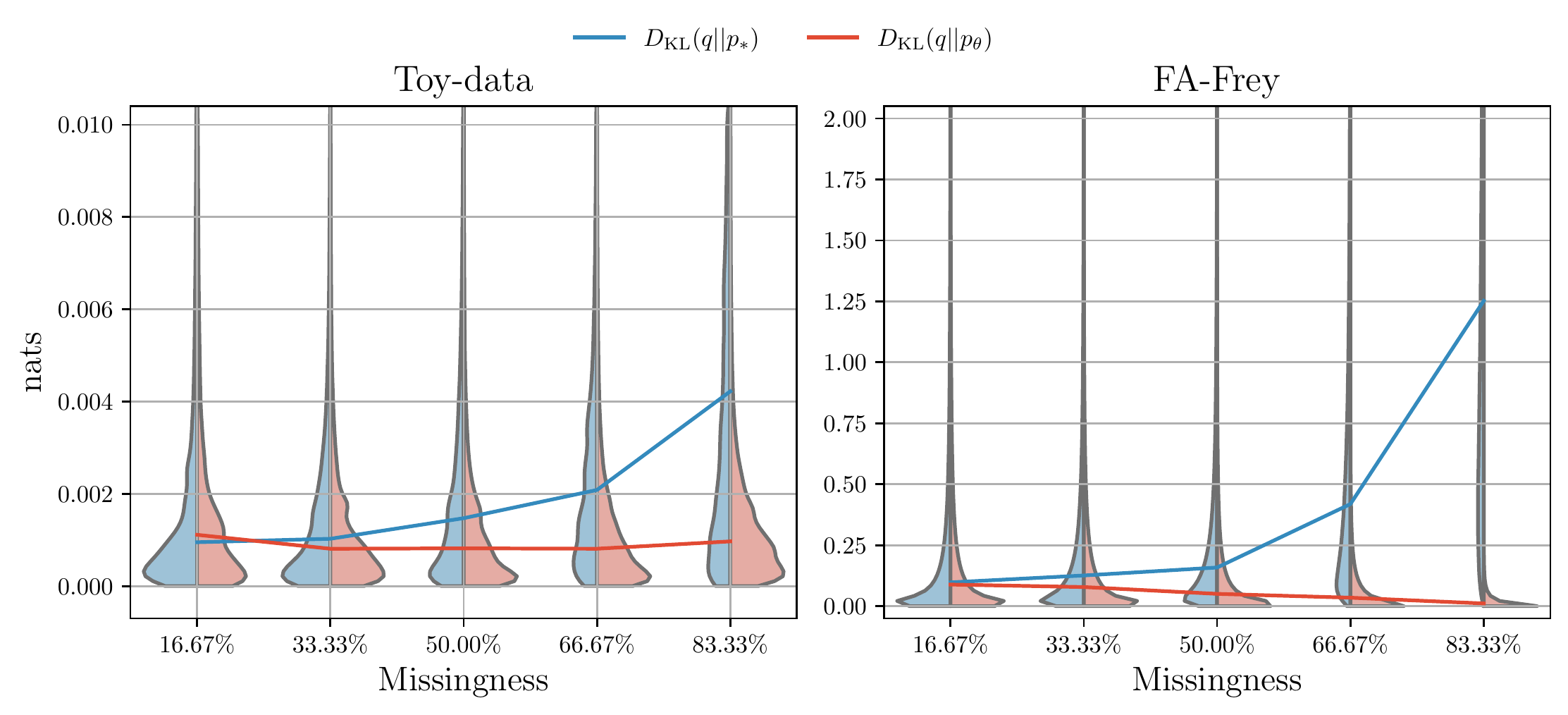}
    \caption{KL divergence, conditioned on the test set (left: toy data, right: FA-Frey data), between the univariate variational conditional distributions and the ground truth distribution $\KLD{\qpj(\xj \mid \xnj)}{\ps(\xj \mid \xnj)}$ (blue), and the posterior under the learnt model $\KLD{\qpj(\xj \mid \xnj)}{\pt(\xj \mid \xnj)}$ (red). The lines show the median KL divergence conditioned on the test set. In Appendix~\ref{apx:additional-figures}, Figure~\ref{fig-apx:toy-posterior-wasserstein} we show a comparison using Wasserstein distance that displays a similar behaviour.}
    \label{fig:toy-posterior-kldiv}
\end{figure}

\begin{figure}[tpb]
    \includegraphics[width=1\linewidth]{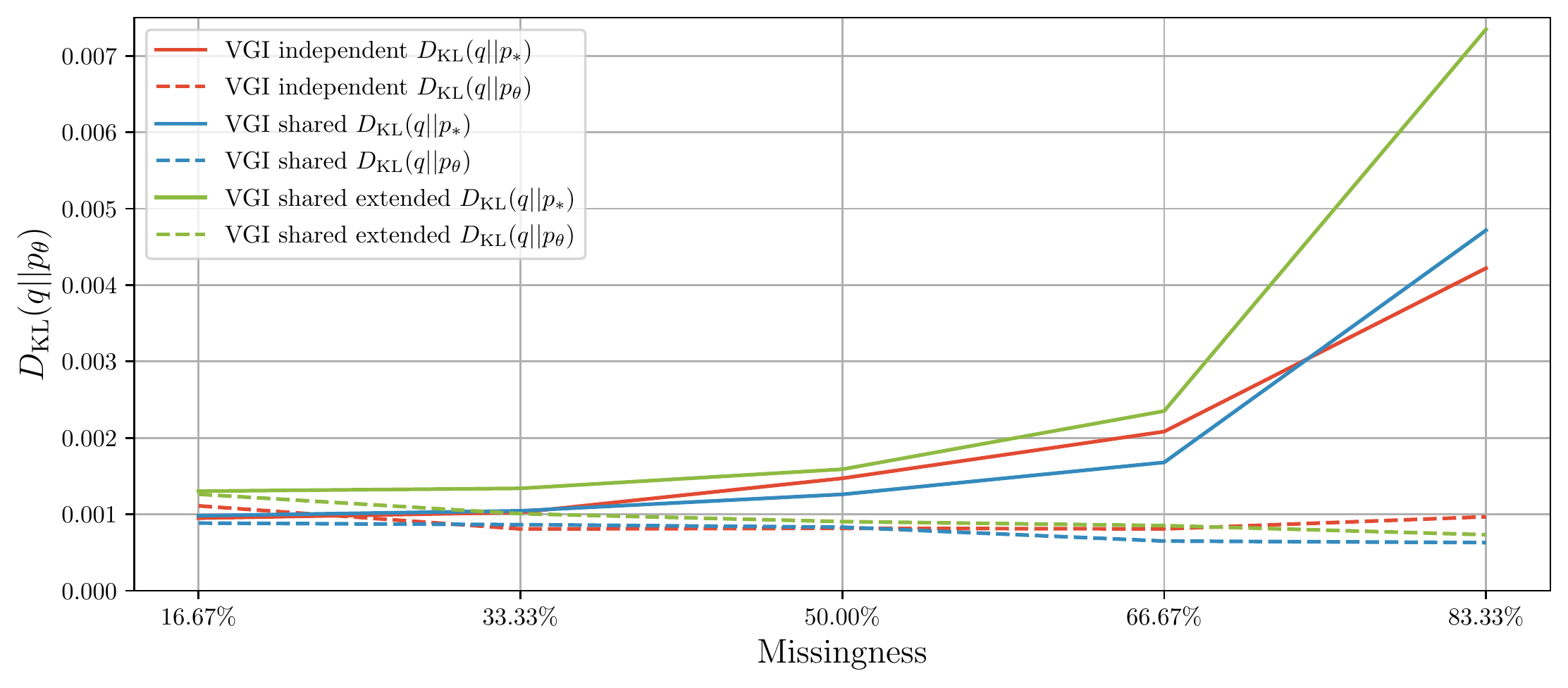}
    \caption{Median KL divergence of the univariate variational conditional distributions, conditioned on the toy test set. Comparing independent against shared-parameter variational models.}
    \label{fig:toy-posterior-kldiv-individual-vs-shared}
\end{figure}

The imputation accuracy depends on the fit of the variational conditionals, which subsequently influences the accuracy of the fitted model.
Hence, we evaluate, and show in Figure~\ref{fig:toy-posterior-kldiv}, the quality of the fitted variational conditional distributions using the KL divergence between the learnt conditionals and the ground truth distribution $\KLD{\qp(x_j \mid \xnj)}{\ps(x_j \mid \xnj)}$ in blue and the conditionals of the learnt statistical model $\KLD{\qp(x_j \mid \xnj)}{\pt(x_j \mid \xnj)}$ in red, where the conditional distributions are computed on test data.\footnote{For the toy data, we also show results dissected by dimension in  Appendix~\ref{apx:additional-figures}, Figure~\ref{fig-apx:toy-posterior-kldiv-separate}.}

The KL divergence to the ground truth indicates how good the imputations sampled from the variational conditionals are. It can be seen that the approximation of the ground truth conditionals gets worse as the missingness increases, which is also in line with the accuracy of the target model in Figure~\ref{fig:toy-fitted-model-results}. This is expected since the variational conditionals approximate the conditional distribution of the target model $\pt$.

In contrast, the KL divergence to the learnt model $\pt$ shows how well the variational conditionals approximate the true conditional distribution under the fitted model, this is the objective that is minimised by $\JVGI$ so we expect it to be low. 
We observe that the variational conditionals are good at the majority of test data-points (see red curves), however we see a long tail where the variational approximations are poor (see the tail of the red violin plot). This under-performance of the variational conditionals without fine-tuning corresponds to the approximation gap discussed in Section~\ref{sec:vgi-eval-held-out}.
Hence, when performing Gibbs sampling with the variational conditionals \revisiontwo{for model selection}, some light fine-tuning of the variational model might be necessary in order to prevent divergent Gibbs imputation chains, as discussed in Section~\ref{sec:vgi-eval-held-out}.

Furthermore, we see that the median KL divergence to the conditionals of the fitted model remains constant for all fractions of missingness on the toy data, but it goes down with increasing missingness on the FA-Frey data (solid red lines).
We attribute this to the partial weight sharing used in the variational model on FA-Frey data. 
We verify this by additionally comparing an independent model (red) against a shared, but not extended, model (blue) on the toy data in Figure~\ref{fig:toy-posterior-kldiv-individual-vs-shared}.\footnote{We also compare the accuracy of the fitted FA models with both variational models on toy data in Appendix~\ref{apx:additional-figures} Figure~\ref{fig-apx:toy-fitted-model-results-individual-vs-shared}, where we find both approaches comparable in the model $\pt$ accuracy.} We can see that with the partially-shared model the KL divergence on toy data also improves with missingness (dashed blue line), whilst it remains almost constant for the independent model (dashed red line).
The improved approximation of the conditionals with increasing missingness and partially-shared variational models can be explained by additional amortisation---it is akin to having more data to learn the same number of parameters, which results in a better fit of the variational model.

We further investigated the effect of the extended variational model on the conditional distributions using the toy data (Section~\ref{sec:variational-model}):
In Figure~\ref{fig:toy-posterior-kldiv-individual-vs-shared} we compare a shared variational model with the extended conditionals (green) to models with standard conditionals (red and blue). 
We observe that the extended conditionals are close approximations of the true Gibbs conditionals, and only slightly worse than the standard variational conditionals. Moreover, the estimate of the model $\pt$ is only slightly affected for missingness larger than $66\%$ (see Figure~\ref{fig-apx:toy-fitted-model-results-individual-vs-shared} in Appendix~\ref{apx:additional-figures}). 
Hence, we conclude that conditioning on an additional variable $\txj$ in the extended-Gibbs variational model (Appendix~\ref{apx:vgi-derivation-extended}) results in inconsequential additional variability (``noise'') of the variational distributions.
Importantly, in higher dimensional problems, the noise due to conditioning on a single extra variable decreases. There is no significant difference between VGI estimation accuracy with the standard variational conditionals and the extended conditionals when the data dimensionality is larger (see Figure~\ref{fig-apx:fa-time-and-kldiv-relaxed-vs-not-relaxed} in Appendix~\ref{apx:additional-figures}).
Hence, when the dimensionality is small and the computational cost is low it is best to use the standard variational conditionals, however, when the dimensionality is large, it is more favourable to use the extended variational model for computational reasons.

\section{Experiments on VAE Models}
\label{sec:vae-experiments}

In this section we estimate VAE models from incomplete data. We compare the general-purpose VGI method, VAE-specific methods based on VBGI, and existing VAE-specific methods in the literature in terms of estimation accuracy and computational efficiency.

\subsection{Variational Autoencoder Model}

The variational autoencoder (VAE) is a non-linear descendant of the factor analysis model. 
It is a latent-variable model with observables $\x$ and latent variables $\xz$, defined via
\begin{align*}
    \pt(\x \mid \xz) = \N(\x; \mub_\thetab(\xz), \Psi_\thetab(\xz)), \quad \text{ and } \quad p(\xz) = \N(\xz ; \bm{0}, \bm{I}),
\end{align*}
where the $\mub_\thetab$ and $\Psi_\thetab$ are deep neural networks with parameters $\thetab$, and $\Psi_\thetab(\xz)$ is diagonal. We use the Gaussian family for the generator, which is the most common case in the VAE literature, although another family of distributions could also be used.
The model is typically optimised using variational inference \citep{Kingma2013, Rezende2014a}, where a variational distribution $\qp(\xz \mid \x)$ is used to approximate the intractable posterior $\pt(\xz \mid \x)$. 
The parameters $\thetab$ and $\phib$ are then optimised with stochastic gradient ascent by maximising the following lower-bound
\begin{align}
    \log \pt(\x) \geq \E_{\qp(\xz \mid \x)} \left[ \log \frac{\pt(\x \mid \xz) p(\xz)}{\qp(\xz \mid \x)} \right]. \label{eq:vae-elbo}
\end{align}

\subsection{Data}
\label{sec:vae-data}
We evaluate VGI on a synthetic 560-dimensional VAE-Frey data set, where a VAE model with one hidden layer in the encoder and decoder and a 10-dimensional latent space was first fitted on the Frey data set, and then it was used to synthesise a new data set that we call VAE-Frey. We used the same VAE architecture as \citet{Kingma2013}. The training data set has 2400 data-points and the test data set has 3000 data-points.
The rest of the data setup is identical to Section~\ref{sec:fa-data}.

\subsection{Experimental Settings}
\label{sec:vae-experimental-settings}

As before, we specify a target VAE model to fit data following a ground truth VAE model with the same architecture, so that the evaluation can focus on the effect of model estimation from incomplete data, rather than on the robustness of the specified model.

We consider three VGI-based methods for fitting the VAE model:

    \paragraphJMLR{VGI.} This method does not use any assumptions of the statistical model $\pt$ and hence uses the univariate variational conditional formulation of VGI, as before. We specify the variational encoder architecture to be equivalent to the encoder of the ground truth and replace the $\log \pt(\x)$ in $\JVGI$ in \eqref{eq:vgi-objective} with the VAE ELBO from \eqref{eq:vae-elbo}. 
    
    We use the same univariate variational model with extended conditionals and partially-shared parameters as in the FA-Frey data experiments in Section~\ref{sec:fa-vgi-settings} and optimise the variational parameters using AMSGrad \citep{Reddi2018}.
    The other VGI hyper-parameters are also equivalent to those used in the FA-Frey data experiments: $K=5$ imputation chains, $G=5$ Gibbs update steps, and $M=10$ sampled missing dimensions in $\hJVGI$.
    
    \paragraphJMLR{\VBGIVAE.} Here we use the structure and assumptions of a VAE to adapt the VGI method to the model $\pt$. 
    
    First, VAE is a latent-variable model so we can treat the latent variables as missing and group them together as we discussed in Section~\ref{sec:vbgi}. We then specify a joint variational distribution $q_{\phib_z}(\xz \mid \txm, \xo)$, which corresponds to the encoder of the standard VAE. 
    
    Second, the missing variables $\xm$ are assumed to be conditionally independent Gaussians given the latents $\xz$, hence we can easily specify the joint distribution of the conditionals $\qpj(\xj \mid \xmnj, \txz, \xo), j \in \idx{\m}$ using a shared variational model $q_{\phib_\mis}(\xm \mid \txz, \xo)$ for all patterns of missingness, similar to the target generative model. To parametrise $q_{\phib_\mis}$ using a neural network, we pad the $\xo$ with zeros for the missing dimensions to get a fixed-size vector for all patterns of missingness.

    Whilst the VAE model assumes that $\xm$ and $\xo$ are independent given $\xz$, in the variational conditionals $q_{\phib_\mis}$ we do not simplify the conditioning set but condition on $\xo$ as in the general VGI formulation.
    Conditioning on $\xo$ compensates for possible information loss about the true $\xz^* \sim \pt(\xz \mid \xo)$ due to the use of an approximation $q_{\phib_z}$, and hence eases the learning of $q_{\phib_\mis}$.
    We empirically validated this theoretical argument against a simplification of the variational conditional and observed that conditioning $q_{\phib_\mis}$ on $\xo$, as dictated by the general VGI methodology, is indeed crucial to the performance of the method. 
    
    Then, the index $j \in \{ \mis, z\}$ in the VGI objective refers to either all the missing variables $\xm$ or the latents $\xz$. 
    Incorporating the VAE assumptions into the VBGI objective for latent-variable models from \eqref{eq:vbgi-latent-objective} we get the VAE-specific objective
\begin{align*}
    \J^t_\text{\VBGIVAE}(\thetab, \phib; \xo) &= \E_{f^{t-1}(\txm, \txz \mid \xo)} \Bigg(\\ 
    &\phantom{= }
    \pib(j=\mis) \E_{q_{\phib_\mis}(\xm \mid \txz, \xo) } \left[ \log \frac{\pt(\xm, \xo \mid \txz)p(\txz)}{q_{\phib_\mis}(\xm \mid \txz, \xo)} \right]\\
    &\phantom{= }+ \pib(j=z) \E_{q_{\phib_z}(\xz \mid \txm, \xo) } \left[ \log \frac{\pt(\txm, \xo \mid \xz)p(\xz)}{q_{\phib_z}(\xz \mid \txm, \xo)} \right]
    \Bigg),
\end{align*}
    where $\pib(j) = \frac{1}{2}$ for all $j$, meaning that we choose a uniform Gibbs selection probability.
    We use $K=5$ imputation chains and $G=2$ Gibbs updates corresponding to one full update of all unobserved variables $\xm$ and $\xz$, and optimise the above objective using the Adam optimiser.
    
    \paragraphJMLR{\VBGIVAEM.} Alternatively, we can use the conditional independence of the observable variables given the latents for the VAE model to marginalise the missing observable dimensions $\xm$ from the likelihood, which is commonly done in the VAE literature for incomplete data. Then, $j=\mis$ and the objective above simplifies to
\begin{align*}
    \J^t_\text{\VBGIVAEM}(\thetab, \phib; \xo ) &= \E_{f^{t-1}(\xm \mid \xo) q_{\phib_z}(\xz \mid \xm, \xo)} \left[ \log \frac{\pt(\xo \mid \xz)p(\xz)}{q_{\phib_z}(\xz \mid \xm, \xo)} \right]
    \Bigg).
\end{align*}
    To update the Markov chain distribution $f^t$ we perform block-Gibbs sampling using $q_{\phib_z}(\xz \mid \xm, \xo)$ for the latents and $\pt(\xm \mid \xz)$ for the missing observables. 
    This imputation procedure was initially proposed by \citet{Rezende2014a} for missing data imputation on test data, however to the best of our knowledge it has not been considered for learning from incomplete data.
    We use $K=5$ imputation chains and $G=2$ Gibbs updates corresponding to one full update of all unobserved variables $\xm$ and $\xz$, and optimise the above objective using the Adam optimiser.

In all of the above methods we use the reparametrisation trick \citep{Kingma2013} to compute the gradients of the variational models.

\subsection{Comparison Methods}
\label{sec:vae-comparison-methods}

We compare the VGI-based methods against:

\begin{description}
    \item[VAE (Complete).] The ideal case where no data is missing, fitted using standard variational inference for VAEs \citep{Kingma2013, Rezende2014a}.
    \item[MICE \citep{VanBuuren2000}.] The same as in Section~\ref{sec:fa-comparison-methods}.
    \revisiontwo{
    \item[missForest \citep{stekhovenMissForestNonparametricMissing2012}.] An iterative FCS method that uses random forest to model the conditionals. Unlike MICE, the imputations are not sampled but instead are replaced with the deterministic values predicted by the random forest. Hence, while the method can be used on non-linear data, the imputations may be biased by lack of uncertainty representation. We implemented the method in Python using IterativeImputer and RandomForestRegressor for the conditional models from the scikit-learn package \citep{Pedregosa2011}. 
    The default settings for this method proved to be too computationally expensive on this data set, hence to make this method computationally feasible we traded-off the expressivity of the random forest for better computational performance by limiting the number of decision trees to 20 and the maximum depth of the trees to 20.\footnote{One imputation of the data set with missForest took about 27 hours on CPU with the scikit-learn implementation.}
    Random forest methods are typically piecewise-constant regressors, which means that by limiting the depth and the number of trees, for reasons above, we may get a too coarse an approximation of the target variable, on the other hand, if the computational budget allows, the trees could be ``fully grown'' such that each leave corresponds to a unique training data point and hence, in the idealistic scenario of a finely sampled training set, the random forest regressor could be made arbitrarily accurate.
    Like MICE, missForest was used to produce $K=5$ imputations and then the VAE model was fitted as described in Appendix~\ref{apx:combining-mi}.
    }
    \revisiontwo{
    \item[MICEForest.] One classical way to calibrate the uncertainty of imputations for methods like missForest is predictive mean matching \citetext{\citealp{littleMissingDataAdjustmentsLarge1988}; \citealp[][Chapter~3.4]{VanBuuren2018}}. We hence use the MICEForest Python package that provides an implementation of missForest enhanced with predictive mean matching.\footnote{MICEForest implementation can be found here \url{https://github.com/AnotherSamWilson/miceforest}.} The uncertainty is controlled via a hyperparameter that specifies the number of nearest neighbours (based on the predictive mean) considered for a random imputation draw.
    As with missForest above, to make this method feasible on the high-dimensional data in this section, the number of estimators was set to 20, the maximum depth of the decision trees was set to 20, and the number of nearest neighbours was set to the default 5.\footnote{One imputation of the data set with MICEForest took about 17 hours on CPU.}
    }
    \item[MVAE \citep{Nazabal2018, Mattei2019}.] The missing dimensions are marginalised in the generator of the VAE and the encoder masks the missing dimensions with zeros.
    \item[HIVAE \citep{Nazabal2018}.] The missing dimensions are marginalised in the generator of the VAE. 
    The model uses a hierarchical prior and zero masking in the encoder for missing data.
    \item[PartialVAE+ \citep{Ma2019}.] The missing dimensions are marginalised in the generator of the VAE. Uses a permutation-invariant encoder network with learnable location embeddings, which can handle partially-observed inputs.
    \item[MIWAE \citep{Mattei2019}.] The missing dimensions are marginalised in the generator of the VAE. Uses a tighter importance weighted lower-bound and zero masking in the encoder.
\end{description}
\noindent The above methods are all fitted using the Adam optimiser.

\subsection{Accuracy of the Fitted VAE Models}
\label{sec:vae-accuracy}

As is common in machine learning the accuracy of the VAE models increased with training iterations and then decreased due to over-fitting, hence we used model selection on an incomplete held-out validation data set to mitigate this effect (for more details see Appendix~\ref{apx:vae-model-selection} and Figure~\ref{fig-apx:vae-finetuning-curves}).
We then use the selected model parameters $\hat \thetab$ to assess the accuracy of the fitted model.

In order to evaluate the accuracy of the selected VAE models, we estimated the log-likelihood on complete test data. In general, computing the marginal log-likelihood of a VAE is intractable, therefore we estimate it using the IWAE bound \citep{Burda2015} that uses samples from \revisiontwo{a variational proposal distribution $q$} and self-importance weighting,
\begin{align*}
    \mathcal{L}_{\text{IWAE}}(\x) &= \log \left( \frac{1}{L} \sum_{l=1}^L \frac{p_{\hat \thetab}(\x, \xz^l)}{q(\xz^l \mid \x)} \right), \quad \text{ where } \quad \xz^1, \ldots, \xz^L \sim q(\xz \mid \x).
\end{align*}
The IWAE bound approaches $\log p_{\hat \thetab}(\x)$ monotonically from below as the number of importance samples $L$ increases \citep[Theorem 1]{Burda2015} 
\revisiontwo{irrespective of the proposal distribution (subject to minor conditions).
However, we note that the non-asymptotic bias of the estimator is in the order of $\mathcal{O}(1/L)$ \citep{owenMonteCarloTheory2013, paananenImplicitlyAdaptiveImportance2021} and depends on the accuracy of the proposal distribution $q(\xz \mid \x)$, and is zero if $q(\xz \mid \x)=\pt(\xz \mid \x)$.
Hence, to attain a good proposal and mitigate the bias we fine-tune the variational encoder $\qp$ from the training stage on the complete test data \citep{matteiRefitYourEncoder2018}.\footnote{\revisiontwo{We have also considered fitting a randomly-initialised encoder on the complete test data but due to poor initialisation the fitted variational approximations were worse \citep[also observed in previous works \eg][]{altosaarProximityVariationalInference2017}.}}
We then estimate the log-likelihood using $\mathcal{L}_{\text{IWAE}}$ and the fine-tuned encoder with a large number $L=50000$ of importance samples.%
}

\begin{figure}[tb]
    \includegraphics[width=1\linewidth]{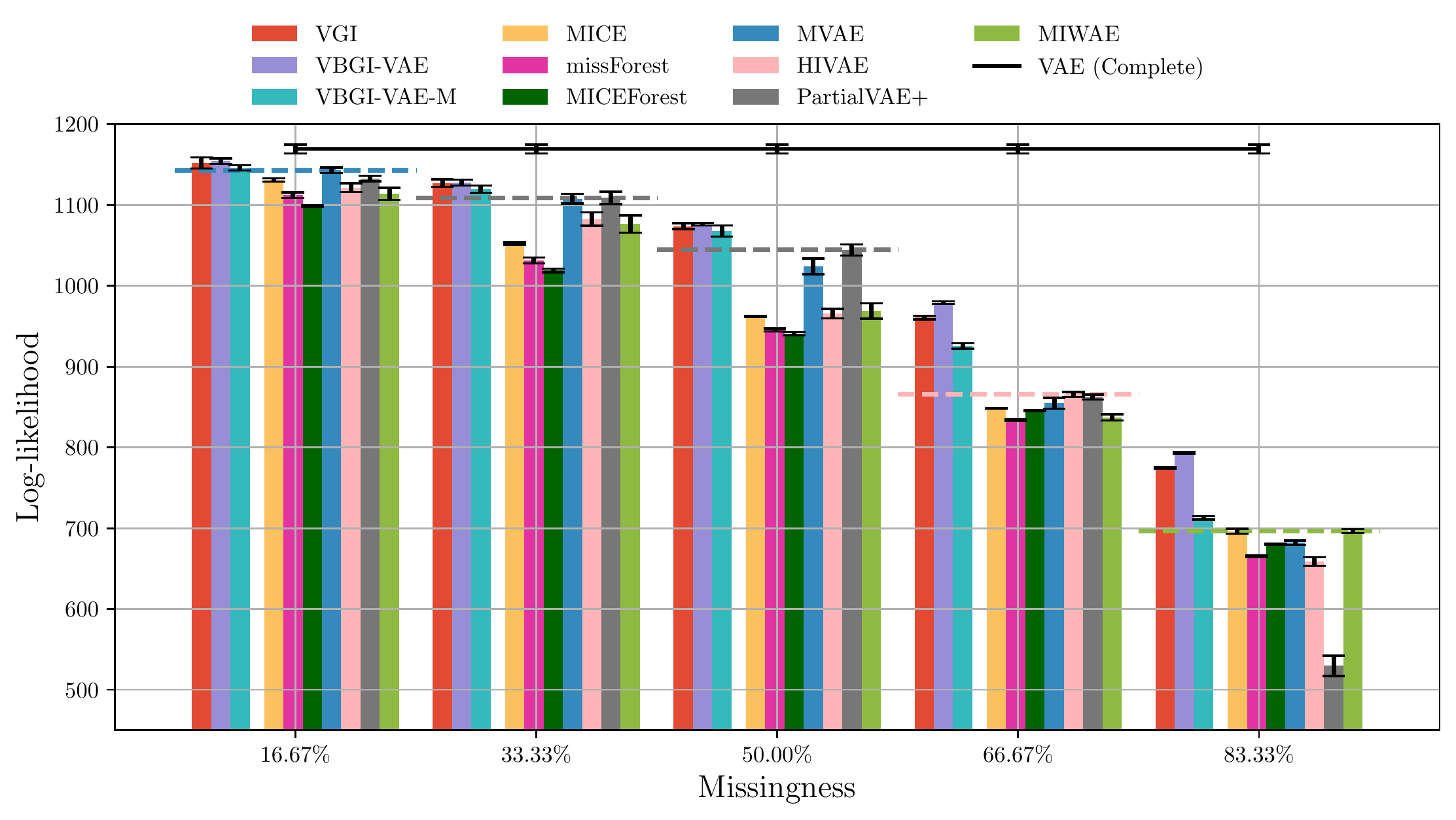}
    \caption{Importance weighted estimates of the marginal log-likelihood on \revisiontwo{complete} test data. \revisiontwo{The error bars show standard error on the mean of five experiment repetitions with different missingness masks and model initialisation.} The dashed horizontal lines show the best performing competitor model, which we note is always outperformed by the VGI-based methods.}
    \label{fig:fcvae-fitted-model-results}
\end{figure}

In Figure~\ref{fig:fcvae-fitted-model-results} we show the estimated marginal log-likelihood averaged over the test data. We see that the VGI-based methods perform consistently better than the existing VAE-specific methods, particularly in settings with missingness fraction greater than 50\%. 
It is interesting that VGI and \VBGIVAE, which learn a variational imputation model of the missing observable variables, both outperform \VBGIVAEM, which marginalises the missing variables from the likelihood.
This suggests that forcing the VAE model to reconstruct a completed data-point is overall better for the accuracy of the generator than marginalising the missing variable dimensions in the generator as it is often done in the VAE-specific methods and VGI-VAE-M. 
We conjecture that marginalising the missing variable dimensions allows the latent representation in VAEs to systematically forget (collapse) some of the information presented in the inputs. In contrast, forcing the VAE model to reconstruct a completed data-point provides an adaptive regularisation that prevents such loss of information.

The learning curves for the different methods are shown in Figure~\ref{fig-apx:vae-learning-curves} in Appendix~\ref{apx:additional-figures}. 
We note that for missingness fractions greater than 50\%, 
the methods that marginalise the missing dimensions from the likelihood (VAE-specific and \VBGIVAEM)
start to over-fit on the training data (compare solid and dash-dotted curves) early,
whereas the VGI and \VBGIVAE monotonically improve throughout the training, which allows them to surpass the other methods in terms of the model accuracy, especially at greater than $50\%$ missingness, which we also observe in Figure~\ref{fig:fcvae-fitted-model-results}.

Moreover, we show in Figure~\ref{fig-apx:vae-time-and-loglik-relaxed-vs-not-relaxed} in Appendix~\ref{apx:additional-figures} that working with the extended variational model in VGI had no adverse effect on the estimation accuracy while greatly reducing the computational cost. This adds additional evidence to the results in Section~\ref{sec:fa-accuracy-variational-model}.

Regarding the \revisiontwo{impute-then-fit} methods, we notice that MICE performs well despite being a linear method. This can be explained as follows: the generator is only slightly non-linear since it is parametrised by a one-hidden layer neural network and the data is based on the Frey data set, which can be fitted rather well by linear methods, such as factor analysis.
\revisiontwo{
Figure~\ref{fig:fcvae-fitted-model-results} shows that the non-linear imputation methods, missForest or MICEForest, did not perform well in this experiment.
We attribute the poor performance of these methods to two factors: modelling limitations due to computational cost restrictions, see Section~\ref{sec:vae-comparison-methods}, and, possibly, an incorrect representation of imputation uncertainty. 
If the computational budget allows, the former issue can be addressed by using more or larger decision trees although preventing overfitting will then become important. %
The second issue could be addressed in MICEForest by tuning the number of nearest neighbours in predictive mean matching, although current recommendations are more based on heuristics than firm principles \citep[\eg][Chapter~3.4]{VanBuuren2018}, which can make finding an optimal setting difficult.
}

Finally, we observe that MVAE performs almost always better than MIWAE, even though MIWAE optimises a tighter importance-weighted lower-bound. 
This suggests that when the statistical model $\pt$ matches the true generative model but, due to missingness, we do not have enough data to actually fit the model, MIWAE might be more prone to over-fitting since it can tighten the bound more effectively. To confirm this intuition, we fitted an under-specified model to the incomplete data (by reducing the latent dimension to 5), and in line with our explanation above, MIWAE was then performing better than MVAE.
Moreover, PartialVAE+ performed considerably well on missingness levels up to 83.33\%, however the performance went down significantly at the largest missingness which \revisiontwo{points to a} difficulty of approximating the variational distribution \revisiontwo{with} a permutation-invariant encoder at very large missingness.

\subsection{Computational Aspects}
\label{sec:vae-computational-aspects}

\begin{figure}[tbp]
    \includegraphics[width=1\linewidth]{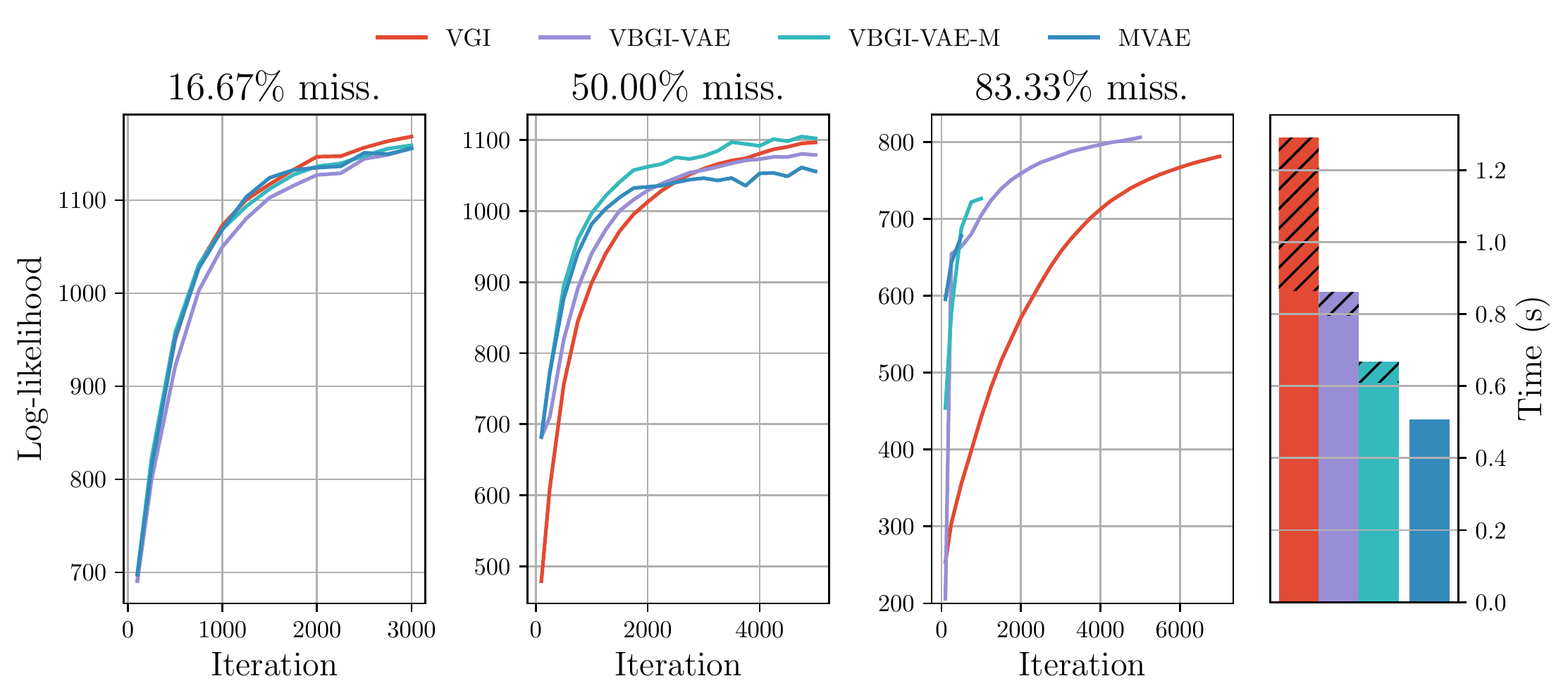}
    \caption{Left: Estimated test log-likelihood on complete data against training iteration. Right: Average duration of one training iteration in seconds. The hatched part of the bars indicates the time spent on updating the imputations via pseudo-Gibbs sampling. Plots including other fractions of missingness and the other VAE-specific methods can be found in the Appendix~\ref{apx:additional-figures} Figures~\ref{fig-apx:fcvae-time-per-epoch} and \ref{fig-apx:fcvae-test-loglik-vs-epoch-all}.}
    \label{fig:fcvae-test-loglik-vs-epoch-and-time-per-epoch}
\end{figure}

We here discuss the computational aspects of VGI-based methods for VAE estimation, and contrast it with the VAE-specific methods. 

In the left-hand side of Figure~\ref{fig:fcvae-test-loglik-vs-epoch-and-time-per-epoch} we plot the estimated test log-likelihood against the training iteration and compare the VGI-based methods to MVAE. We see that out of all VGI-based methods \VBGIVAEM has the highest rate of likelihood improvement, which is comparable to the rate of MVAE, while achieving consistently better accuracy of the fit. 
However, the methods that marginalise the missing variables from the likelihood (MVAE and \VBGIVAEM) quickly over-fit to the training data in the higher missingness settings and hence perform worse than VGI and \VBGIVAE.
\VBGIVAE closely follows \VBGIVAEM at a slightly slower rate of improvement, due to having to learn an additional variational model of the missing observables, but it offers a generally better accuracy of the fit, as seen in the previous section, and does not over-fit to the training data like MVAE and \VBGIVAEM.
The general-purpose VGI method, which does not use the assumptions available in a VAE, has the slowest rate of improvement, which is most clearly seen in high missingness settings. However, as \VBGIVAE, VGI generally outperforms the other methods that marginalise the missing variables from the likelihood in terms of the accuracy of the fit.

The average duration of a training iteration is shown in the right-hand side of Figure~\ref{fig:fcvae-test-loglik-vs-epoch-and-time-per-epoch}. 
We used the same VGI hyper-parameters for all levels of missingness in our experiments, which means that the average per-iteration cost is about the same. While the cost of an iteration for VAE-specific methods does not depend on the missingness, for VGI-based methods, the iteration cost could be adjusted to the level of missingness by, for example, adapting the number of Gibbs update steps ($G$ in Algorithm~\ref{alg:pseudo-Gibbs}), the number of imputation chains $K$, or the number of sampled missing dimensions ($M$ in \eqref{eq:vgi-objective-approx}).
The figure thus shows results that were not optimised for computational efficiency.
The average duration of an iteration of the VGI-based methods are about 2.4 (VGI), 1.6 (\VBGIVAE), and 1.25 (\VBGIVAEM) times longer than that of the other methods.
Moreover, about 0.6 (VGI), 0.2 (\VBGIVAE), and 0.45 (\VBGIVAEM) of the excess iteration cost over the other methods is due to the Gibbs sampling of imputations, denoted by the hatched bars in the figure.
We attribute the rest of the excess cost to the fitting of the variational kernel and evaluating the generative model on multiple imputed samples.
We also note that for the experiments in this section we used a shared variational model with extended variational conditionals as discussed in Section~\ref{sec:variational-model} which resulted in about 2.8 times lower per-iteration cost than VGI with the standard-Gibbs conditionals (see Appendix~\ref{apx:additional-figures} Figure~\ref{fig-apx:vae-time-and-loglik-relaxed-vs-not-relaxed}).

In summary, VGI-based methods offer better estimation accuracy over the other methods at (i) a slightly higher per-iteration computational cost and (ii) a potentially slower rate of likelihood improvement in the high-missingness settings.

\section{Experiments on Normalising Flows}
\label{sec:flow-experiments}

In this section we estimate normalising flow models from incomplete UCI data using the general-purpose VGI and a flow-specific Monte Carlo EM method. We compare them in terms of model accuracy and computational efficiency.

\subsection{Normalising Flow Model}

A normalising flow is a probabilistic model that models complex distributions $\pt(\x)$ via a simple base distribution $p(\ub)$ and an invertible deterministic transformation $T_\thetab{\,:\,} \ub \mapsto \x =T_\thetab(\ub)$ \citep{Rezende2015}. The density of $\x$ can be obtained from the change of variables formula \citep[\eg][]{murphyProbabilisticMachineLearning2021} as
\begin{align*}
    \pt(\x) = p(\ub) |\det{J_{T_{\thetab}}(\ub)}|^{-1}, \quad \text{where} \quad \ub = T_{\thetab}^{-1}(\x),
\end{align*}
and $J_{T_\thetab}$ is the Jacobian matrix of the transformation.
In practice, the transformation $T_\thetab$ is often composed of multiple chained transformations $T_\thetab = T_L \circ \cdots \circ T_1$, where we suppress the dependency of the decomposed transformations $T_l$ on $\thetab$ for notational simplicity. The parameters of each transformation are given by a neural network, thus enabling complex overall transformations $T_\thetab$. Moreover, both transformation $T_\thetab$ and $T_\thetab^{-1}$ must be differentiable, which allows us to fit the parameters $\thetab$ using stochastic gradient ascent.

The expressivity of the flow model greatly depends on the choice of the transformations $T_l$, hence to show that our method can fit expressive models, we specify the statistical model $\pt$ as a rational-quadratic neural spline flow \citep[RQ-NSF,][]{durkanNeuralSplineFlows2019}, which has been shown to be able to model complex distributions efficiently. 
In RQ-NSF, each transformation $T_l$ is composed of an invertible linear transformation and a monotonic rational-quadratic spline transformation.
The spline transformation maps a fixed interval $[-B, B]$ to $[-B, B]$ using a spline function and an identity mapping outside of this range. 
The rational-quadratic splines are monotonic piece-wise functions with $P$ sections (\emph{bins}) that are characterised by the boundary coordinates (\emph{knots}) and the (positive) derivatives at the knots, where both the knot coordinates and the derivative values are parametrised by a residual neural network.

\subsection{Data}

We evaluate VGI on a selection of tabular data sets from the UCI machine-learning repository \citep{duaUCIMachineLearning2017}, which are commonly used to evaluate normalising flow models, and follow the pre-processing in \citep{papamakariosMaskedAutoregressiveFlow2017}.

\begin{description}
    \item[POWER.] Measurements of electric power consumption in one household with a one-minute sampling rate collected over a period of 47 months. Contains 6 dimensions and \textapproxVGI 2M samples.
    \item[GAS.] A collection of gas sensor measurements in several gas mixtures. Contains 8 dimensions and \textapproxVGI 1M samples.
    \item[HEPMASS.] A collection of measurements from high-energy physics experiments to detect a new particle of unknown mass. Contains 21 dimensions and \textapproxVGI 525K samples.
    \item[MINIBOONE.] Data taken from a MiniBooNE experiment used to distinguish electron neutrinos from background noise. Contains 43 dimensions and \textapproxVGI 36K samples.
\end{description}

\noindent The missingness is rendered uniformly at random (MCAR) at three levels of missingness from 16.6\% to 83.3\%. The rest of the data setup is the same as in the previous sections.

\subsection{Experimental Settings}

We implement the rational-quadratic neural spline flow (RQ-NSF) with coupling following \citet[][Appendix~B.1]{durkanNeuralSplineFlows2019}.

In order to match the expressiveness of the target model $\pt$ we parametrise the univariate conditional distributions using flow-like element-wise distributions. 
We use the standard normal distribution as the base distribution and $R$ sets of linear and rational-quadratic spline transformations, where the parameters of the spline transformations (knot coordinates and derivatives) are given by a partially-shared residual network.
The transformations are defined over the same interval as the target flow model, and we use 4 bins and $R=3$ element-wise transformations.

As before, we use the extended conditionals with a partially-shared model which consists of a shared network and independent per-conditional networks. The shared network takes a completed data-point and computes a shared representation. This representation is then used to compute the element-wise transformation parameters using independent networks for each of the $R*D$ transformations.
The shared network consists of one residual block with 256 hidden features, and outputs a 128-dimensional shared representation. Each element-wise transformation network consists of 2 residual blocks with 32 hidden features.

Like in the experiments in the previous sections, we optimise the variational parameters $\phib$ using AMSGrad \citep{Reddi2018}.
The other VGI hyper-parameters are as follows: $K=5/10/20$ imputation chains for corresponding missingness of 16.6\%/50\%/83.3\%, $G=5$ Gibbs update steps, and $M=1$ sampled missing dimensions in \eqref{eq:vgi-objective-approx}.

\subsection{Comparison Methods}
\label{sec:flows-comparison-methods}

We compare VGI against Monte Carlo expectation-maximisation \citep[MCEM,][]{Wei1990} with the flow-specific MCMC sampler of \citet{cannellaProjectedLatentMarkov2021}.
MCEM maximises the observed data log-likelihood by iteratively maximising the ELBO
\begin{align*}
    \hat \thetab^t &= \argmax_\thetab \frac{1}{N} \sum_{i=1}^N \E_{p_{\hat \thetab^{t-1}}(\xm \mid \xo^i)} \left[ \log \pt(\xm, \xo^i) - \log p_{\hat \thetab^{t-1}}(\xm \mid \xo^i) \right]\\
    &= \argmax_\thetab \frac{1}{N} \sum_{i=1}^N \E_{p_{\hat \thetab^{t-1}}(\xm \mid \xo^i)} \left[ \log \pt(\xm, \xo^i) \right]\\
    &\approx \argmax_\thetab \frac{1}{N} \sum_{i=1}^N \frac{1}{K} \sum_{k=1}^K \log \pt(\xm^\sik, \xo^i), \quad \text{where} \quad \xm^\sik \sim p_{\hat \thetab^{t-1}}(\xm \mid \xo^i).
\end{align*}
We obtain the samples from the conditional distributions of missing data $p_{\hat \thetab^{t-1}}(\xm \mid \xo^i)$ using projected latent Markov chain Monte Carlo \citep[PLMCMC,][]{cannellaProjectedLatentMarkov2021}.
Rather than generating proposals in the data space, PLMCMC generates Metropolis-Hastings proposals \citep[\eg][Chapter~11.2.2]{Bishop2006} in the better-behaved space of the base distribution of a normalising flow using simple proposal distributions, and hence provides a more efficient way to sample from any conditional distribution under a joint normalising flow model. 

The use of PLMCMC with MCEM (in the next subsections denoted as PLMCMC for simplicity) for the estimation of normalising flows from incomplete data is not new to this paper and has been demonstrated by \citet{cannellaProjectedLatentMarkov2021}, and in our experiments we follow their experimental configuration. 
During the first set of iterations the imputations are sampled from a standard Gaussian distribution, and afterwards the conditional imputations are re-sampled using PLMCMC at regular intervals using chains of increasing length (from 200 to 1000 steps), except for the MINIBOONE data for which we could afford running longer chains of up to 3000 steps due to its relatively small size.

After a limited hyper-parameter search we have found that the same proposal distribution as used in the model estimation experiments by \citet{cannellaProjectedLatentMarkov2021} also performed best in our experiments.
The proposal distribution used is a uniform mixture of a conditional normal distribution with mean as the previous state and a standard deviation of 0.01, and an unconditional normal distribution with mean 0 and standard deviation of 1.0.

We observed that the PLMCMC sampler can experience numerical instability due to the inversion of the flow on the GAS and POWER data sets.
To stabilise the sampler on the GAS data set we clipped the accepted MCMC proposals to the observed data hypercube, defined by the minimum and maximum observed values for each dimension. A similar fix has been used in some experiments in the original work by \citet{cannellaProjectedLatentMarkov2021}. 
However, this fix did not help on the POWER data set, and the results in the following sections partly reflect the instability of the sampler.

\subsection{Accuracy of the Fitted Normalising Flow Models}
\label{sec:flow-accuracy}

We did not observe any over-fitting in the flow model experiments, hence the model parameters $\hat \thetab$ from the final training iteration were used in the following evaluation for both methods.

\begin{figure}[tb]
    \includegraphics[width=1\linewidth]{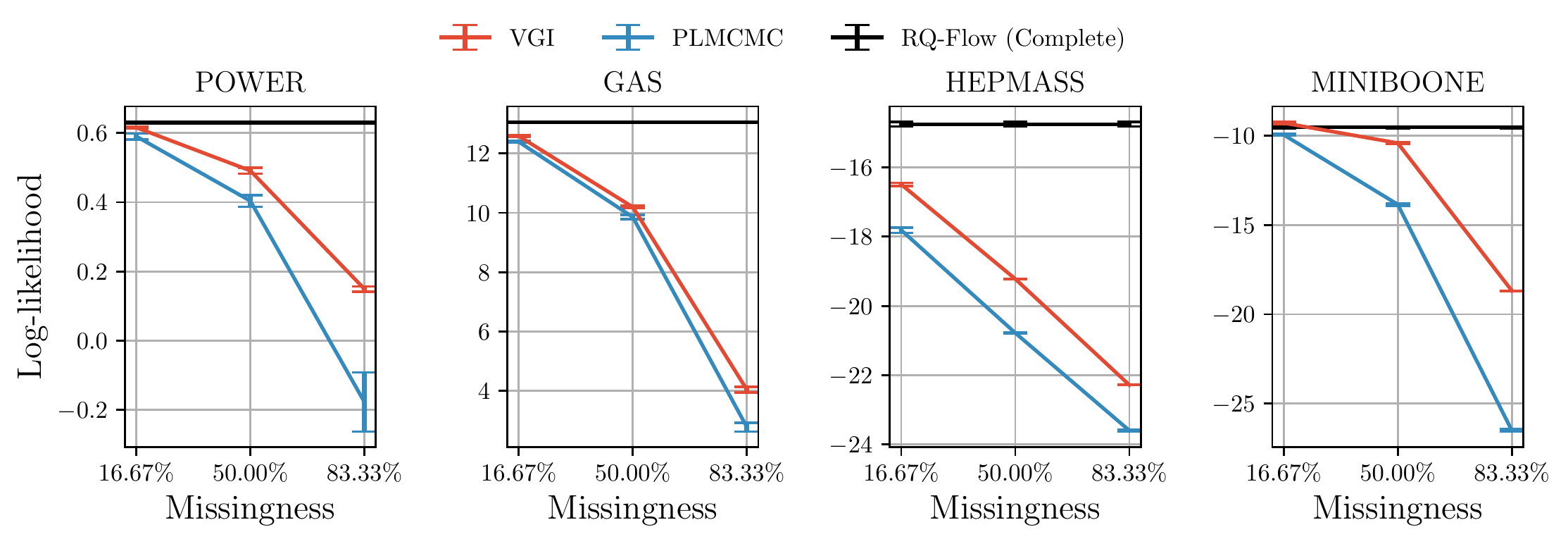}
    \caption{Log-likelihood on complete test UCI data. \revisiontwo{The error bars show standard error on the mean of five experiment repetitions with different missingness masks and model initialisation.} VGI consistently yields a more accurate model $\pt$ than MCEM with the PLMCMC sampler. %
    }
    \label{fig:rqcspline-fitted-model-results}
\end{figure}

We evaluate the accuracy of the fitted models using the log-likelihood on complete test data, shown in Figure~\ref{fig:rqcspline-fitted-model-results}. 
Uniformly across the data sets, VGI produces a more accurate fit than MCEM with the PLMCMC sampler. 
In principle, MCEM should be able to produce model fits that are equally good or better than any variational method due to sampling from the true conditional distributions instead of using an approximation.
However, in practice the performance of MCEM is limited by the performance of the MCMC sampler, which, due to computational constraints, often does not generate exact samples from the true conditional distribution. 
Even when using PLMCMC as the sampler, which takes advantage of the characteristics of a normalising flow to improve the performance of the sampler, the adverse effect of using finite-length Markov chains is significant.
In an attempt to mitigate this pitfall of MCEM, we have empirically investigated the use of persistent imputations chains, similar to VGI, but found that this drastically reduced the acceptance rate of the proposed transitions in the later training epochs and therefore also impaired the accuracy of the estimated statistical model.

In this evaluation we have trained the models for a fixed number of iterations with both methods---we investigate the accuracy-compute trade-off in the following subsection.
We thus note that the results could be further improved by running the methods for more iterations (see Appendix~\ref{apx:additional-figures} Figure~\ref{fig-apx:rqcspline-longrun} for possible gains when using larger compute budgets), or via other means in stochastic optimisation.%

\subsection{Computational Aspects}
\label{sec:flow-computational-ascpects}

\begin{figure}[p]
    \centering
    \vspace{-1.5em}
    \includegraphics[width=0.99\linewidth]{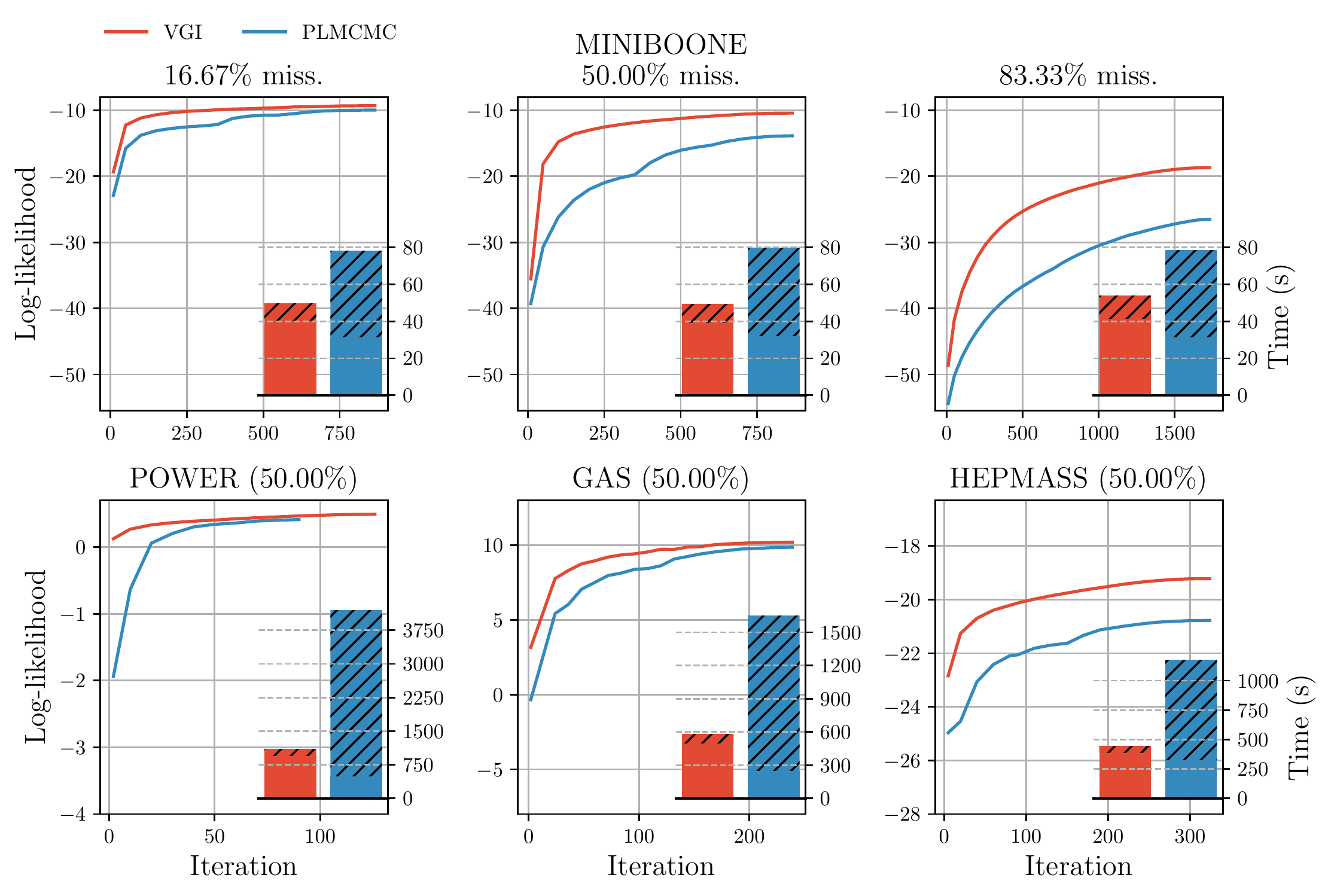}
    \vspace{-1.em}
    \caption{Estimated test log-likelihood on complete data against training iteration and average duration of one training iteration in seconds. The hatched part of the bars indicates the time spent on pseudo-Gibbs sampling in VGI and PLMCMC during MCEM.
    Top row: results on the MINIBOONE data set for all fractions of missingness. Bottom row: results on the other data sets (POWER, GAS, HEPMASS) for 50\% missingness (for the other fractions, see Appendix~\ref{apx:additional-figures} Figure~\ref{fig-apx:rqcspline-computational-aspects-all}).
    }
    \label{fig:rqcspline-computational-aspects}
\end{figure}

To investigate the computational performance, we plot the test log-likelihood against the training iteration in Figure~\ref{fig:rqcspline-computational-aspects} and further show the average duration of a training iteration for each method. 
First, we note that the log-likelihood curves of VGI (red) are consistently ahead of PLMCMC (blue). Hence, even without taking the iteration cost into account, VGI improves the statistical model $\pt$ significantly faster than PLMCMC.
Moreover, the average per-iteration cost of VGI is also smaller than PLMCMC as shown in the bar plots, making VGI overall more efficient than PLMCMC.

\begin{figure}[p]
    \includegraphics[width=0.99\linewidth]{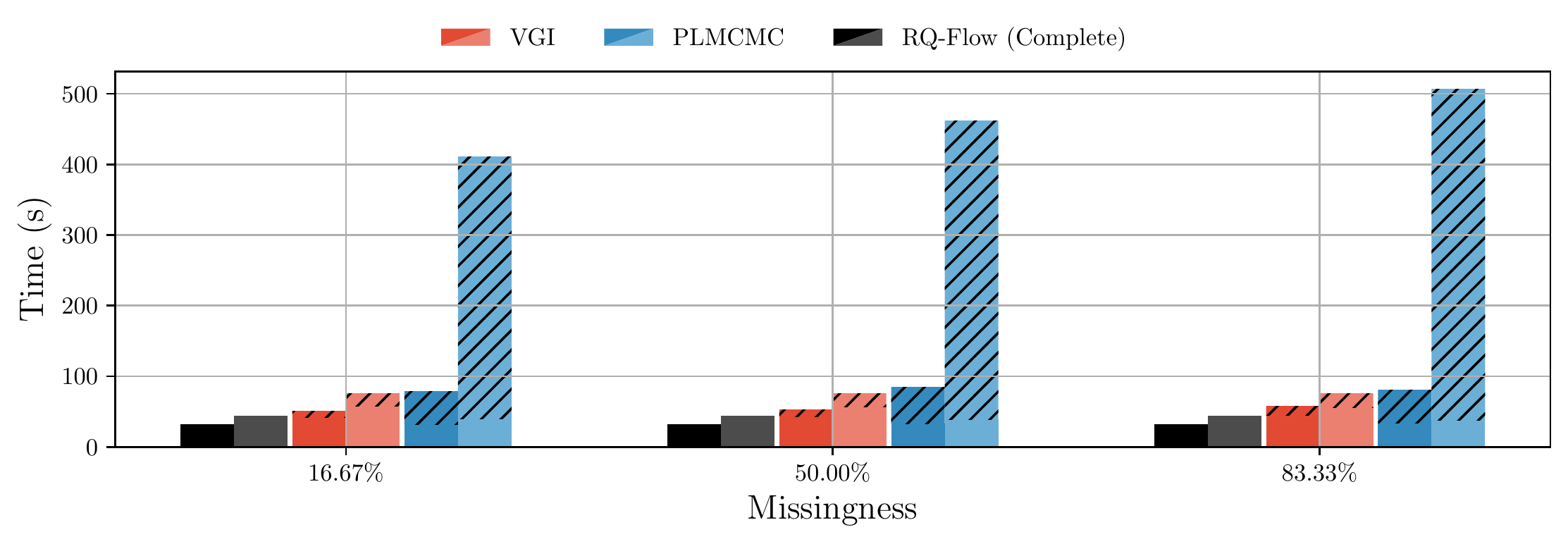} %
    \vspace{-1.em}
    \caption{The average duration of a training iteration on the MINIBOONE data set when using coupling-based (darker) and autoregressive  (lighter) flow models. VGI easily handles autoregressive models without a significant overhead while PLMCMC scales poorly to such models due to a computationally inefficient inverse.}
    \label{fig:rqarspline-miniboone-avg-time}
\end{figure}

The figure further shows that PLMCMC spends a large fraction of its per-iteration cost on performing MCMC (blue hatched bars). Since the PLMCMC sampler requires the inverse transformation to be efficiently computable, the experiments up to this point used a flow model with coupling layers whose inverse can be computed at a similar cost as the forward transformations. However, many normalising flow models cannot be inverted efficiently. One example of such a model class are the autoregressive flows, where the forward transformation can be computed in constant time with the dimensionality of the model, but the inverse requires $d$ sequential transformations \citep[\eg][Section 3.1]{Papamakarios2021}.
In Figure~\ref{fig:rqarspline-miniboone-avg-time}, we compare the training iteration cost in seconds on a coupling-based (darker colour) and autoregressive (lighter colour) flow models on the MINIBOONE data.\footnote{For the autoregressive flow we use the autoregressive rational-quadratic neural spline flow model with the setup of \citet[][Appendix~B.1]{durkanNeuralSplineFlows2019}.}
It can be seen that the cost of PLMCMC on the autoregressive model is significantly larger than for the coupling-based model, which prohibits its application to larger data sets, whilst VGI handles such models much more gracefully.

The results in this section contrast with Section~\ref{sec:vae-computational-aspects}, where the VGI-based methods were computationally more expensive than the other VAE-specific estimation methods: here we have found that the general-purpose VGI method is more efficient than the existing method for flow model estimation, whilst also producing a more accurate fit.

\section{Discussion}
\label{sec:discussion}

In this work we have presented a novel approach to (approximate) maximum-likelihood estimation of statistical models from incomplete data.
Our method is applicable to general statistical models and is based on variational inference and Gibbs sampling, both of which have been extensively researched. \revisiontwo{Hence, the proposed method, termed variational Gibbs inference (VGI),} can adopt the diversity of \revisiontwo{techniques} developed for them.
\revisiontwo{We have further introduced} a version of VGI for problems where missingness is known to occur in blocks (Section~\ref{sec:vbgi}), for example, in latent-variable models with incomplete data (as demonstrated on a VAE in Section~\ref{sec:vae-experiments}).

VGI \revisiontwo{addresses limitations of (amortised) variational inference (VI) that has strongly limited its applicability to model estimation from incomplete data (see Sections~\ref{sec:background-density-estimation-and-vi}\nobreakdash{-}\ref{sec:background-vi}):
(i) VGI} only requires $d$ amortised variational conditionals and thus scales linearly in the dimensionality of the data, gracefully handling the exponential explosion in the number of possible missingness patterns \revisiontwo{and subsequently the number of conditional variational distributions%
, and (ii) it enables the specification of a flexible variational family for the conditional distribution of the missing variables using general probabilistic models without restrictions}. 
The parameters of the statistical model $\pt$ and the variational conditionals $\qpj$ are fitted using stochastic gradient ascent, hence it can be used in large-scale data settings that are common in machine learning.
\revisiontwo{The proposed method thus enables efficient learning by drawing on the strengths of amortised variational inference, thanks to property (i), and allows us to well-optimise the ELBO and obtain a good estimate of the target statistical model, thanks to property (ii).}

\revisiontwo{We have validated our method on a toy factor analysis (FA) model} against the EM algorithm, which corresponds to the best solution that can be achieved by any variational method, and against a multiple-imputation approach using a popular MICE imputation method in order to \revisiontwo{evaluate against the two-stage impute-then-fit approaches to statistical model estimation from incomplete data}.
Our method achieved comparable performance to the classical EM algorithm in a low-dimensional setting and performed similarly to the Monte Carlo EM in the higher-dimensional setting.
This suggests that given sufficient compute resources, VGI can achieve good estimates of the model.
Moreover, VGI outperformed the impute-then-fit approaches in a higher-dimensional setting even \revisiontwo{when} the (MICE) imputer used the correct family of imputation models relative to the target (FA) model.
\revisiontwo{This suggests that optimising a single variational lower-bound might be a more efficient approach to estimating statistical models from incomplete data than the two-stage impute-then-fit approaches.}

\revisiontwo{We have demonstrated that VGI is particularly well-suited for modern models}
by successfully fitting important \revisiontwo{modern} statistical models in machine learning, namely VAEs and normalising flows in Sections~\ref{sec:vae-experiments} and \ref{sec:flow-experiments}, respectively.\footnote{The evaluated models are for continuous random variables, and we focused on them due to their popularity, but the theory presented in Section~\ref{sec:vgi} is general and also holds for discrete distributions.}
\revisiontwo{In our experiments, VGI achieved better performance than the existing state-of-the-art model-specific estimation methods.}

\subsection{Future Work}

\revisiontwo{We see a number of interesting future directions following this work.}

\revisiontwo{Due to the generality of the method,} it is a candidate to be implemented in probabilistic modelling platforms as a general-purpose inference engine for incomplete data, or as a library in machine learning frameworks. 
A readily available framework may help improve the current practice of handling missing data, where simple fixes are often preferred over the principled probabilistic methods like VGI. 

\revisiontwo{We also believe that there is area for improvement in the computational efficiency of the method.}
It is widely known that the mixing of the Markov chain may be inefficient in Gibbs sampling if the variables are highly correlated, which may slow down the rate of improvement in VGI.
However, we can use the rich literature on Gibbs sampling to improve VGI:
To improve the mixing rate of the Gibbs imputation chains one could use ordered over-relaxation \citep{nealSuppressingRandomWalks1995} or apply adaptive-scan Gibbs methods to find Gibbs selection probabilities $\pi(j)$ that are better than the uniform ones that we used in this paper \citep{Latuszynski2013, Chimisov2018, grathwohlOopsTookGradient2021}, thereby increasing the overall performance of VGI. 
Alternatively, one could use the (near\nobreakdash-)conditional-independencies of the variational conditionals, if such exist, to update some of the dimensions simultaneously, thus breaking the sequential Gibbs update requirement but potentially increasing the mixing rate of the imputation chains \citep{Angelino2016, Terenin2020}. However, the effect of these changes on the performance of VGI is yet unknown.

Another area of potential improvement is the computational efficiency of evaluating and sampling the variational model.
In this work we have implemented the variational conditionals moderately efficiently by using the extended-Gibbs kernel with partially-shared parameters and then computing all conditional distributions simultaneously using optimised matrix operations.
However, we usually do not require all conditional distributions for each data-point; only a subset of sampled missing dimensions in the VGI objective in \eqref{eq:vgi-objective-approx} is required.
We believe this could be addressed by leveraging the efficient sparse matrix computation libraries built for graph and compressed neural network processing \citep{bellEfficientSparseMatrixVector2008, hanEIEEfficientInference2016, huillcen-bacaEfficientSparseMatrixVector2019}.

Finally, while our investigation considered statistical model estimation under the ignorable missingness assumption, which is a standard assumption in the missing data field, in practice, the missingness mechanisms are often non-ignorable, and making the ignorability assumption can result in incorrect inferences.
Some recent work on VAEs has shown promising results for non-ignorable missingness \citep{ipsenNotMIWAEDeepGenerative2020, collierVAEsPresenceMissing2021, maIdentifiableGenerativeModels2021}, however no such work has yet been done on normalising flows.
Considering the importance of non-ignorable missingness in practice, 
extending VGI to statistical model estimation from incomplete data that is subject to non-ignorable missingness is an important future direction towards a general-purpose estimation engine for incomplete data.

\acks{Vaidotas Simkus gratefully acknowledges the support from the University of Edinburgh and the financial support from Huawei through their grant for PhD Studentships in Dialogue Systems and Data Systems. Benjamin Rhodes was supported in part by the EPSRC Centre for Doctoral Training in Data Science,
funded by the UK Engineering and Physical Sciences Research Council (grant EP/L016427/1) and
the University of Edinburgh. We would also like to thank the anonymous reviewers for their constructive feedback and valuable suggestions that have helped us improve the paper.}

\appendix

\section{Ignorable Missingness Assumption}
\label{apx:ignorable-missingness}

In this section we summarise the three classes of missingness as proposed by \citet{Rubin1976} and their effects on statistical model estimation.
An important aspect of the incomplete data model in \eqref{eq:incomplete-joint-model} is that the missingness mask $\m$ depends on both observed $\xo$ and missing $\xm$ values of the data. This generally means that the estimation of the statistical model $\pt(\x)$ is coupled with the missingness model $p(\m \mid \x)$.
\citet{Rubin1976} proposed a taxonomy of missing data mechanisms that consists of three levels of simplifying assumptions:
\begin{enumerate}
    \item Missing not at random (MNAR): no independence assumptions,
    \item Missing at random (MAR): $p(\m \mid \xo, \xm) = p(\m \mid \xo)$,
    \item Missing completely at random (MCAR): $p(\m \mid \xo, \xm) = p(\m)$.
\end{enumerate}
MAR and MCAR assumptions have been widely used in many incomplete data methods since they decouple the model $\pt(\x)$ estimation task from the missingness model $p(\m \mid \x)$
\begin{align*}
    \hat \thetab &= \argmax_{\thetab} \prod_{i=1}^N \int \pt (\xo^i, \xm^i) p(\m^i \mid \xo^i) \dif \xm^i \\
    &= \argmax_{\thetab} \prod_{i=1}^N p(\m^i \mid \xo^i) \prod_{i=1}^N \int \pt (\xo^i, \xm^i) \dif \xm^i \\
    &= \argmax_{\thetab} \prod_{i=1}^N \int \pt (\xo^i, \xm^i) \dif \xm^i,
\end{align*}
where the product is due to computing the likelihood for N i.i.d.\@ observed data-points.
These assumptions are also known as \emph{ignorable missingness mechanism} because the missingness model can be simply ignored when the only goal is to estimate the model $\pt(\x)$.\footnote{Assuming the model $\pt(\x)$ is flexible enough, otherwise the estimate can be biased even with MAR misingness \citep[][Section~3.4]{marlinMissingDataProblems2008}.}

\section{Gibbs Sampling}
\label{apx:gibbs-sampling}

\revisiontwo{
Gibbs sampling \citep{Geman1984} is a Markov chain Monte Carlo \citep[MCMC, \eg][Chapter~27.3]{Barber2017} method for (approximately) sampling a joint model $\pt(\x)$ where directly sampling the joint distribution is intractable.
The method factorises the distribution using the chain rule $\pt(\x) = \pt(\xj \mid \xnj) \pt(\xnj)$, and assumes that the conditional distribution $\pt(\xj \mid \xnj)$ is easy to sample from for any index $j$. 
Then, the sampler starts with any random initial sample $\x^{0}$, selects one variable index $j$ to sample, and updates the sample $\x^{t+1} = \{\xj^{t+1}, \xnj^{t}\}$ with $\xj^{t+1} \sim \pt(\xj \mid \xnj^{t})$. 
By continuing this procedure, the Gibbs sampler asymptotically in the number of iterations $t$ samples the joint distribution of the model $\pt(\x)$. 
}

\revisiontwo{
The order in which each dimension is updated by the Gibbs sampler is called the ``scan''. The two most common scans are: random, where the dimension index $j$ at every iteration $t$ is chosen uniformly at random, and systematic, in which a fixed order of the variables is selected and the sampler repeatedly iterates over the variables in that order.
Our VGI method uses the randomised scan as presented in Section~\ref{sec:vgi}.
However, we feel that it is important to note that neither of these scans are necessarily optimal, and as highlighted in the discussion there are ways to adapt the scan of the Gibbs sampler \citep{Latuszynski2013, Chimisov2018,grathwohlOopsTookGradient2021}.}

\revisiontwo{
Gibbs sampling methods have been crucial for sampling and learning some important classes of models in physics, such as the Ising model \citep[\eg][Section 4.2.5]{Barber2017}.
However, the required conditionals $\pt(\xj \mid \xnj)$ are typically intractable for most modern statistical models, such as VAEs and normalising flows, and hence, efficient sampling is not tractable either. 
In VGI, we approximate the Gibbs conditionals with an amortised variational model, which allows us to overcome this issue.}

\section{Multiple Imputation by Chained Equations (MICE)}
\label{apx:mice}

The MICE algorithm starts with random imputations of missing data and then alternates between fitting the univariate conditional distributions $f(\xj \mid \xnj; \phibj)$ using the data-points where $\xj$ is observed and imputing the data-points where $\xj$ is missing for $\forall j \in 1 \ldots d$. For example, the $t$-th iteration of the method proceeds as follows 
\begin{align*}
    &\hat \phib_1^{(t)} = \argmax_{\phib_1} \sum_{i \in \obs(\D, 1)} \log f(x_1^{(i, t-1)} \mid \x_{\smallsetminus 1}^{(i, t-1)}; \hat \phib_1^{(t-1)}) \\
    &x_1^{(i, t)} \sim f(x_1^{(i, t)} \mid \x_{\smallsetminus 1}^{(i, t-1)}; \hat \phib_1^{(t)}) \text{ for } \forall i \in \mis(\D, 1) \\
    &\phantom{----------}\vdots \\
    &\hat \phib_d^{(t)} = \argmax_{\phib_d} \sum_{i \in \obs(\D, d)} \log f(x_d^{(i, t-1)} \mid \x_{\smallsetminus d}^{(i, t)}; \hat \phib_d^{(t-1)}) \\
    &x_d^{(i, t)} \sim f(x_d^{(i, t)} \mid \x_{\smallsetminus d}^{(i, t)}; \hat \phib_d^{(t)}) \text{ for } \forall i \in \mis(\D, d),
\end{align*}
where $\obs(\D, j)$ and $\mis(\D, j)$ are respectively the sets of indices of data-points in the data set that are observed and missing in the $j$-th feature dimension. The procedure is usually repeated several times over all the missing variables to converge to reasonable imputations.
Alternatively, if the conditional models $f(\xj \mid \xnj, \phi_j)$ admit it, the approach can be made Bayesian by specifying a prior over the parameters $\phib_j$ and then sampling the parameters $\phib_j^{(t)}$ from the posterior distribution $f(\phib_j \mid \{\x_{< j}^{(t)}, \x_{\geq j}^{(t-1)}\}_{\obs(\D, j)})$ instead of the $\argmax$ in the above procedure. 

Moreover, as dictated by the multiple-imputation framework, the overall imputation procedure is usually repeated $K$ times, each time starting with random baseline imputations, to obtain $K$ independent imputations of the missing data, which can then be used for many downstream tasks.

\section{Combining Multiple Imputations for Statistical Model Estimation}
\label{apx:combining-mi}

\begin{figure}[tb]
    \centering
    \begin{minipage}[c]{.45\linewidth}
    \begin{tikzpicture}[x=1.2cm,y=0.3cm]
      \node[latent]                (D)      {$\D$} ;

      \node[latent, right=of D, yshift=1.5cm]   (D1)      {$\D_1$} ;
      \node[latent, below=of D1]   (D2)      {$\D_2$} ;
      \node[latent, below=of D2]   (D3)      {$\D_3$} ;
      \node[latent, below=of D3]   (D4)      {$\D_4$} ;
      
      \edge {D} {D1} {} ;
      \edge {D} {D2} {} ;
      \edge {D} {D3} {} ;
      \edge {D} {D4} {} ;

      \node[latent, right=of D1]   (R1)      {$\hat \thetab_1$} ;
      \node[latent, right=of D2]   (R2)      {$\hat \thetab_2$} ;
      \node[latent, right=of D3]   (R3)      {$\hat \thetab_3$} ;
      \node[latent, right=of D4]   (R4)      {$\hat \thetab_4$} ;

      \node[latent, right=of R1, yshift=-1.5cm]  (R)  {$\hat \thetab$} ;
      
      \node[inner sep=0pt] (R_fig) at ([xshift=({0.12cm})]R)
      {\includegraphics[width=.18\textwidth]{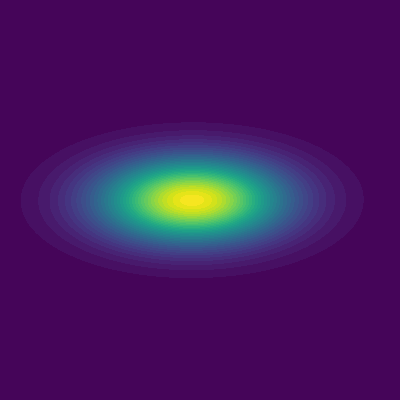}};
        
      \node[inner sep=0pt] (R1_fig) at (R1)
      {\includegraphics[width=.14\textwidth]{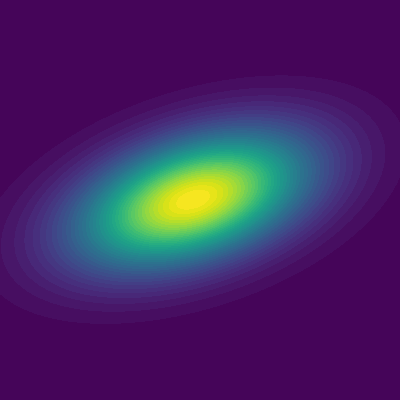}};
        
      \node[inner sep=0pt] (R2_fig) at (R2)
      {\includegraphics[width=.14\textwidth]{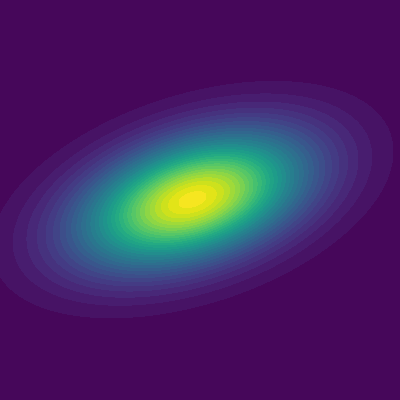}};
        
      \node[inner sep=0pt] (R3_fig) at (R3)
      {\includegraphics[width=.14\textwidth]{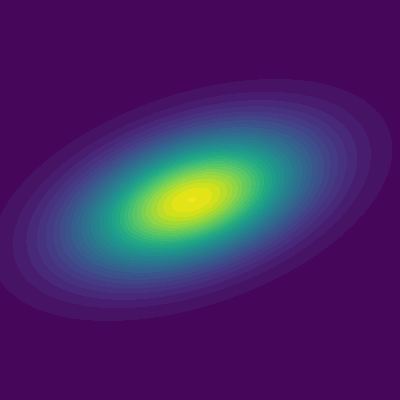}};
        
      \node[inner sep=0pt] (R4_fig) at (R4)
      {\includegraphics[width=.14\textwidth]{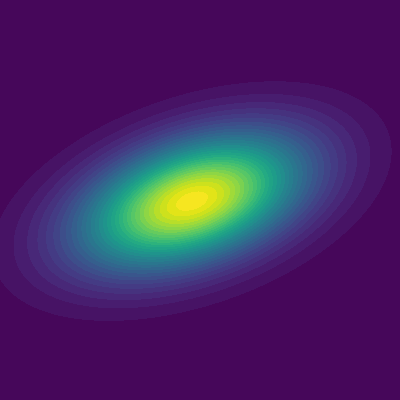}};
      
      \small{
      \node[above=of R1, yshift=-0.8cm] (Fk) {$\hat F_k \hat F_k^{\top} + \hat \Psi_k$} ;
      
      \node[above=of R, yshift=-0.5cm] (F) {$\hat F \hat F^{\top} + \hat \Psi$} ;
      }
      
      \edge {D1} {R1_fig} {} ;
      \edge {D2} {R2_fig} {} ;
      \edge {D3} {R3_fig} {} ;
      \edge {D4} {R4_fig} {} ;
      
      \edge {R1_fig} {R_fig} {} ;
      \edge {R2_fig} {R_fig} {} ;
      \edge {R3_fig} {R_fig} {} ;
      \edge {R4_fig} {R_fig} {} ;
      
      \small{
      \node[below=of D, align=center, yshift=-0.8cm] (D_text) {Incomplete\\data set} ;
      \node[below=of D4, align=center, yshift=0.75cm] (D4_text) {Completed\\data sets} ;
      \node[below=of R4, align=center, yshift=0.75cm] (R4_text) {Fitted\\estimates} ;
      \node[below=of R, align=center, yshift=-0.8cm] (R_text) {Na\"ive\\averaging} ;
      }
    \end{tikzpicture}
    \end{minipage}
    \qquad
    \begin{minipage}[c]{.45\linewidth}
    \includegraphics[width=1\linewidth]{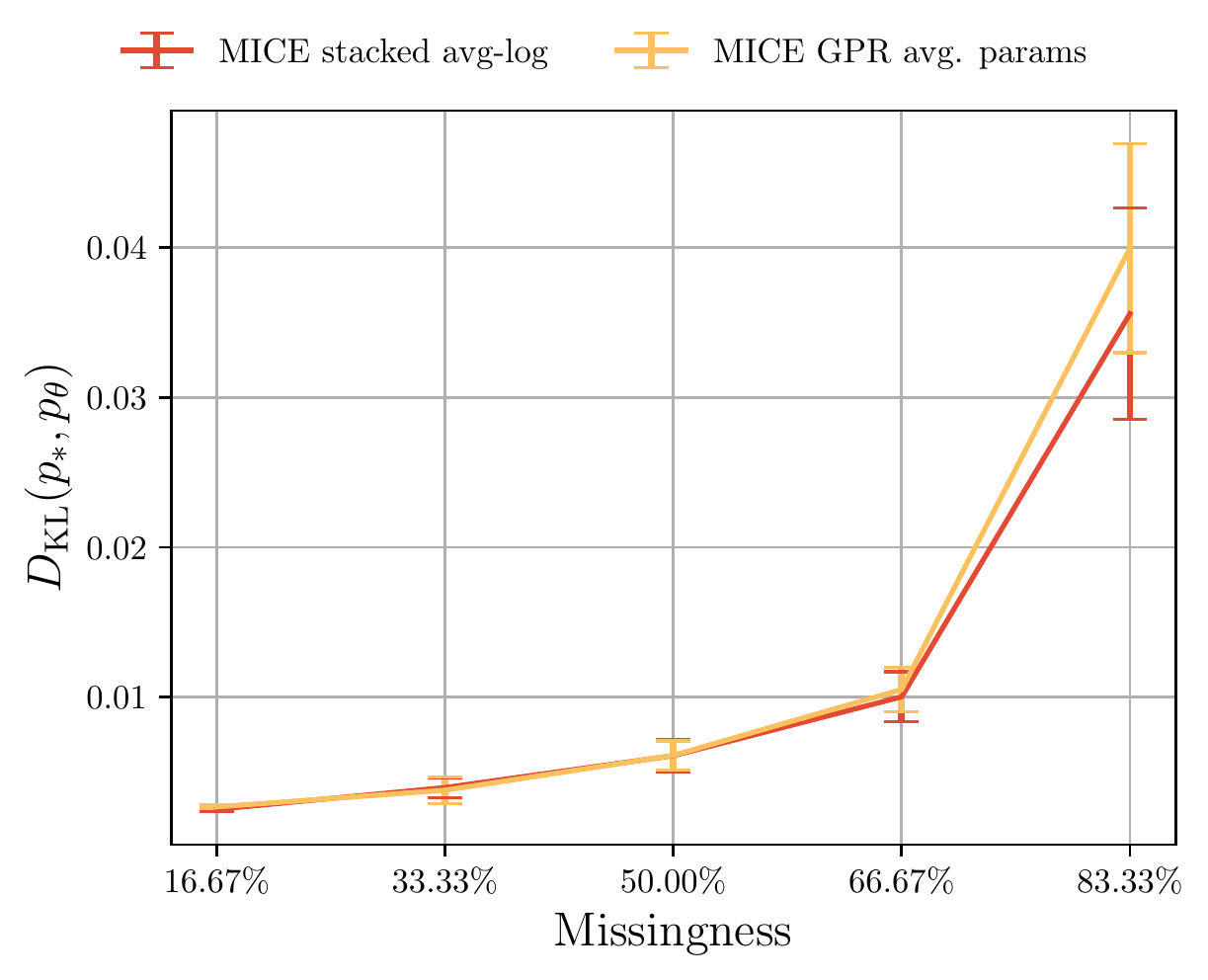}
    \end{minipage}
    \caption{The effect of parameter averaging on a FA model when fitted on imputations from MICE. Left: na\"ive averaging of parameter estimates from different imputations removes the correlations in the averaged model.
    Right: comparison of model accuracy after parameter averaging of models trained on independent completed data set (yellow), where generalised Procrustes rotation was applied to the factor loading matrices prior the averaging, and a model trained on several completed data sets stacked together using average log-likelihood (red). 
    Accuracy is measured using KL divergence of the factor analysis model to the ground truth model. Fitting on a stacked completed data set (red) yields equally good parameter estimates as the GPR-based averaging approach.}
    \label{fig-apx:toy-kldiv-mice-stacked-vs-avg-params}
\end{figure}

Multiple imputation \citep[MI,][]{Rubin1987a} has been the primary tool for the analysis of incomplete data for more than two decades. Many books have covered the standard workflow of principled statistical analysis with multiple imputation and the risks of following invalid procedures \citep{Schafer1997, gelmanBayesianDataAnalysis2013, VanBuuren2018}.
In this section we first describe the standard analysis workflow with MI and then discuss why it is problematic or cannot be used for statistical models that have non-identifiable parameters. We demonstrate the issue on a classical factor analysis (FA) model, where, unlike for modern deep models (for example, VAEs), there are known methods to resolve the issue. We then explain the procedure we followed in our baseline experiments that use MI-based methods (for example, MICE) and verify on the FA model that it obtains equally good estimates of the statistical model as the standard analysis procedure.

In standard analysis with MI, initially K sets of imputed data are generated using a preferred multiple-imputation method such as the FCS described in Section~\ref{sec:background-fcs}. Then, for each imputed data set an independent analysis is performed treating the imputed data as complete and producing $K$ estimates of a desired analysis quantity $\thetab^1, \ldots, \thetab^K$, namely one from each set of imputed data.
Following Rubin's rules \citep[][Result~3.2]{Rubin1987a} the estimated quantities are then generally combined by averaging
\begin{align*}
    \hat \thetab = \frac{1}{K} \sum_{k=1}^K \thetab^k.
\end{align*}
The estimated quantities $\thetab^1, \ldots, \thetab^K$ can further be used to perform statistical tests and provide confidence intervals of the estimate $\hat \thetab$. 

However, such averaging requires that $\thetab$ is identifiable. If identifiability is not possible this procedure may result in meaningless estimates.
Parametric non-identifiability is a common property of many statistical models starting from classical models, such as mixtures of Gaussians and factor analysis, to modern deep models, such as variational autoencoders and normalising flows, where the non-identifiability stems from the use of neural networks.

We illustrate the issues arising in MI analysis from non-identifiability on a simple example, the factor analysis model, where the factor loading matrix is generally only identifiable up to a multiplication with an orthonormal matrix. 
Hence, estimating the factor matrix using MI may produce estimates $W^1, \ldots, W^K$ that could be written as the product of the true factor matrix $F$ and an orthonormal matrix $R^k$, $W^k = F R^k$. Na\"ively averaging such estimates would give
\begin{align*}
    \hat W = \frac{1}{K} \sum_{k=1}^K W^k = \frac{1}{K} \sum_{k=1}^K F R^k = F \frac{1}{K} \sum_{k=1}^K R^k,
\end{align*}
where the average over orthonormal matrices is not a valid orthonormal matrix, and hence the average result is not a valid estimate of the parameters.  
On the left side of Figure~\ref{fig-apx:toy-kldiv-mice-stacked-vs-avg-params} we
demonstrate that such na\"ive averaging approach produces a meaningless model estimate.

For classical models, such as mixtures of Gaussians and factor analysis it is often possible to resolve the indeterminacies of the $K$ estimates before averaging.
One solution for the FA model is to transform the factor loading matrices before averaging using the generalised (orthogonal) Procrustes rotation \citep[GPR,][]{tenbergeOrthogonalProcrustesRotation1977,vanginkelUsingGeneralizedProcrustes2014,lorenzo-sevaMultipleImputationMissing2016}.
However, no obvious solution exists to resolve the indeterminacy of deep models, which are the main interest of this paper, and hence the standard MI analysis methodology via parameter averaging described above is generally not applicable.

Therefore to avoid averaging, an alternative approach was used in our experiments where MI methods (such as MICE) were used to estimate baseline statistical models. 
Specifically, we have stacked the imputed data into a single K-times imputed data set and then estimate the model $\pt$ using weighted maximum-likelihood by weighting each imputation by $1/K$
\begin{align*}
    \hat \thetab &= 
    \argmax_{\thetab} \frac{1}{N} \sum_{i=1}^N \frac{1}{K} \sum_{k=1}^K \log \pt (\txm^\sik, \xo^i), %
\end{align*}
where $\txm^\sik$ denotes the $k$-th imputation of the $i$-th sample.

On the right side of Figure~\ref{fig-apx:toy-kldiv-mice-stacked-vs-avg-params} we validate that the stacking approach (red) described above produces equally good parameter estimates as the GPR-based parameter averaging (yellow) for all fractions of missingness. 
While standard statistical tests may provide invalid test statistics or confidence intervals if (incorrectly) applied to the stacked data \citep[\eg][Chapter~5.1]{VanBuuren2018} here we are primarily interested in obtaining point estimates of the parameters and hence use the stacked approach since no method (such as the GPR for FA) exists to resolve the parametric indeterminacy of neural networks.

\section{VGI Derivation with Extended-Gibbs Kernel}
\label{apx:vgi-derivation-extended}

We here generalise the VGI lower-bound \eqref{eq:variational-Gibbs-elbo} to the case of the extended-Gibbs kernel where the variational conditionals can depend on all variables imputed in the previous iteration. We use this extension to achieve a computationally more efficient method by enabling computation sharing among all variational conditionals (see  Section~\ref{sec:variational-model}).

We define the extended-Gibbs kernel by letting the variational conditionals depend on the previous imputation value $\txj$ in dimension $j$
\begin{align}
    \tilde \tau_\phib(\xm \mid \xo, \txm) &= \sum_{j \in \idx{\m}} \pib(j) \qpj(\xj \mid \xmnj, \txj, \xo) \delta (\xmnj - \txmnj). \label{eq:extended-gibbs-transition}
\end{align}
Marginalising out $\txm$ in the above equation with respect to the previous imputation distribution $f^{t-1}$ yields the imputation distribution after a single Gibbs update
\begin{align}
    f_\phib^t(\xm \mid \xo) &= \int \tilde \tau_\phib(\xm \mid \xo, \txm) f^{t-1}(\txm \mid \xo) \dif \txm \nonumber\\
    &= \sum_{j \in \idx{\m}} \pib(j) \int f^{t-1}(\txj \mid \xo) \nonumber\\
    &\phantom{= \sum_{j \in \idx{\m}} \pib(j) \int } f^{t-1} (\xmnj \mid \xo, \txj) \qpj(\xj \mid \xmnj, \txj, \xo) \dif \txj \nonumber\\
    &= \sum_{j \in \idx{\m}} \pib(j) \int f^{t-1}(\txj \mid \xo) f_\phib^t(\xm \mid \xo, \txj, j) \dif \txj %
    \nonumber\\
    & = \E_{ \pib(j) f^{t-1}(\txj \mid \xo)} \left[ f_\phib^t(\xm \mid \xo,\txj, j) \right] \label{eq:joint-var-distribution-extended}\\  
    f_\phib^t(\xm \mid \xo, \txj, j) &= \qpj(\xj \mid \xmnj, \txj, \xo) f^{t-1}(\xmnj \mid \xo, \txj). \label{eq:joint-var-distribution-extended-j}
\end{align}
Now we continue from the standard ELBO in \eqref{eq:elbo-marginal-markov-chain} and use \eqref{eq:joint-var-distribution-extended} and \eqref{eq:joint-var-distribution-extended-j}
\begin{align}
    \Lc^t(\thetab, \phib; \xo)
    &\oset[1.2ex]{\eqref{eq:joint-var-distribution-extended}}{=} \E_{ \pib(j) f^{t-1}(\txj \mid \xo) f_\phib^t(\xm \mid \xo, \txj, j)} \left[ \log \frac{\pt(\xo, \xm)}{f_\phib^t(\xm \mid \xo)} \right] \nonumber\\
    &=\E_{ \pib(j) f^{t-1}(\txj \mid \xo) f_\phib^t(\xm \mid \xo, \txj, j)} \Bigg[ \log \frac{\pt(\xo, \xm)}{f_\phib^t(\xm \mid \xo, \txj, j)} \nonumber\\
    &\phantom{=\E_{ \pib(j) f^{t-1}(\txj \mid \xo) f_\phib^t(\xm \mid \xo, \txj, j)} } +\log \frac{f_\phib^t(\xm \mid \xo, \txj, j)}{f_\phib^t(\xm \mid \xo)} \Bigg] \nonumber\\
    &= \E_{ \pib(j) f^{t-1}(\txj \mid \xo) f_\phib^t(\xm \mid \xo, \txj, j)} \left[ \log \frac{\pt(\xo, \xm)}{f_\phib^t(\xm \mid \xo, \txj, j)} \right] \nonumber\\
    &\phantom{==} + \E_{\pib(j) f^{t-1}(\txj \mid \xo)} \KLD{f_\phib^t(\xm \mid \xo, \txj, j) }{f_\phib^t(\xm \mid \xo)} \nonumber\\
    &\geq \E_{ \pib(j) f^{t-1}(\txj \mid \xo) f_\phib^t(\xm \mid \xo, \txj, j)} \left[ \log \frac{\pt(\xo, \xm)}{f_\phib^t(\xm \mid \xo, \txj, j)} \right] \nonumber\\
    &\oset[1.2ex]{\eqref{eq:joint-var-distribution-extended-j}}{=} \E_{ \pib(j) f^{t-1}(\txj \mid \xo) f^{t-1}(\xmnj \mid \xo, \txj) \qpj(\xj \mid \xmnj, \txj, \xo)} \Bigg[ \nonumber\\
    &\phantom{====} \log \frac{\pt(\xo, \xm)}{\qpj(\xj \mid \xmnj, \txj, \xo)} \Bigg] \nonumber\\
    &\phantom{==} - \E_{{\pib(j)} f^{t-1}(\txj \mid \xo) f^{t-1}(\xmnj \mid \xo, \txj)} \left[ \log f^{t-1}(\xmnj \mid \xo, \txj) \right] \nonumber\\
    &= \E_{ \pib(j) f^{t-1}(\textcolor{red}{\txj}, \xmnj \mid \xo) \qpj(\xj \mid \xmnj, \textcolor{blue}{\txj}, \xo)} \left[ \log \frac{\pt(\xo, \xm)}{\qpj(\xj \mid \xmnj, \textcolor{blue}{\txj}, \xo)} \right] \nonumber\\
    &\phantom{==}- \E_{{\pib(j)} f^{t-1}(\textcolor{magenta}{\txj}, \xmnj \mid \xo)} \left[ \log f^{t-1}(\xmnj \mid \xo, \textcolor{magenta}{\txj}) \right] \label{eq:variational-Gibbs-elbo-extended},
\end{align}
where the inequality follows from the non-negativity of KL divergence.
Dropping the entropy term in the second line of \eqref{eq:variational-Gibbs-elbo-extended} yields the VGI objective with extended conditionals.
We note that the VGI ELBO in \eqref{eq:variational-Gibbs-elbo-extended} obtained using the extended-Gibbs kernel is analogous to \eqref{eq:variational-Gibbs-elbo}, only the coloured terms are different. To sum up:
\begin{itemize}
    \item It uses the imputation value $\txj$ from the previous step in the Markov chain, highlighted in red.
    \item The variational conditionals $\qpj$ depend on the previous step in the Markov chain, highlighted in blue.
    \item The entropy term of the imputation distribution $f^{t-1}$ from \eqref{eq:variational-Gibbs-elbo} became a conditional entropy, highlighted in magenta. Since the term does not include the parameters of interest $\thetab$ and $\phib$, it does not need to be computed during optimisation.
\end{itemize}

The extended-Gibbs conditionals improve the efficiency of the main stage of the VGI algorithm. We note that the extended-Gibbs conditionals need to be separately defined for the  variational warm-up stage (line~\ref{alg:line:var-warmup} of Algorithm~\ref{alg:VGI} and Algorithm~\ref{alg:var-warm-up}), since $\xj$ has no ``previous imputation'' to condition on. Including the true $\xj$ in the conditioning set for the $j$-th dimension would cause the variational model to approximate the identity function, which may not be a desirable initialisation. To enable variational model warm-up with the extended conditionals we mask the $j$-th input to the conditional $\qpj$ with a zero.

Furthermore, in the VBGI objective for latent-variable models in \eqref{eq:vbgi-latent-objective}, we can similarly let the univariate conditionals depend on the previous imputation, which results in the following VBGI objective with extended conditionals
\begin{align*}
    \J^t_\text{VBGI}(\thetab, \phib; \xo) &= \E_{f^{t-1}(\txm, \txz \mid \xo)} \Bigg( \nonumber \\
    &\phantom{=}\sum_{j \in \idx{\m}} \pib(j) \E_{\qpj(\xj \mid \textcolor{blue}{\txm}, \txz, \xo) } \left[ \log \frac{\pt(\xj, \txmnj, \xo \mid \txz) \pt(\txz)}{\qpj(\xj \mid \textcolor{blue}{\txm}, \txz, \xo)} \right] \nonumber\\
    &\phantom{=}+ \pib(j=z) \E_{q_{\phib_z}(\xz \mid \txm, \textcolor{blue}{\txz} \xo) } \left[ \log \frac{\pt(\txm, \xo \mid \xz) \pt(\xz)}{q_{\phib_z}(\xz \mid \txm, \textcolor{blue}{\txz}, \xo)} \right] \Bigg).
\end{align*}
It allows us to achieve a similar computational improvement as in the standard VGI case with the extended conditionals.

\section{Properties of the Gibbs Kernel in the Stationary Regime}
\label{apx:gibbs-kernel-stationary-distribution}

We show that in the stationary regime the distribution obtained after a Gibbs transition is independent of the updated dimension $j$. We use a modified kernel from \eqref{eq:gibbs-transition}, which updates a fixed dimension $j$
\begin{align*}
    \tau_\phib(\xm \mid \xo, \txm, j) = \qpj(\xj \mid \xmnj, \xo) \delta (\xmnj - \txmnj).
\end{align*}
We now marginalise the kernel with respect to the stationary distribution $q^*(\xm \mid \xo)$ with conditionals $\qpj(\xj \mid \xmnj, \xo)$
\begin{align*}
    q^*(\xm \mid \xo, j) &= \int \tau_\phib(\xm \mid \xo, \txm, j) q^*(\txm \mid \xo) \dif \txm\\
    &= \qpj(\xj \mid \xmnj, \xo)  q^*(\xmnj \mid \xo)\\
    &= q^*(\xm \mid \xo).
\end{align*}
Hence, the stationary distribution is invariant to the updated dimension $j$.

\section{The Variational Inference Network}
\label{apx:variational-network}

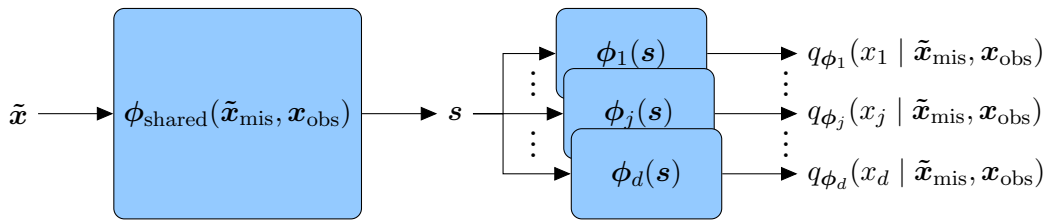
\begin{figure}[t]
  \centering
  \definecolor{shared}{HTML}{94CAFF}
  \definecolor{net}{HTML}{94CAFF}
  \tikzset{shared/.style={black,draw=black,fill=shared,rectangle,rounded corners,minimum height=2.8cm}}
  \tikzset{net/.style={black,draw=black,fill=net,rectangle,rounded corners,minimum height=1.2cm,minimum width=2cm}}
  
  \begin{tikzpicture}[x=1.2cm,y=0.3cm]
    \node (x) {$\tx$};
    \node[style=shared,right=of x] (shared) {$\phib_{\text{shared}}(\txm, \xo)$};
    \node[right=of shared] (s) {$\bm{s}$};
    
    \edge {x} {shared} {} ;
    \edge {shared} {s} {} ;
    
    \foreach \i in {1,2,3}
       {
       \node[style=net,right=of s,xshift=0.1cm*\i, yshift=1.6cm-0.8cm*\i]  (indep\i) {$\phib_{\ifthenelse{\i=2}{j}{\ifthenelse{\i=3}{d}{\i}}}(\bm{s})$};

        \draw [->] (s.east) -- +(0.4cm,0) |- node[right=of s] {} (indep\i.west);
        
        \node[right=of indep\i,xshift=0.1cm*3-0.1cm*\i] (out\i) {$q_{\phib_{\ifthenelse{\i=2}{j}{\ifthenelse{\i=3}{d}{\i}}}}(x_{\ifthenelse{\i=2}{j}{\ifthenelse{\i=3}{d}{\i}}} \mid \txm, \xo)$};
        
        \edge {indep\i} {out\i} {} ;
       }

    \node[right=of s, xshift=-0.4cm, yshift=0.5cm] (left_vdots1) {\Large$\vdots$};
    \node[right=of s, xshift=-0.4cm, yshift=-0.3cm] (left_vdots2) {\Large$\vdots$};
    \node[left=of out2, xshift=1.05cm, yshift=0.5cm] (right_vdots1) {\Large$\vdots$};
    \node[left=of out2, xshift=1.05cm, yshift=-0.3cm] (right_vdots2) {\Large$\vdots$};
  \end{tikzpicture}
  \caption{Partially-shared model of the extended variational conditionals.}
  \label{fig:relax-shared-var-model}
\end{figure}

Using VGI to fit a statistical model $\pt$ first requires us to specify and model $d$ variational distributions. 
One way to model them is to use $d$ inference networks with parameters $\phibj$, one for each variational distribution.
Thus, the neural network for the $j$-th variational distribution takes an (imputed) data-point $\txnj$ without the target dimension $j$ and outputs the parameters of the distribution $\qpj(\xj \mid \txnj)$.
Note that all inference networks that parametrise the $d$ Gibbs conditionals have the same input-output dimensionality, hence the computation of all variational distribution parameters can be efficiently parallelised using optimised matrix operations.

Considering that the variational distributions are approximating the conditionals of a joint target distribution we motivate that some parameter sharing in the variational model can be beneficial.
One way to incorporate partial parameter sharing is to use a two-part inference network, where the first part of the network is shared among all conditionals and the second part is independent for each conditional.
The shared network can then be used for the different conditionals by fixing the $j$-th input dimension to zero when computing the intermediate representation for the $j$-th conditional.
In our evaluations we compare an independent model with a partially-shared model and find that they both perform similarly well (see Section~\ref{sec:fa-accuracy-variational-model}). Importantly, a partially-shared model allows us to scale VGI to higher-dimensional data, by increasing the parameter-efficiency of the variational model.

The VGI objective in \eqref{eq:vgi-objective-approx} requires computing $M$ conditional distributions for each data-point, and the dimensions $j^m$ of those conditionals can differ for each example.
One way to compute the parameters of the conditional distributions is to evaluate the inference network for each dimension separately, however, this can require up to $d$ evaluations and hence can be slow in higher dimensions.
Alternatively, we could compute all $d$ conditionals in parallel and then select only the required conditionals, however, this would not always be possible, since it would require $d$-times more memory.
To mitigate this, for high-dimensional data we suggest working with the extended-Gibbs variational conditionals where the $j$-th conditional distribution is allowed to depend on the current imputation in the $j$-th dimension, thus giving variational distributions of the form $\qpj(\xj^t \mid \xo, \txj^{t-1}, \txmnj^{t-1})$. 
Figure~\ref{fig:relax-shared-var-model} depicts such an extended model.
We now need a \emph{single} pass through the shared layers, rather than $d$ passes, which allows for a significant amount of the computation to be shared and hence gives a significant gain in computational efficiency.
We empirically investigate the effect of the extended variational conditionals in Section~\ref{sec:fa-accuracy-variational-model}, where we find no significant detrimental effects of the extended model when estimating higher-dimensional statistical models $\pt$.

\section{VGI Fine-Tuning Algorithm}
\label{apx:vgi-finetuning}

In this section we describe the VGI fine-tuning algorithm (Algorithm~\ref{alg-apx:VGI-finetuning}) that is used to evaluate the VGI objective in \eqref{eq:vgi-objective} on held-out data. It is similar to the VGI training algorithm (Algorithm~\ref{alg:VGI}) when only the parameters $\phib$ of the variational model are updated.
In our experiments, we use this procedure to select model parameters $\hat \thetab$ that perform best on the validation data, which is not otherwise possible with amortised variational inference without fine-tuning due to the inference generalisation gap discussed in Section~\ref{sec:vgi-eval-held-out}. 

\begin{algorithm}[tb]
    \caption{VGI fine-tuning algorithm}
    \label{alg-apx:VGI-finetuning}
    
    \textbf{Input:} $p_{\hat \thetab}(\x)$, a fitted probabilistic model with parameters $\hat \thetab$ (kept fixed)\\
    \phantom{\textbf{Input:}} $q_{\hat \phib_j}(\xj \mid \xnj)$ for $j \in \{1 \ldots d\}$, var. conditionals fitted on train data with params $\hat \phib$\\
    \phantom{\textbf{Input:}} $\D^{\text{val}}$, incomplete validation data set\\
    \phantom{\textbf{Input:}} $K$, number of imputations of each incomplete data-point\\
    \phantom{\textbf{Input:}} $f^0(\xm \mid \xo)$, initial imputation distribution\\
    \phantom{\textbf{Input:}} $G_\text{W}$, number of Gibbs steps to warm up imputations\\
    \phantom{\textbf{Input:}} $G$, number of Gibbs imputation update steps in each epoch\\
    \phantom{\textbf{Input:}} \emph{max\_epochs}, number of epochs\\
    \textbf{Output:} Validation loss $\hJVGI$ and $K$-times imputed validation data $\D_K^\text{val}$
    \begin{algorithmic}[1]
        \State \textbf{Find} the maximum $\x_\text{max}$ and minimum $\x_\text{min}$ values for each dimension in $\D^\text{val}$
        \State \textbf{Create} $K$-times imputed data $\D_K^{\text{val}}$ using $f^0$
        \State $\phib^0 \leftarrow \hat \phib$ \Comment{Set the initial parameters to the parameters of the fitted variational model}
        \For{$t$ in $[1, \text{\emph{max\_epochs}}]$}
            \For{mini-batch $\B_K$ in $\D_K^{\text{val}}$}
                \If{$t = 1$}
                    \State \parbox[t]{0.9\textwidth}{%
                    \textbf{Update} $\B_K$ with $G_\text{W}$ steps of (pseudo-)Gibbs sampler using the variational conditionals, rejecting any Gibbs update values $\xj$ outside $[\x_\text{min}(j), \x_\text{max}(j)]$\strut}
                \Else
                    \State \textbf{Update} $\B_K$ with $G$ steps of (pseudo-)Gibbs sampler using Algorithm~\ref{alg:pseudo-Gibbs}
                \EndIf
                \State \textbf{Store} the updated imputations in $\B_K$ for use in the next epoch
                \State \textbf{Compute} $\hJVGI^t$ in \eqref{eq:vgi-objective-approx} using Algorithm~\ref{alg:vgi-objective}
                \State $\phib^t = \phib^{t-1} + \alpha_\phib \nabla_\phib \hJVGI^t$ \Comment{Update params of $\qp$ with a stochastic gradient step}
          \EndFor
        \EndFor 
    \end{algorithmic}
\end{algorithm}

It starts with an incomplete held-out data set and finds the minimum and maximum values in the observed data that defines an imputation acceptance region for the warm-up stage. 
Then, the incomplete data are imputed using the initial imputation distribution $f_0$.
In the first iteration of the algorithm, it first ``warms up'' the incomplete data imputations by performing $G_\text{W}$ Gibbs updates, rejecting any imputations outside of the acceptance hypercube. Rejecting imputations that are far from observed data distribution mitigates some of the adverse effects of the inference generalisation gap, but other potentially better rejection strategies, such as the use of a Metropolis-Hastings acceptance step, may also be used.
Then, the algorithm continues in the same way as Algorithm~\ref{alg:VGI}, fine-tuning the variational parameters $\phib$ and updating imputations with $G$ Gibbs updates in each iteration, accepting any imputations that are produced by the Gibbs sampler.

\section{More on Markov Chain Variational Inference}
\label{apx:mcvi}

We review the lower-bound from Markov chain variational inference \citep[MCVI, ][]{Salimans2015}, and show that replacing the reverse transition model $r$ with the true reverse operator restores the tighter lower-bound using the marginal of a Markov chain in \eqref{eq:elbo-marginal-markov-chain}.

The MCVI lower-bound is defined in equation~4 in \citet{Salimans2015} as follows
\begin{align*}
    \Lmcvi(\x) &=\E_{q(\z_0, \ldots, \z^T \mid \x)} \left[ \log \frac{p(\x, \z^T)}{q_0(\z^0 \mid \x)} + \sum_{t = 1}^T \log \frac{r(\z^{t-1} \mid \x, \z^t) }{q (\z^t \mid \x, \z^{t-1})}  \right]
\end{align*}
where $T$ is the total number of transitions in a Markov chain, $q$ and $r$ are transition and reverse transition functions, and $\z$ denotes the latent variable. The joint variational distribution $q$ and the reverse distribution $r$ are assumed to follow a Markov structure $q(\z^0, \ldots, \z^T \mid \x) = q_0(\z^0 \mid \x) \prod_{t=1}^T q(\z^t \mid \x, \z^{t-1})$ and $r(\z^0, \ldots, \z^{T-1} \mid \x, \z^T) = \prod_{t=1}^T r(\z^{t-1} \mid \x, \z^t)$.

We now show that $\Lmcvi$ is a looser lower-bound than the variational ELBO using the marginal of a Markov chain in \eqref{eq:elbo-marginal-markov-chain}, denoting the marginal distribution $q_{T}(\z^{T}\mid \x) = \int q_0(\z^0 \mid \x) \prod_{t=1}^{T} q(\z^t \mid \x, \z^{t-1}) \dif \z^{t-1}$.
\begin{align*}
    \log \pt(\x) &\geq \Lc_T(\x)\\
    \Lc_T(\x) &= \E_{ q_T(\z^T\mid \x)} \left[ \log  \frac{p(\x, \z^T)}{q_T(\z^T \mid \x)} \right]\\
    &= \E_{ q_T(\z^T\mid \x)} \left[ \log  \frac{p(\x, \z^T)}{q_T(\z^T \mid \x)} + \E_{q(\z^{0 \ldots T-1} \mid \x, \z^T)} \left[ \log \frac{r(\z^{0 \ldots T-1} \mid \x, \z^T) q(\z^{0 \ldots T-1} \mid \x, \z^T) }{ q(\z^{0 \ldots T-1} \mid \x, \z^T) r(\z^{0 \ldots T-1} \mid \x, \z^T)} \right] \right]\\
    &= \E_{ q(\z^0, \ldots, \z^T \mid \x)} \left[ \log  \frac{p(\x, \z^T)}{q_T(\z^T \mid \x)} + \log \frac{r(\z^{0 \ldots T-1} \mid \x, \z^T) }{ q(\z^{0 \ldots T-1} \mid \x, \z^T) } \right]\\
    &\phantom{=} + \E_{q_T(\z^T \mid \x)} \left[ \KLD{q(\z^{0 \ldots T-1} \mid \x, \z^T)}{r(\z^{0 \ldots T-1} \mid \x, \z^T)} \right]\\
    &= \E_{ q(\z^0, \ldots, \z^T \mid \x)} \left[ \log  p(\x, \z^T) + \log \frac{r(\z^{0 \ldots T-1} \mid \x, \z^T) }{ q(\z^{0 \ldots T} \mid \x) } \right]\\
    &\phantom{=} + \E_{q_T(\z^T \mid \x)} \left[ \KLD{q(\z^{0 \ldots T-1} \mid \x, \z^T)}{r(\z^{0 \ldots T-1} \mid \x, \z^T)} \right]\\ 
    &= \E_{ q(\z^0, \ldots, \z^T \mid \x)} \left[ \log  \frac{p(\x, \z^T)}{q_0(\z^0 \mid \x)} + \log \frac{\prod_{t=1}^T r(\z^{t-1} \mid \x, \z^t) }{ \prod_{t=1}^T q_t(\z^t \mid \x, \z^{t-1}) } \right]\\
    &\phantom{=} + \E_{q_T(\z^T \mid \x)} \left[ \KLD{q(\z^{0 \ldots T-1} \mid \x, \z^T)}{r(\z^{0 \ldots T-1} \mid \x, \z^T)} \right]\\ 
    &= \E_{ q(\z^0, \ldots, \z^T \mid \x)} \left[ \log  \frac{p(\x, \z^T)}{q_0(\z^0 \mid \x)} + \sum_{t = 1}^T \log \frac{r(\z^{t-1} \mid \x, \z^t) }{ q_t(\z^t \mid \x, \z^{t-1}) } \right]\\
    &\phantom{=} + \E_{q_T(\z^T \mid \x)} \left[ \KLD{q(\z^{0 \ldots T-1} \mid \x, \z^T)}{r(\z^{0 \ldots T-1} \mid \x, \z^T)} \right]\\ 
    &= \Lmcvi(\x) + \E_{q_T(\z^T \mid \x)} \left[ \KLD{q(\z^{0 \ldots T-1} \mid \x, \z^T)}{r(\z^{0 \ldots T-1} \mid \x, \z^T)} \right]\\
    \log \pt(\x) &\geq \Lc_T(\x) \geq \Lmcvi(\x)
\end{align*}

Moreover, if we choose the reverse transition function $r$ in MCVI to be the true reverse operator, such that it produces the distribution of the chain at the previous step \citep[\eg][Chapter~1.4]{Murray2007}
\begin{align*}
    r(\z^{t-1} \mid \x, \z^t) &= \frac{q(\z^t \mid \x, \z^{t-1}) q_{t-1}(\z^{t-1} \mid \x)}{ q_t(\z^t \mid \x)},
\end{align*}
then from MCVI we can recover the tighter lower-bound in \eqref{eq:elbo-marginal-markov-chain}
\begin{align*}
    \Lmcvi(\x) &=\E_{q(\z_0, \ldots, \z^T \mid \x)} \left[ \log \frac{p(\x, \z^T)}{q(\z^0 \mid \x)} + \sum_{t = 1}^T \log \frac{r(\z^{t-1} \mid \x, \z^t) }{q (\z^t \mid \x, \z^{t-1})}  \right]\\
    &=\E_{q(\z_0, \ldots, \z^T \mid \x)} \left[ \log \frac{p(\x, \z^T)}{q(\z^0 \mid \x)} + \sum_{t = 1}^T \log \frac{q(\z^t \mid \x, \z^{t-1}) q_{t-1}(\z^{t-1} \mid \x) }{q_t(\z^t \mid \x)  q (\z^t \mid \x, \z^{t-1})}  \right]\\
    &=\E_{q(\z_0, \ldots, \z^T \mid \x)} \left[ \log \frac{p(\x, \z^T)}{q(\z^0 \mid \x)} + \sum_{t = 1}^T \log \frac{q(\z^t \mid \x, \z^{t-1}) q_{t-1}(\z^{t-1} \mid \x) }{q_t(\z^t \mid \x)  q (\z^t \mid \x, \z^{t-1})}  \right]\\
    &=\E_{ q_T(\z^T\mid \x)} \left[ \log  \frac{p(\x, \z^T)}{q_T(\z^T \mid \x)} \right]\\
    &=\Lc_T(\x)
\end{align*}

Hence, having to learn a reverse model $r$ loosens the lower-bound on $\log \pt(\x)$, which makes maximising the likelihood more difficult. 
The VGI method, on the other hand, does not require learning a model $r$ to reverse the Markov sampling path, and optimises a lower-bound that asymptotically converges to $\Lc_T(\x)$.

\section{Expectation-Maximisation for Factor Analysis with Incomplete Data}
\label{apx:em-for-fa}

We here derive the EM solution for the FA model with incomplete data.
From Section~\ref{sec:fa-model} we know that the FA model is a Gaussian model, with zero-mean standard Gaussian latent variables $\z$ and a linear Gaussian generative model $p(\x \mid \z) = \N(\x; \Fb \z + \mub, \Psib)$. We assume that $\x$ has observed $\xo$ and missing $\xm$ components, and write down the joint distribution 
\begin{align}
    p\left(\begin{pmatrix}
            \xo\\\xm\\\z
        \end{pmatrix}
    \right) 
    = \N \left( 
        \begin{pmatrix}
            \xo\\\xm\\\z
        \end{pmatrix};
        \begin{pmatrix}
            \muo \\ \mum \\ \bm{0}
        \end{pmatrix},
        \begin{bmatrix}
            \Fo \Fo^\top + \Psio & \Fo \Fm^\top & \Fo \\
            \Fm \Fo^\top & \Fm \Fm^\top + \Psim & \Fm \\
            \Fo^\top & \Fm^\top & \bm{I}
        \end{bmatrix}
    \right). \label{eq:fa-joint}
\end{align}

Following the standard EM procedure \citep[\eg][Chapter~11]{Barber2017} we write down the energy, where we remove a constant term due to the prior on $\z$
\begin{align}
    E(\thetab; \thetab^{(t-1)}) &= \sum_{i=1}^N \E_{p_{\thetab^{(t-1)}}(\xm, \z \mid \xo^i)}  \left[ \log \pt(\xo^i, \xm \mid \z) \right] \nonumber\\
    &= - \frac{1}{2} \sum_{i=1}^N \E \left[ d\log (2 \pi) + \log \det \abs{\Psib} + (\x^i - \mub - \Fb\z)^\top \Psib^{-1} (\x^i - \mub - \Fb\z) \right] \nonumber\\
    &\propto -\frac{N}{2}\log \det \abs{\Psib} - \frac{1}{2} \sum_{i=1}^N \E \left[ (\x^i - \mub - \Fb\z)^\top \Psib^{-1} (\x^i - \mub - \Fb\z) \right] \label{eq:fa-em-energy},
\end{align}
where $\thetab = (\Fb, \mub, \Psib)$ and $\thetab^{(t-1)}$ is the current estimate of $\thetab$.
The expectation operator $\E[\cdot]$ in \eqref{eq:fa-em-energy} means expectation with respect to $p_{\thetab^{(t-1)}}(\xm, \z \mid \xo^i)$, which is 
\begin{align}
    p_{\thetab^{(t-1)}}\left( \begin{pmatrix}
        \xm \\ \z
    \end{pmatrix} \;\middle|\; \xo^i \right) &= 
    \N \left(
        \begin{pmatrix}
            \xm \\ \z
        \end{pmatrix};
        \begin{pmatrix}
            \mub_{\mis\mid \obs} \\ \mub_{\text{z}\mid \obs}
        \end{pmatrix},
        \begin{bmatrix}
            \Sigmab_{\mis\mid \obs} & \Cb_{\mis, z\mid \obs}\\
            \Cb_{\mis, z\mid \obs}^\top & \Sigmab_{z\mid \obs}
        \end{bmatrix}
    \right). \label{eq:fa-em-prob-xm-z}
\end{align}
In the E-step of the algorithm we compute the statistics required in the above from \eqref{eq:fa-joint} using the rules for conditionals of multivariate Gaussian distributions \citep{Petersen2012}.
\begin{align}
    \Sigmab_{z\mid \obs} &= \bm{I} - \Fo^\top(\Fo \Fo^\top + \Psio)^{-1}\Fo = (\bm{I} + \Fo^\top \Psio^{-1} \Fo)^{-1} \label{eq:fa-em-stats-sigmazobs}\\
    \mub_{z \mid \obs} &= \Fo^\top (\Fo \Fo^\top + \Psio)^{-1}(\xo - \muo) = \Sigmab_{z\mid \obs} \Fo^\top \Psio^{-1} (\xo - \muo) \label{eq:fa-em-stats-muzobs}\\
    \Sigmab_{\mis \mid \obs} &= \Fm \Fm^\top + \Psim - \Fm \Fo^\top (\Fo \Fo^\top + \Psio)^{-1} \Fo \Fm^\top \nonumber\\
    &= \Fm \Sigmab_{z \mid \obs} \Fm^\top + \Psim \nonumber\\
    \mub_{\mis \mid \obs} &= \mum + \Fm \Fo^\top (\Fo \Fo^\top + \Psio)^{-1}(\xo - \muo) = \mum + \Fm \mub_{z \mid \obs} \nonumber\\
    \Cb_{\mis, z \mid \obs} &= \Fm - \Fm \Fo^\top (\Fo \Fo^\top + \Psio)^{-1} \Fo = \Fm \Sigmab_{z \mid \obs}, \nonumber
\end{align}
where in \eqref{eq:fa-em-stats-sigmazobs} we used Woodbury's identity \citep[\eg][]{Petersen2012} and in \eqref{eq:fa-em-stats-muzobs} we used the push-through identity \citep{Henderson1981}. Also note that these statistics depend on the observed data $\xo^i$ or the observed dimensions $\m$, we suppressed the index $i$ in the above notation.

In the M-step of the algorithm we maximise the energy with respect to $\thetab$. First, to derive the solution for $\mub$ we write down a simplified energy term where $\z$ are marginalised out
\begin{align}
    E_{\x}(\thetab; \thetab^{(t-1)}) &= \sum_{i=1}^N \E_{p_{\thetab^{(t-1)}}(\xm \mid \xo^i)} \left[ \log \pt(\xo^i, \xm) \right] \nonumber\\
    &= -\frac{1}{2} \sum_{i=1}^N \E \left[ d \log (2 \pi) + \log \det \abs{\Sigmab} + (\x^i - \mub)^\top \Sigmab^{-1} (\x^i - \mub) \right]. \label{eq:fa-em-energy-mu}
\end{align}
The expectation operator $\E[\cdot]$ in the above equation depends on $p_{\thetab^{(t-1)}}(\xm \mid \xo^i)$, which can be directly read from \eqref{eq:fa-em-prob-xm-z} as the marginal of a Gaussian \citep[\eg][Section~8.1.2]{Petersen2012}.
Then, taking the partial derivative and setting it to zero we get $\hat \mub$
\begin{align}
    \frac{\partial E_{\x}(\thetab; \thetab^{(t-1)})}{\partial \mub} &= -\frac{1}{2} \sum_{i=1}^N \E \left[\frac{\partial}{\partial \mub} \left((\x^i - \mub)^\top \Sigmab^{-1} (\x^i - \mub) \right) \right] \nonumber\\
    &= - \sum_{i=1}^N \E \left[(\x^i - \mub)^\top \right] \Sigmab^{-1}\nonumber\\
    &= N \mub^\top \Sigmab^{-1} - \sum_{i=1}^N \E \left[ \x^i \right]^\top \Sigmab^{-1} = 0\nonumber\\
    \implies \hat \mub &= \frac{1}{N} \sum_{i=1}^N \hat \x^i, \label{eq:fa-em-mu-update}
\end{align}
where $\hat \x^i = (\xo^i, \mub_{\mis \mid \obs}^i)$. 

Now, from \eqref{eq:fa-em-energy} we derive the updated $\hat \Fb$
\begin{align*}
    \frac{\partial E(\thetab; \thetab^{(t-1)})}{\partial \Fb} &= -\frac{1}{2} \sum_{i=1}^N \E \left[\frac{\partial}{\partial \Fb} \left((\x^i - \mub - \Fb\z)^\top \Psib^{-1} (\x^i - \mub - \Fb\z) \right) \right] \\
    &= - \sum_{i=1}^N \E \left[(\x^i - \mub - \Fb\z) \z^\top \Psib^{-1} \right]\\
    &= - \sum_{i=1}^N \E \left[(\x^i - \mub) \z^\top \right] \Psib^{-1} + \Fb \sum_{i=1}^N \E \left[ \z \z^\top \right] \Psib^{-1} = 0\\
    \implies \hat \Fb \bm{H} &= \bm{A} \\
    \implies\phantom{\bm{H}} \hat \Fb &= \bm{A} \bm{H}^{-1}
\intertext{where}
    \bm{H} &= \frac{1}{N} \sum_{i=1}^N \E \left[ \z \z^\top \right]
    = \frac{1}{N} \sum_{i=1}^N \left( \Sigmab_{z \mid \obs}^i + \mub_{z \mid \obs}^i \mub_{z \mid \obs}^{i\top} \right) \\
    \bm{A} &= \frac{1}{N} \sum_{i=1}^N \E \left[(\x^i - \hat \mub) \z^\top \right] 
    = \frac{1}{N} \sum_{i=1}^N \E \left[\x^i\z^\top \right] - \hat \mub \mub_{z \mid \obs}^{i\top}\\
    &= \frac{1}{N} \sum_{i=1}^N \bar \Cb_{\mis, z \mid \obs}^i + \hat \x^i \mub_{z \mid \obs}^{i\top} - \hat \mub \mub_{z \mid \obs}^{i\top} \\
    &= \frac{1}{N} \sum_{i=1}^N \bar \Cb_{\mis, z \mid \obs}^i + (\hat \x^i - \hat \mub)\mub_{z \mid \obs}^{i\top},
\end{align*}
and $\bar \Cb_{\mis, z \mid \obs}$ is $\Cb_{\mis, z \mid \obs}$ with $0$-rows for the observed dimensions.

Now we derive solution for $\hat \Psib$, using $\text{diag}()$ to denote a function that sets off-diagonal elements to zero
\begin{align*}
    \frac{\partial E(\thetab; \thetab^{(t-1)})}{\partial \Psib} &= - \frac{\partial}{\partial \Psib} \frac{N}{2} \log \det \abs{\Psib} \\
    &\phantom{=}- \text{diag} \left( \sum_{i=1}^N \E \left[ \frac{\partial}{\partial \Psib} \left\{ \frac{1}{2} (\x^i - \mub - \Fb\z)^\top \Psib^{-1} (\x^i - \mub - \Fb\z) \right\} \right] \right)\\
    &= -\frac{N}{2} \Psib^{-1} + \text{diag} \left( \sum_{i=1}^N \E \left[ \frac{1}{2} \Psib^{-1} (\x^i - \mub - \Fb\z) (\x^i - \mub - \Fb\z)^\top \Psib^{-1} \right] \right) = 0\\
    \implies \hat \Psib &= \text{diag} \left( \frac{1}{N} \sum_{i=1}^N \E \left[ (\x^i - \hat \mub - \Fb\z) (\x^i - \hat \mub - \Fb\z)^\top \right] \right)\\
    &= \text{diag} \left( \frac{1}{N} \sum_{i=1}^N \E \left[ (\x^i - \hat \mub) (\x^i - \hat \mub)^\top - 2\Fb\z(\x^i - \hat \mub)^\top + \Fb\z\z^\top \Fb^\top \right]\right) \\
    &= \text{diag} \left( \frac{1}{N} \sum_{i=1}^N \E \left[ (\x^i - \hat \mub) (\x^i - \hat \mub)^\top \right] - 2\Fb\bm{A}^\top + \Fb \bm{H} \Fb^\top \right)\\
    &= \text{diag} \left( \frac{1}{N} \sum_{i=1}^N \E \left[ \x^i \x^{i\top} - 2\x^i \hat \mub^\top +\hat \mub \hat \mub^\top \right] - 2\Fb \bm{A}^\top + \Fb \bm{H} \Fb^\top \right)\\
    &= \text{diag} \left( \bm{V} - 2\Fb\bm{A}^\top + \Fb \bm{H} \Fb^\top \right),\\
    \intertext{with}
    \bm{V} &= \frac{1}{N} \sum_{i=1}^N \E \left[ \x^i \x^{i\top}\right] - 2\hat \x^i \hat \mub^\top +\hat \mub \hat \mub^\top\\
    &= \frac{1}{N} \sum_{i=1}^N \bar \Sigmab_{\mis \mid \obs}^i + \hat \x^i \hat \x^{i\top} - 2\hat \x^i \hat \mub^\top +\hat \mub \hat \mub^\top\\
    &= \frac{1}{N} \sum_{i=1}^N \bar \Sigmab_{\mis \mid \obs}^i + (\hat \x^i - \hat \mub)(\hat \x^i - \hat \mub)^\top,
\end{align*}
and where $\bar \Sigmab_{\mis \mid \obs}^i$ is $\Sigmab_{\mis \mid \obs}^i$ with zero rows and columns for the observed dimensions in $\x^i$.
The solutions from the current iteration are then set to $\thetab^{(t)} = (\hat \Fb, \hat \mub, \hat \Psib)$ and used in the next iteration.

Similar EM update rules for FA with incomplete data have been derived by \citet[][Chapter~4.3.2]{marlinMissingDataProblems2008}, however our update of the mean vector in \eqref{eq:fa-em-mu-update} uses a tighter lower-bound, by marginalising the latents from the energy in \eqref{eq:fa-em-energy-mu}, while theirs uses a looser lower-bound that results in the use of the energy in \eqref{eq:fa-em-energy}. Thus, we can expect that using our update rule should converge faster.
Finally, we note that our derivation closely follows the derivation of EM for FA with complete data by \citet[][Chapter~21.2.2]{Barber2017} with extra terms arising due to incomplete data.

\section{Toy Data Ground Truth}
\label{apx:toy-ground-truth}
The toy data was generated using a FA model with the following parameters as the ground truth:
\begin{align*}
    \Fb = 
    \begin{bmatrix}
    -5 & -2 \\
     \phantom{-}4 &  \phantom{-}0 \\
    -3 & -1 \\
    -3 & -3 \\
     \phantom{-}1 &  \phantom{-}5 \\
    -1 &  \phantom{-}2
    \end{bmatrix}
    \quad
    \mub = 
    \begin{bmatrix}
     \phantom{-}3 \\
    -1 \\
     \phantom{-}0 \\
     \phantom{-}2 \\
    -1 \\
     \phantom{-}0
    \end{bmatrix}
    \quad
    \psidiag = 
    \begin{bmatrix}
    50.4794 \\
    30.0988 \\
    6.766 \\
    17.3357\\
    40.9839\\
    25.1122
    \end{bmatrix}.
\end{align*}
The toy data intentionally has a low signal-to-noise ratio, which made the estimation problem harder.

\section{VAE Model Selection on Incomplete Held-Out Data}
\label{apx:vae-model-selection}

The primary goal of our investigation in Section~\ref{sec:vae-experiments} is to evaluate the accuracy of the VAE model $\pt(\x)$ to understand how well it estimates the distribution of the data.
Like all other amortised variational methods, the evaluation on held-out data requires fine-tuning of the variational models to the target data due to the inference generalisation gap discussed in Section~\ref{sec:vgi-eval-held-out}.
Hence, to perform model selection via validation on held-out data, we store checkpoints of all parameters (including the generator and the variational models) at predefined intervals during training. Then, we retrospectively fine-tune the variational models to the validation data and select, out of all checkpointed states, the generator parameters that performed best on the fine-tuned validation loss. 
For VGI we use the validation loss obtained via the fine-tuning algorithm in Appendix~\ref{apx:vgi-finetuning} and for the other methods we use their respective loss with the fine-tuned encoder networks.

In Figure~\ref{fig-apx:vae-finetuning-curves} of Appendix~\ref{apx:additional-figures} we show the corresponding objective curves during fine-tuning of the VGI methods and MVAE,\footnote{The fine-tuning loss curves of the other VAE-specific methods were similar to MVAE and hence are not shown.} where we observe that the fine-tuning of VGI-based methods converge significantly faster than MVAE. %
Moreover, in Figure~\ref{fig-apx:vae-learning-curves} of Appendix~\ref{apx:additional-figures} we show the validation (dashed) and fine-tuned validation (dash-dotted) learning curves. We can see that the inference generalisation gap affects all evaluated methods (note the significant gap between dashed and dash-dotted curves), hence confirming that fine-tuning of the variational distributions is necessary to perform model selection.

\section{Additional Figures}
\label{apx:additional-figures}

In this section we show additional figures from the experiments in Sections~\ref{sec:toy-experiments}-\ref{sec:flow-experiments}.

\begin{figure}[htb]
    \centering
    \includegraphics[width=1\linewidth]{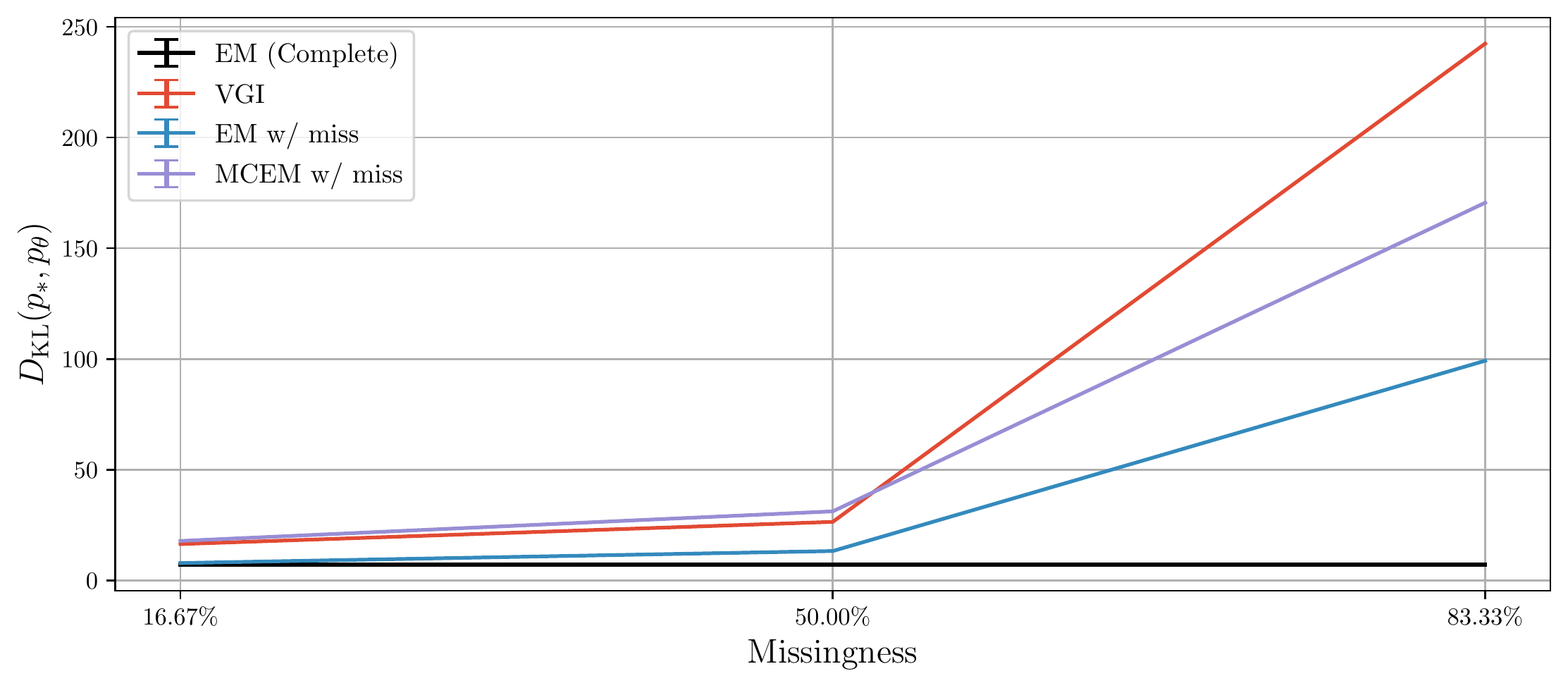}
    \caption{The accuracy of the fitted statistical models as $\KLD{\ps}{\pt}$ on FA-Frey data. MCEM samples imputations from the true conditional $\pt(\xm \mid \xo)$ using the learnt model $\pt$ and using SGA fits the model on the imputed data.
    Note that the curves for VGI (red) and MCEM (purple) are close, hence we attribute the performance gap between the EM and VGI to the stochastic optimisation.}
    \label{fig-apx:fafrey-kldiv-mcem}
\end{figure}

\begin{figure}[tpb]
    \centering
    \includegraphics[width=1\linewidth]{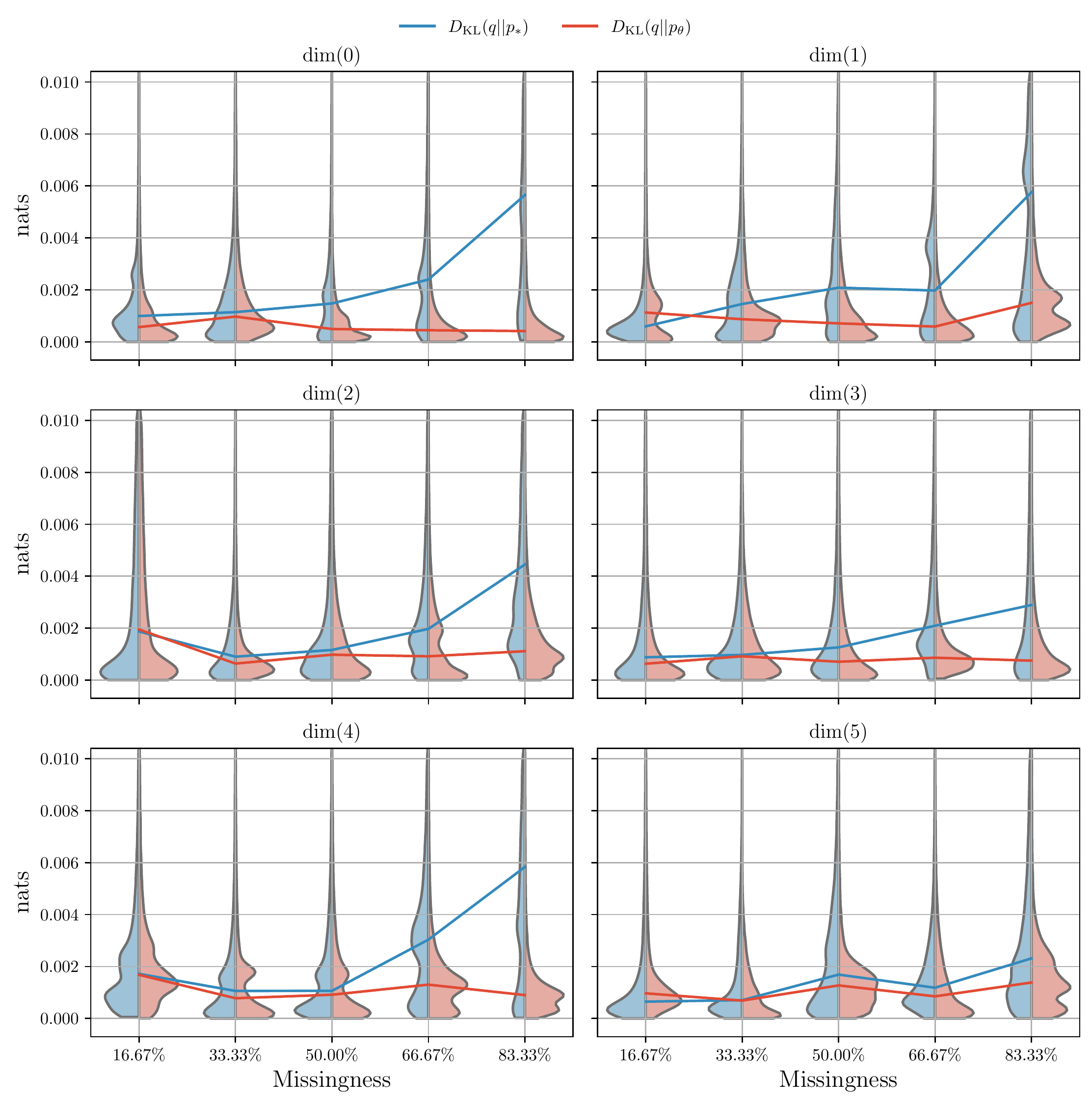}
    \caption{KL divergence, conditioned on the toy test set, between the univariate variational conditional distributions and the ground truth distribution for each feature dimension $\KLD{\qpj(\xj \mid \xnj)}{\ps(\xj \mid \xnj)}$ (blue), and the posterior under the learnt model $\KLD{\qpj(\xj \mid \xnj)}{\pt(\xj \mid \xnj)}$ (red). The lines show the median conditional KL divergence on the test set.}
    \label{fig-apx:toy-posterior-kldiv-separate}
\end{figure}

\begin{figure}[tpb]
        \includegraphics[width=0.495\textwidth]{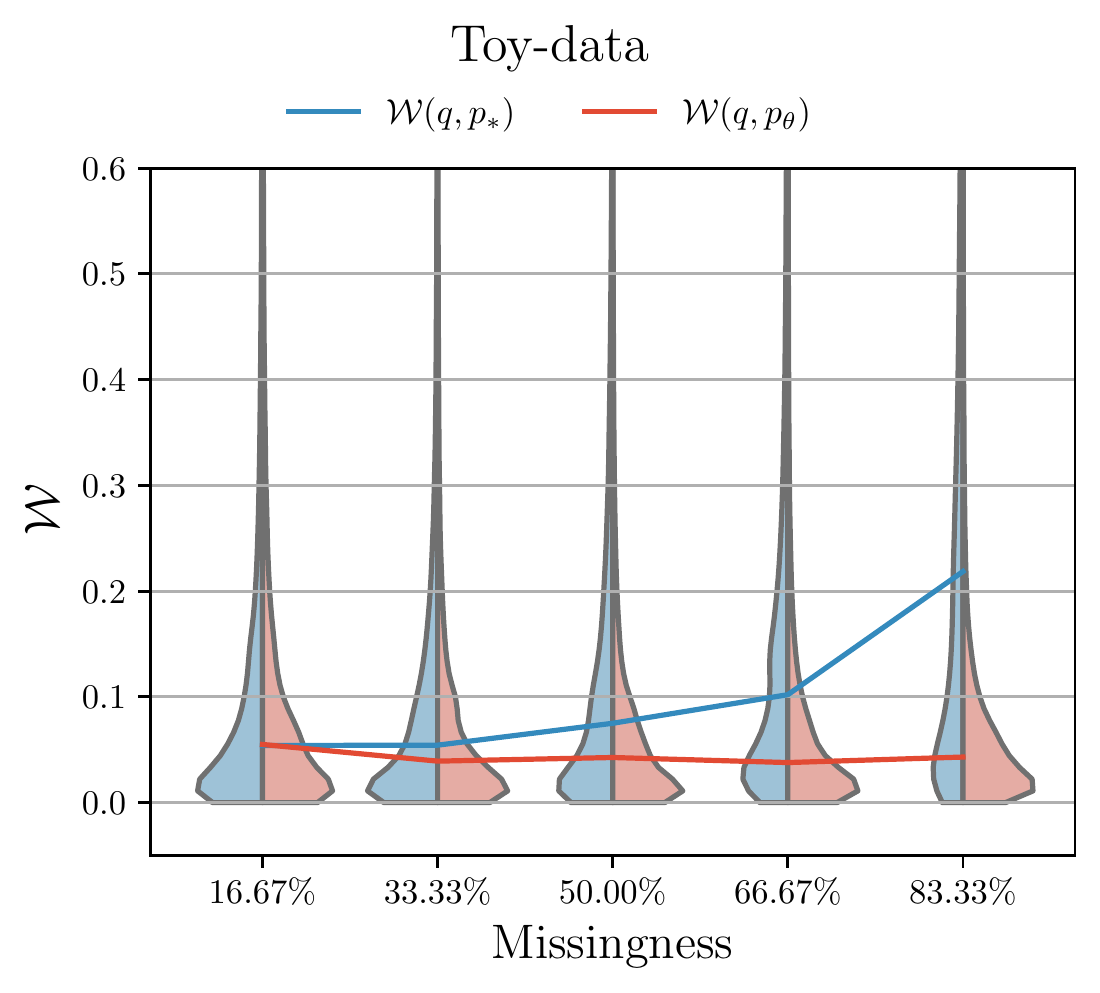}
        \includegraphics[width=0.495\textwidth]{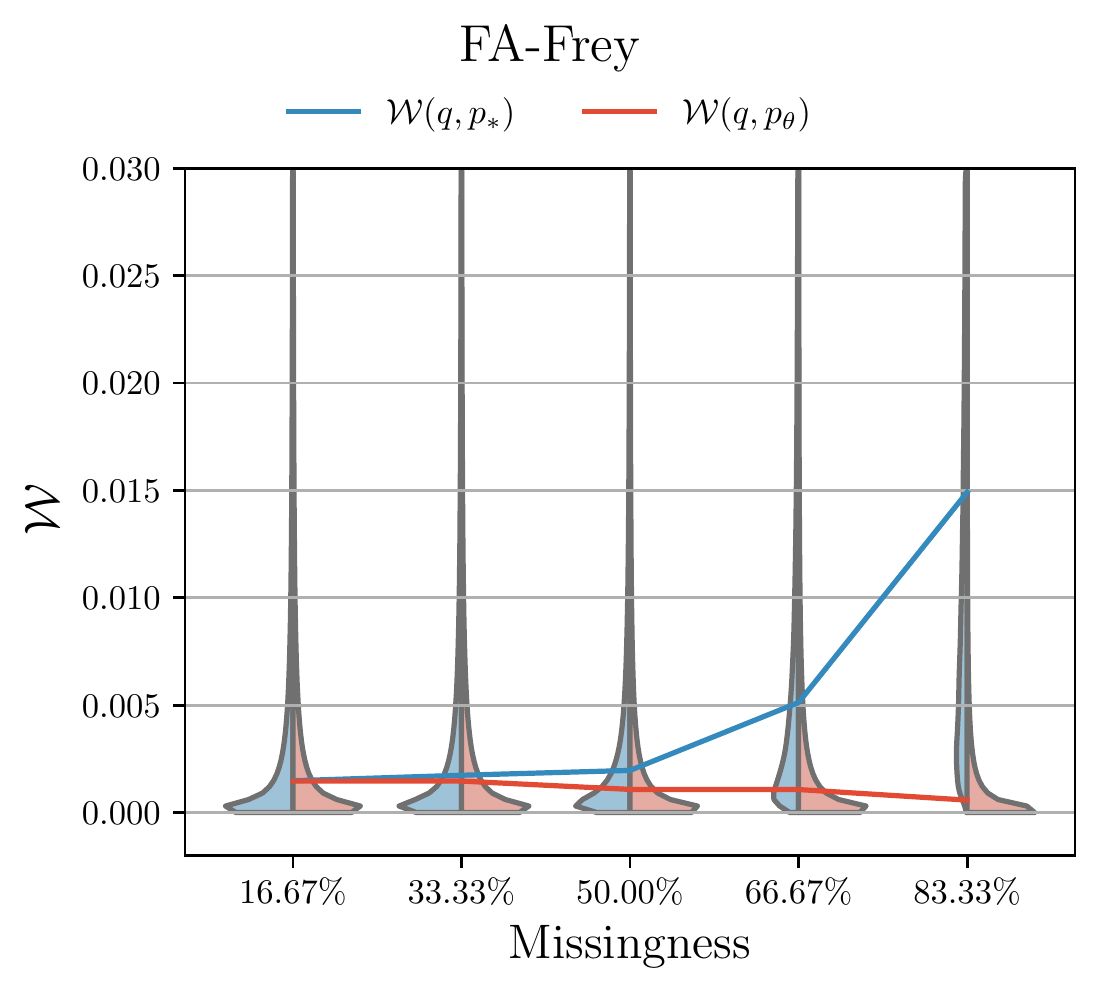}
    \caption{Wasserstein distance, conditioned on the test set (left: toy data, right: FA-Frey), between the univariate variational conditional distributions and the ground truth distribution $\mathcal{W}(\qpj(\xj \mid \xnj),\ps(\xj \mid \xnj))$ (blue), and the posterior under the learnt model $\mathcal{W}(\qpj(\xj \mid \xnj),\pt(\xj \mid \xnj))$ (red). The lines show the median Wasserstein distance conditioned on the test set.}
    \label{fig-apx:toy-posterior-wasserstein}
\end{figure}

\begin{figure}[tpb]
    \centering
    \includegraphics[width=1\linewidth]{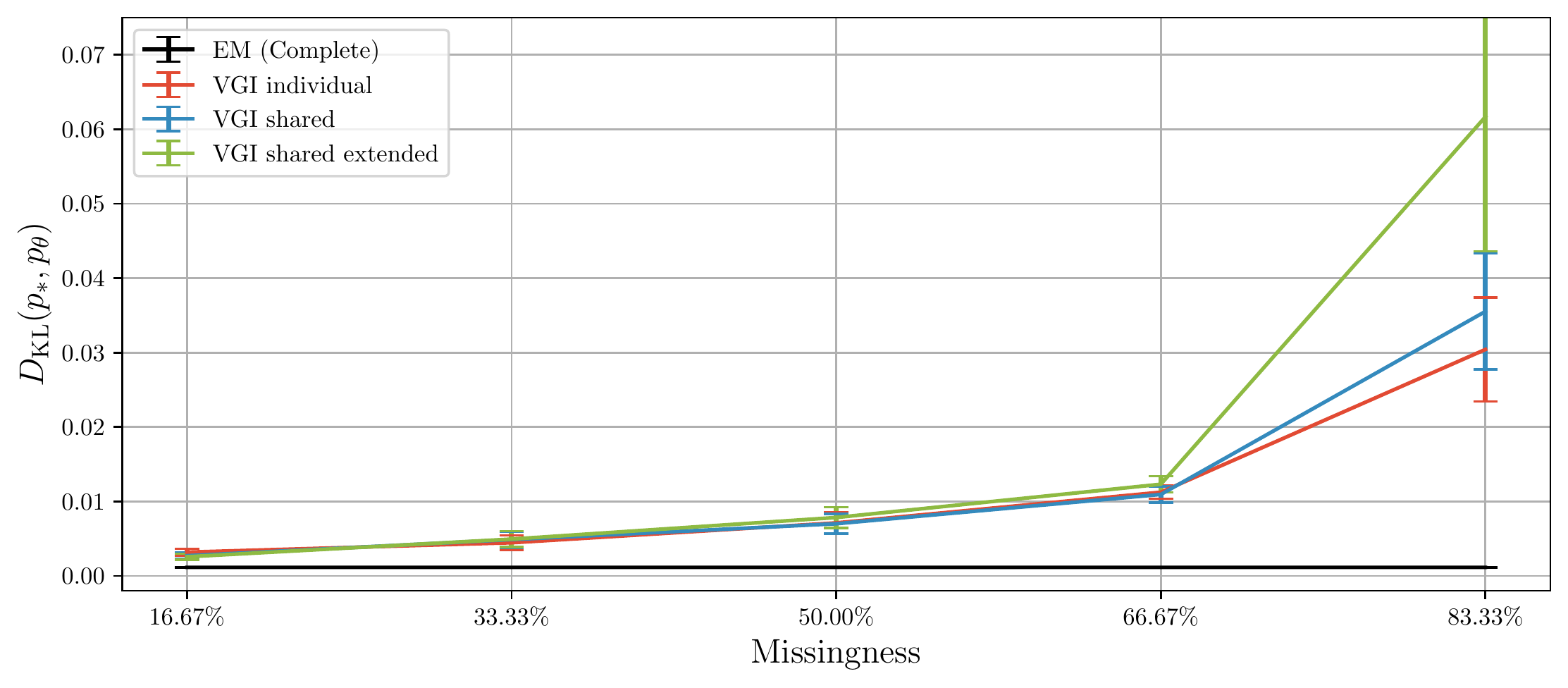}
    \caption{The accuracy of the fitted models on toy data, as measured by $\KLD{\ps}{\pt}$, comparing independent-weight variational models against shared-weight variational models.}
    \label{fig-apx:toy-fitted-model-results-individual-vs-shared}
\end{figure}

\begin{figure}[tpb]
    \centering
    \includegraphics[width=1\linewidth]{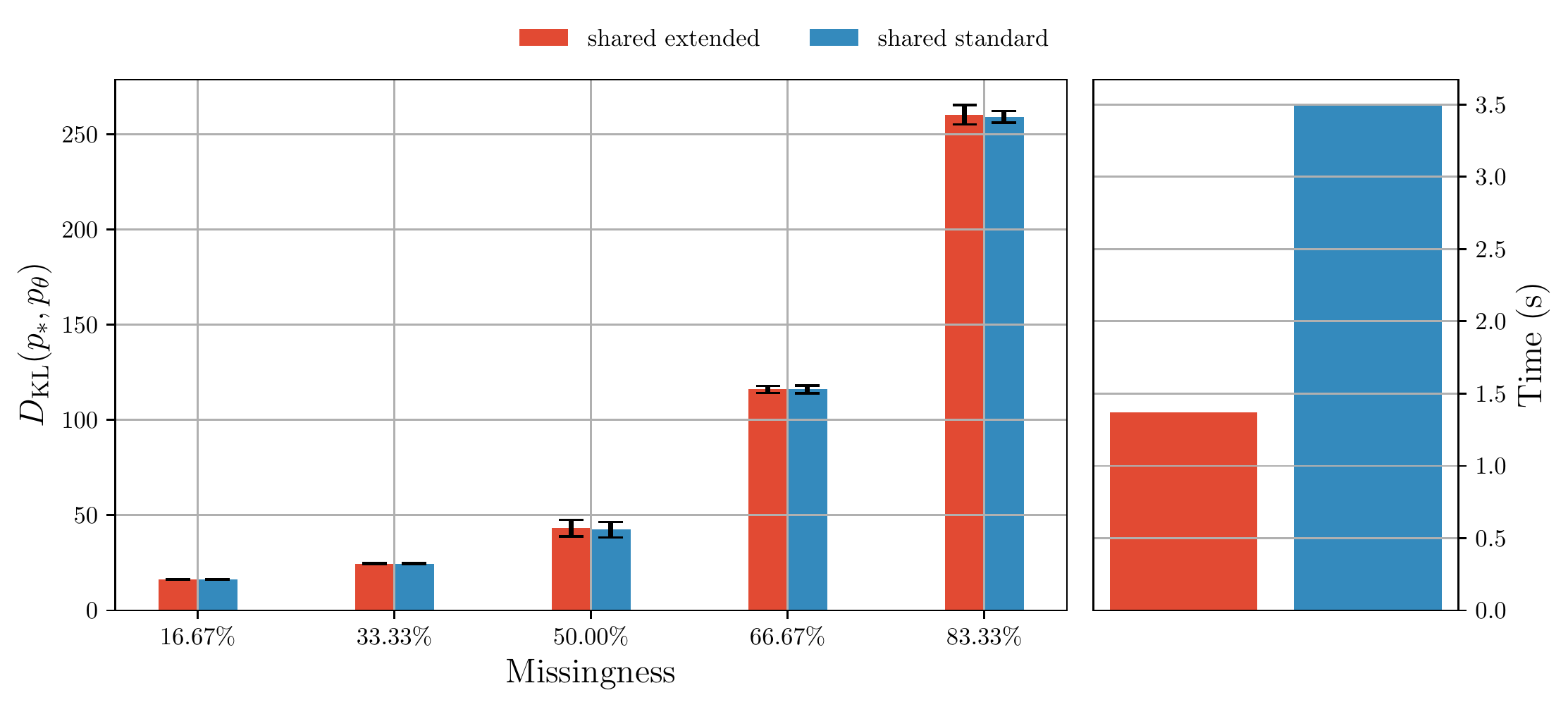}
    \caption{Comparison of VGI estimation accuracy on a FA model using a shared standard variational Gibbs model against a shared extended model, as discussed in Section~\ref{sec:variational-model}. Left: estimated KL divergence between the fitted and the ground truth model. Right: average training time per iteration.}
    \label{fig-apx:fa-time-and-kldiv-relaxed-vs-not-relaxed}
\end{figure}

\begin{figure}[p]
    \centering
    \includegraphics[width=1\linewidth]{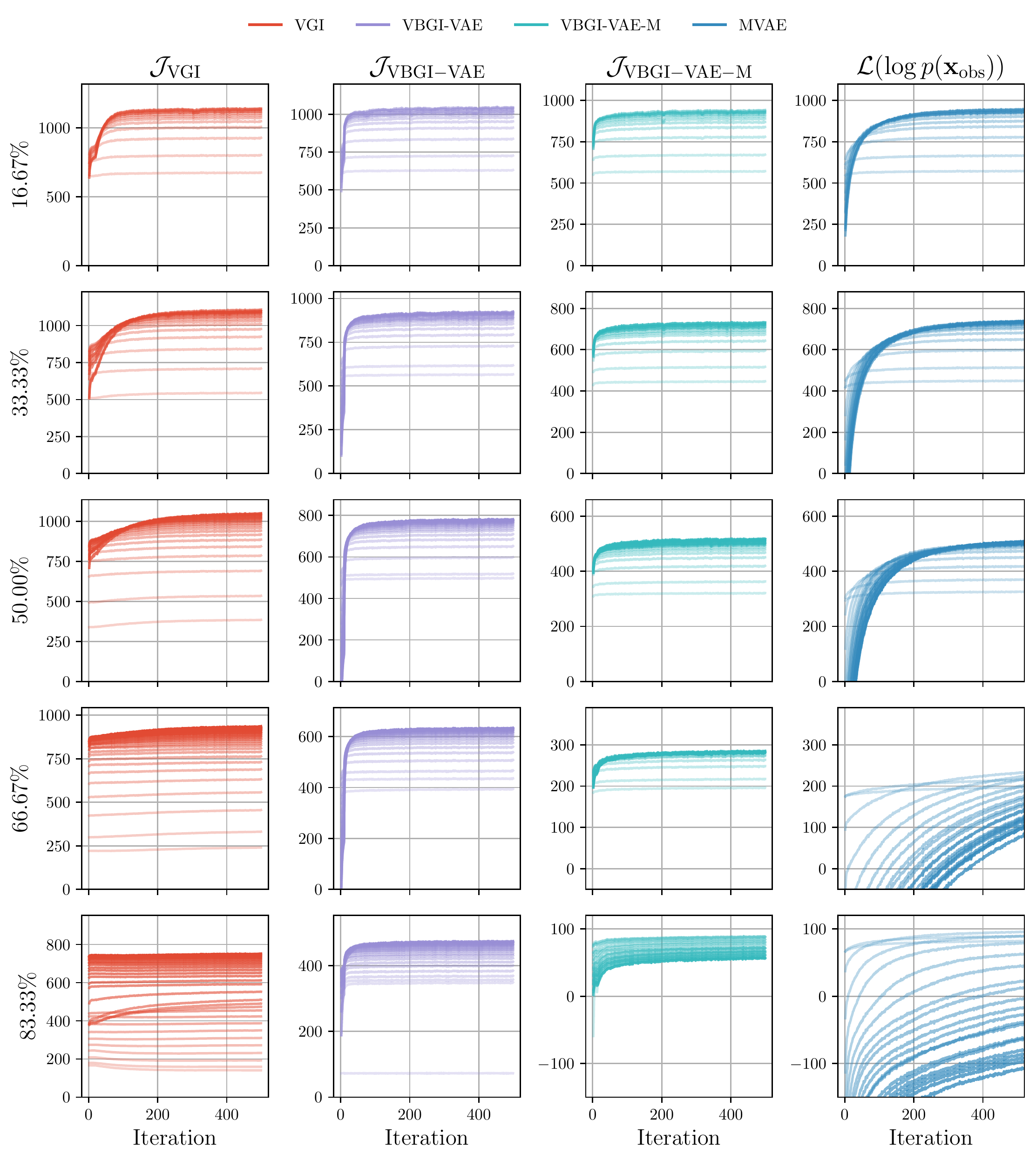}
    \caption{
    The validation fine-tuning loss curves on VAE-Frey data, where the intensity of the curve represents the training iteration at which the weight snapshot was taken. From an early iteration (less intense) to the last iteration (more intense). The fine-tuning of VGI-based experiments (columns 1-3) is comparatively faster then the VAE-specific methods (right-hand column, the curves of all VAE-specific methods were similar).}
    \label{fig-apx:vae-finetuning-curves}
\end{figure}

\begin{figure}[p]
    \centering
    \includegraphics[width=0.985\linewidth]{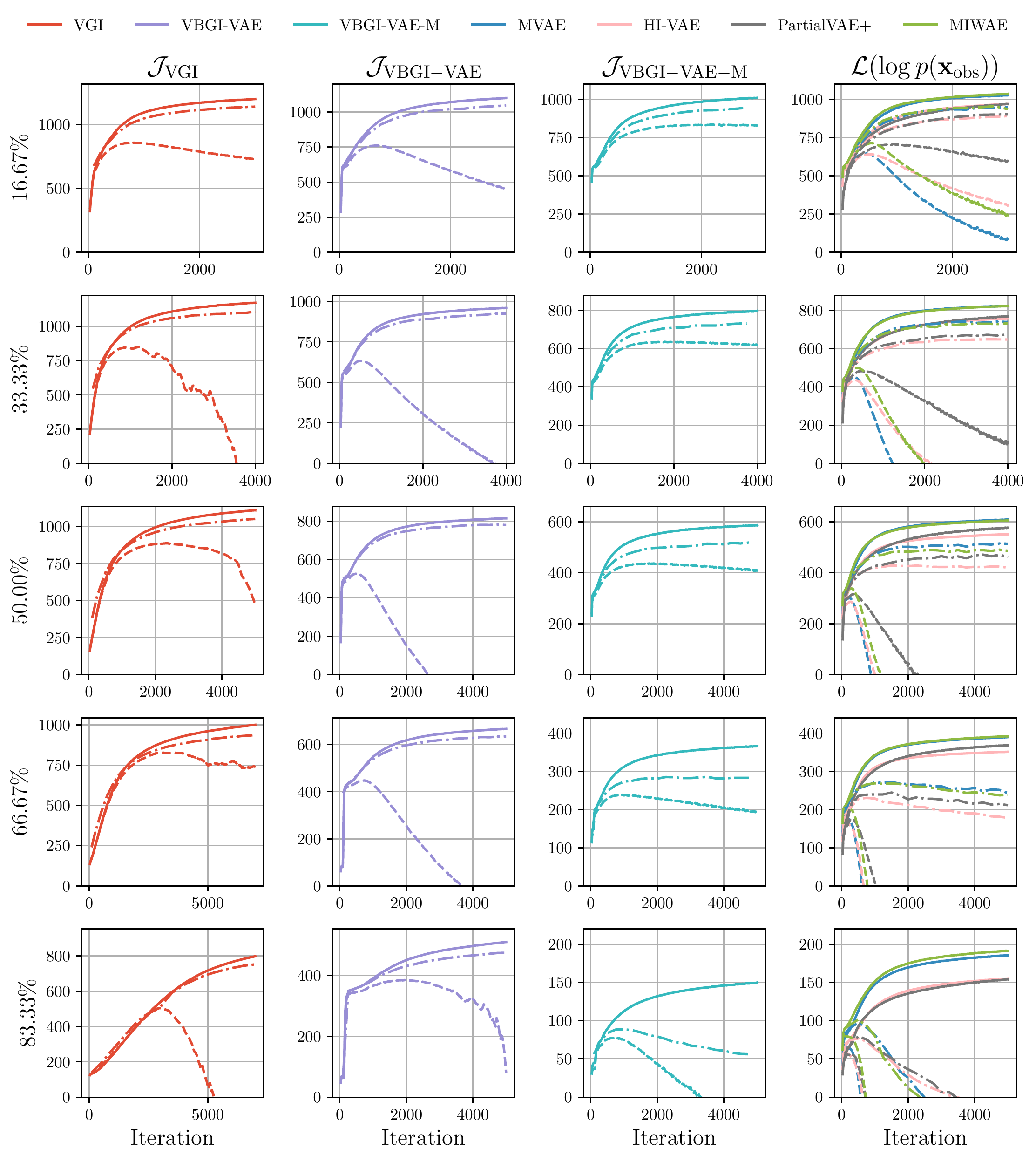}
    \caption{
    The training (solid), validation (dashed), and fine-tuned validation (dash-dotted) loss curves on VAE-Frey data. The gap between the validation and fine-tuned validation curves shows the inference generalisation gap. Note the difference in scale of the objectives, thus they should not be compared directly.
    Because of the generalisation gap, the validation curves before fine-tuning (dashed red) diverged for VGI when the number of variational models was large. For visualisation purposes, we thus only accepted Gibbs updates in the hypercube defined by the minimum and maximum values in the observed data. 
    }
    \label{fig-apx:vae-learning-curves}
\end{figure}

\begin{figure}[tpb]
    \centering
    \includegraphics[width=1\linewidth]{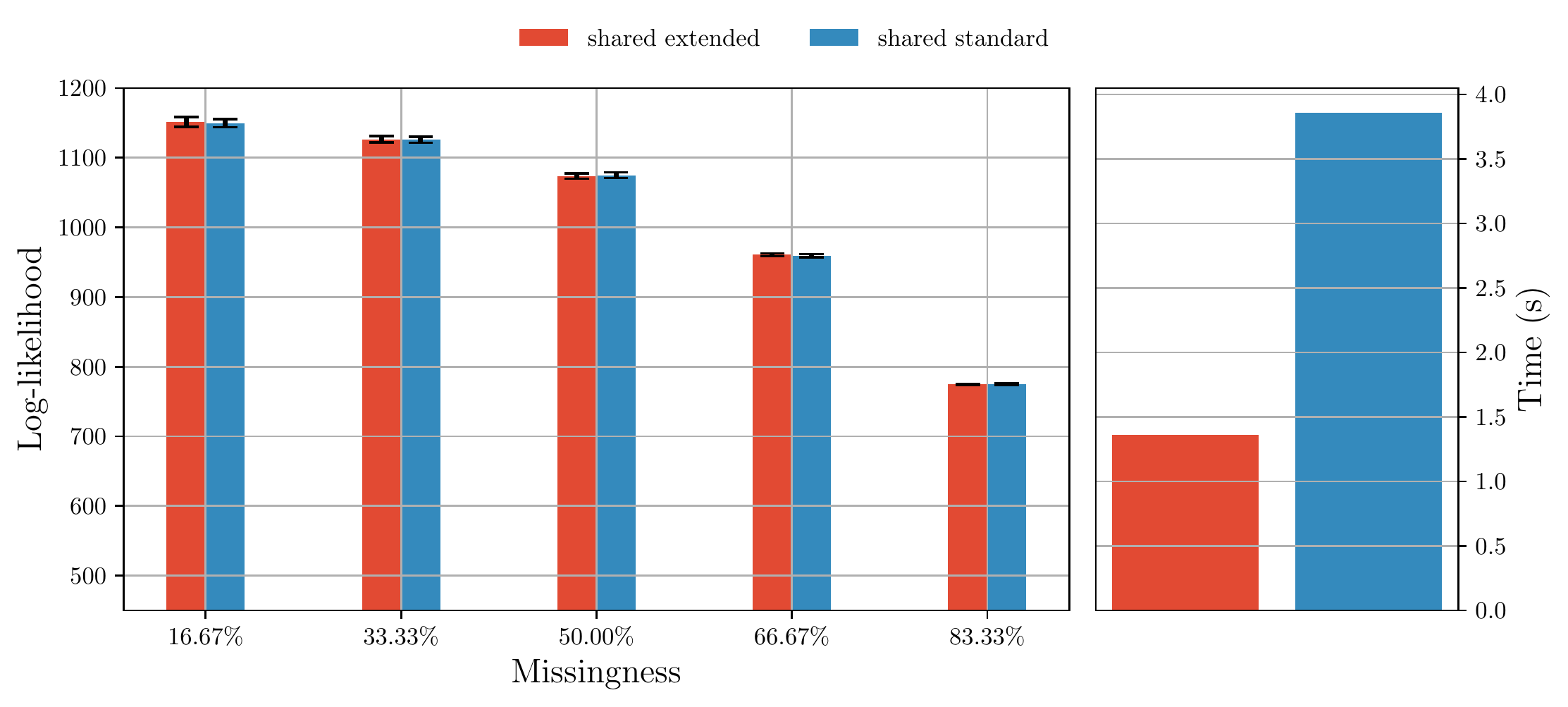}
    \caption{Comparison of VGI estimation on a VAE model using shared standard-Gibbs variational conditionals against shared extended-Gibbs model, as discussed in Section~\ref{sec:variational-model}. Left: estimated test log-likelihood of the fitted model. Right: average training time per iteration.}
    \label{fig-apx:vae-time-and-loglik-relaxed-vs-not-relaxed}
\end{figure}

\begin{figure}[tbp]
    \includegraphics[width=1\linewidth]{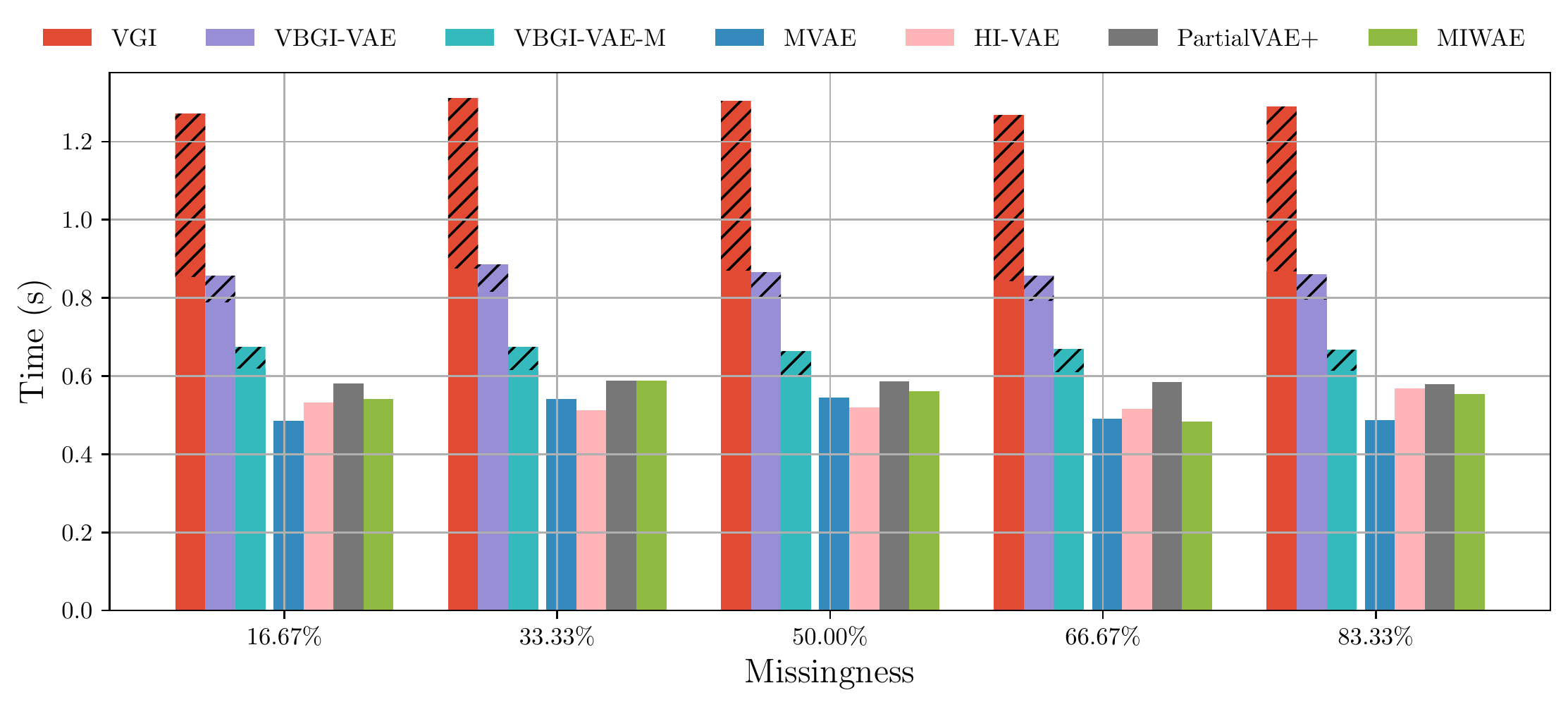}
    \caption{Average time of one training iteration in seconds on VAE-Frey data. The hatched part of the bars indicates the time spent on updating the imputations via pseudo-Gibbs sampling.}
    \label{fig-apx:fcvae-time-per-epoch}
\end{figure}

\begin{figure}[tpb]
    \centering
    \includegraphics[width=1\linewidth]{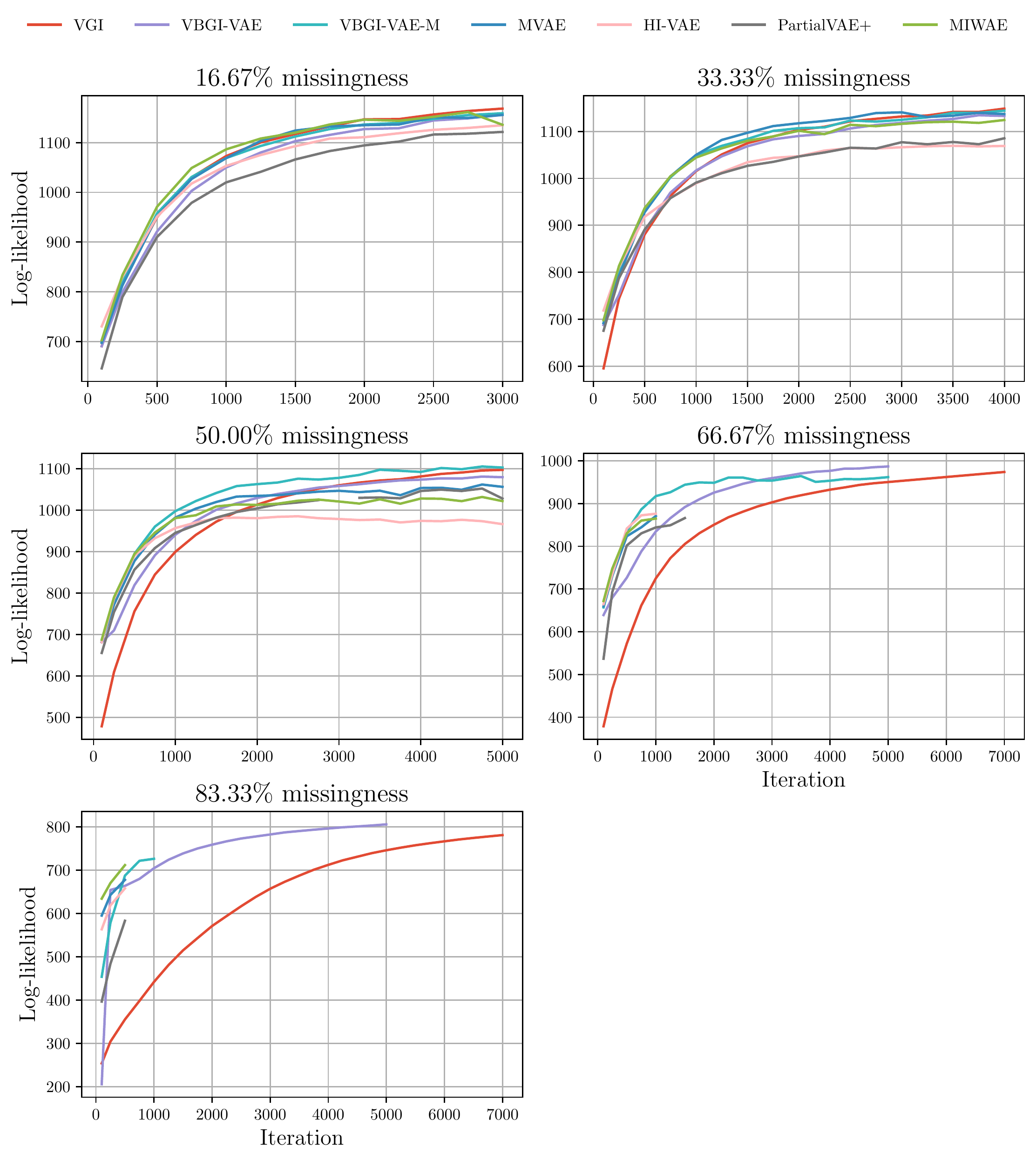}
    \caption{Estimated test log-likelihood against the training iteration on complete VAE-Frey data. Estimated using importance sampling as described in  Section~\ref{sec:vae-accuracy}, where the variational encoder was refitted to the test data to mitigate the inference generalisation gap.}
    \label{fig-apx:fcvae-test-loglik-vs-epoch-all}
\end{figure}

\begin{figure}[tpb]
    \centering
    \includegraphics[width=1\linewidth]{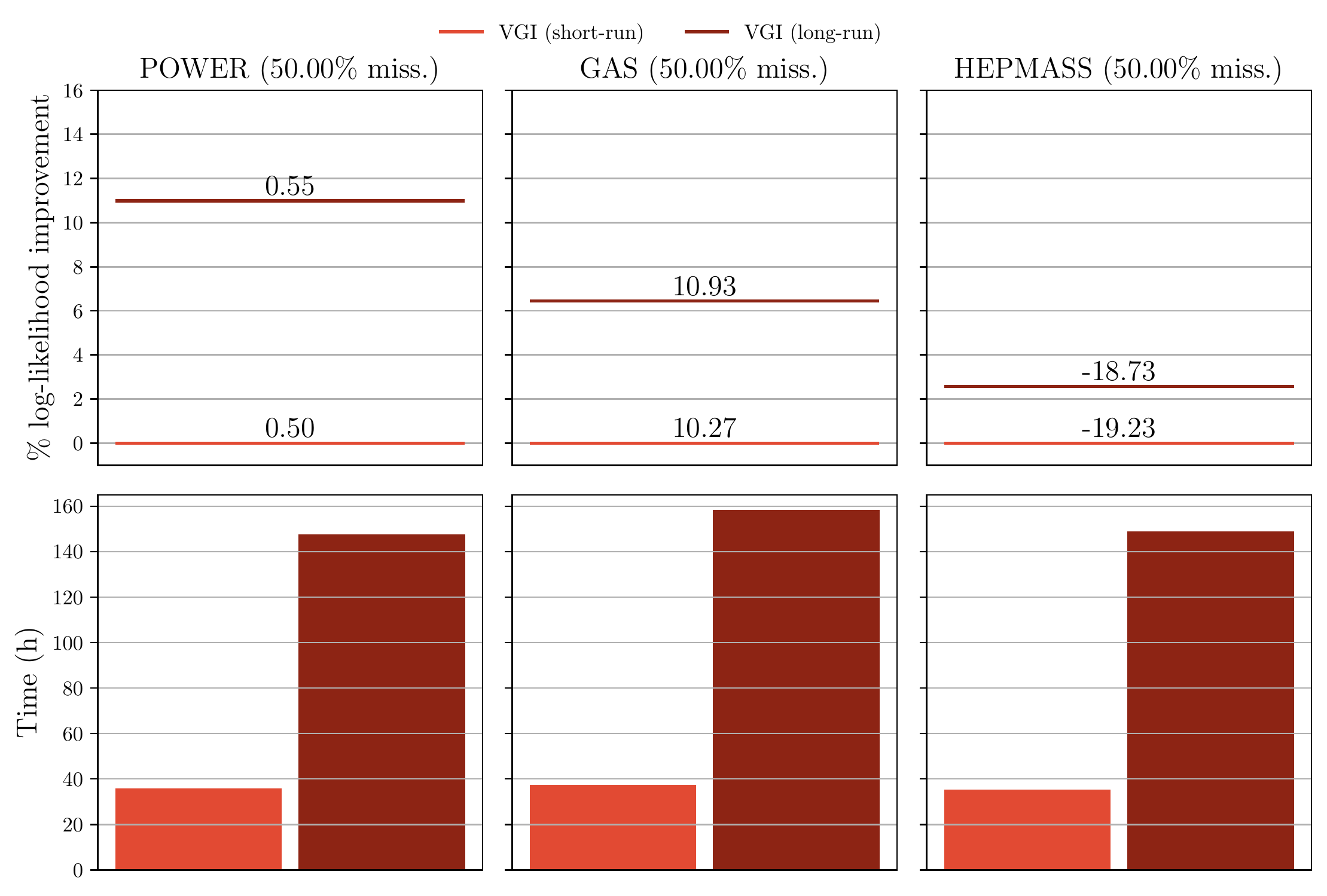}
    \caption{The percentage gain in test log-likelihood by running the VGI significantly longer. The numbers above the horizontal lines show the test log-likelihood, and the bars show the total training time. Other than the number of maximum iterations, the experimental setup was identical to the main experiments. The results show that the model fits in the main evaluation (Section~\ref{sec:flow-accuracy}) can be further improved by using a larger computational budget.}
    \label{fig-apx:rqcspline-longrun}
\end{figure}

\begin{figure}[tpb]
    \centering
    \includegraphics[width=1\linewidth]{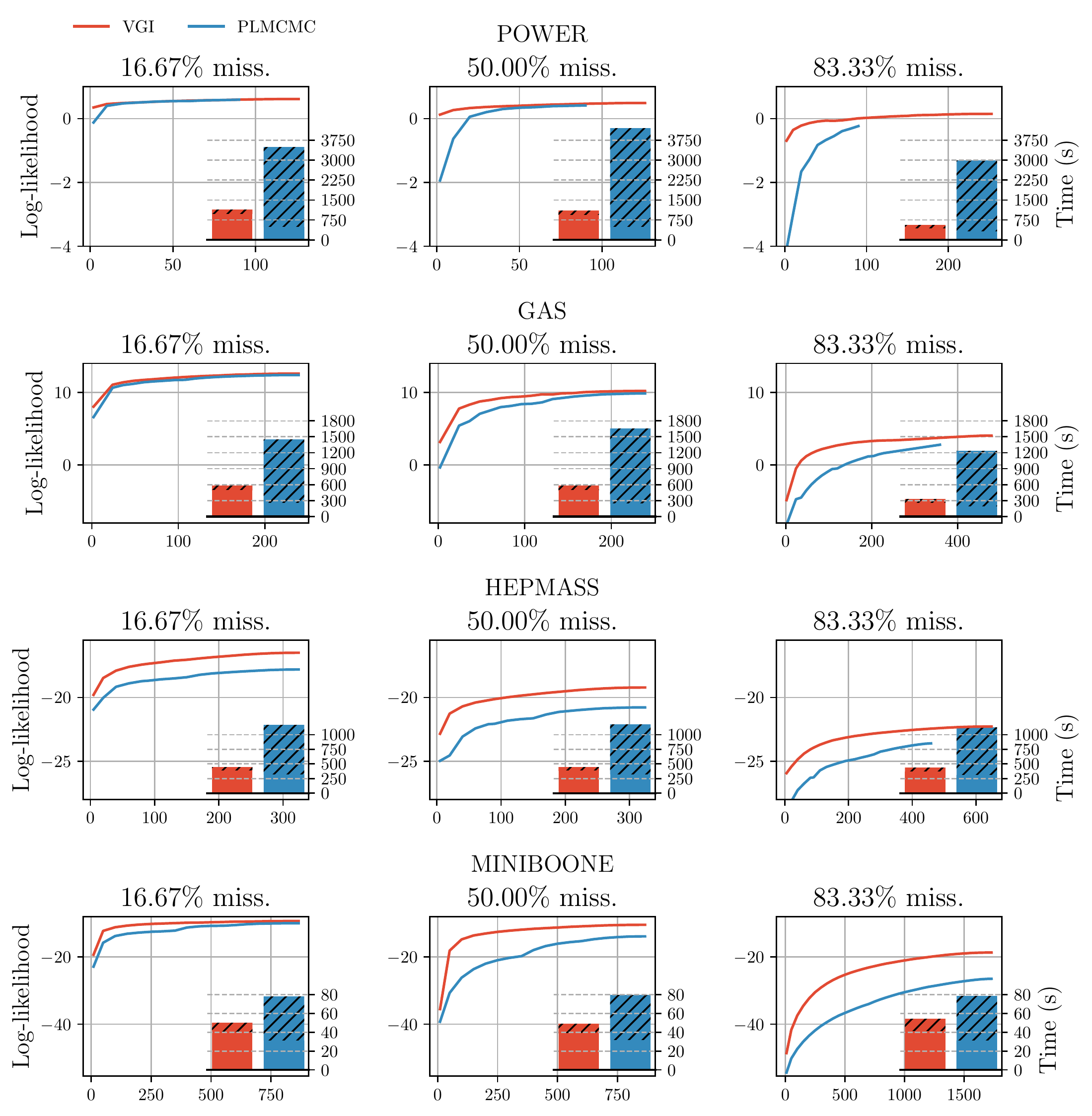}
    \caption{Estimated test log-likelihood on complete data against training iteration and average duration of one training iteration in seconds. The hatched part of the bars indicates the time spent on pseudo-Gibbs sampling in VGI and performing PLMCMC during MCEM. In all of the top row and the third column the PLMCMC log-likelihood curves stop earlier than VGI due to the computational budget, since the average iteration of PLMCMC was significantly more expensive than VGI.}
    \label{fig-apx:rqcspline-computational-aspects-all}
\end{figure}

\clearpage
\vskip 0.2in
\bibliography{references}

\end{document}